\documentclass{article} % For LaTeX2e
\usepackage{iclr2027_conference,times}

\usepackage{amsmath,amsfonts,bm}

\def\eqref#1{equation~\ref{#1}}
\def\1{\bm{1}}

\DeclareMathAlphabet{\mathsfit}{\encodingdefault}{\sfdefault}{m}{sl}
\SetMathAlphabet{\mathsfit}{bold}{\encodingdefault}{\sfdefault}{bx}{n}

\usepackage{wrapfig}
\usepackage{caption}
\usepackage{algorithm}
\usepackage{algorithmic}

\usepackage{xcolor}
\usepackage{hyperref}
\hypersetup{
    colorlinks=true,
    citecolor=blue!60!black,
    linkcolor=blue!60!black,
    urlcolor=blue!60!black
}

\usepackage{url}
\usepackage{booktabs}
\usepackage{arydshln}
\usepackage{multirow}
\usepackage{graphicx} 
\usepackage{nicematrix}
\usepackage{enumitem}
\usepackage{tikz}
\newcommand{\xhdr}[1]{\noindent\textbf{#1.}}

\newcommand{\llmqwen}{\textsc{\small Qwen2.5-7B-Instruct}}
\newcommand{\llmmistral}{\textsc{\small Mistral-7B-Instruct}}
\newcommand{\llmllama}{\textsc{\small Llama-3.1-8B-Instruct}}

\definecolor{tred}{RGB}{227, 11, 92}

\newcommand{\vlmqwen}{\textsc{\small Qwen3-VL-4B-Instruct}}
\newcommand{\vlmblip}{\textsc{\small InstructBLIP-7B}}
\newcommand{\vlmllava}{\textsc{\small LLaVA-1.5-7B}}

\title{One-Shot Calibration via Model Editing}
\author{Jiaxing Qiu \\
University of Virginia \\
\texttt{jq2uw@virginia.edu}
\And
Luke Hakso \\
University of Virginia \\
\And
Roxana Daneshjou \\
Stanford University \\
\AND
Ahmed Alaa \\
University of California, Berkeley \\
\And
Thomas Hartvigsen \\
University of Virginia \\
\texttt{hartvigsen@virginia.edu}
}

\iclrfinalcopy % Uncomment for camera-ready version, but NOT for submission. % Remember to switch lhead in iclr2027_conference.sty file!!!!!!
\begin{document}

\maketitle

\begin{abstract}

Model editing aims to correct errors in large, pretrained models by replacing model generations with user-provided responses. Existing editors assume a single target answer for each edit. However, many real tasks inherently entail uncertainty, resulting in multiple plausible answers with varying likelihoods in user feedback. Such feedback forms a target distribution to not only correct but also calibrate model generations.
We therefore propose SoftEdit, the first model editor to leverage uncertainty in user responses and perform one-shot calibration during editing. To generalize calibration to unseen samples, SoftEdit uses co-plausibility relationships between answers while preserving real distributional shifts, balancing the new distributional generality--locality tradeoff.
Across three language models, three vision-language models, and two large datasets, SoftEdit achieves state-of-the-art performance, ultimately showing that uncertainty in user responses enables generalizable calibration of pretrained models beyond single-answer corrections.\footnote{Our code and data are available at \url{https://github.com/JiaxingQiu/SoftEdit}}
\end{abstract}

\section{Introduction}

% Motivation 1
Large pretrained text generation models achieve state-of-the-art performance on many tasks, but still make errors and miscalibrated generations after deployment as data distributions, world knowledge, and user needs evolve over time~\citep{ma2025large}.
Correct and calibrated generations are important for faithful simulations, such as LLM-based survey simulations~\citep{hu2026simbench,cao-etal-2025-specializing,grief-albert-etal-2026-emulate}, and reliable decision support in high-stakes domains, such as medical language model question answering and differential diagnoses~\citep{singhal2025toward,mcduff2025towards}.
To address this gap, many recent methods have been developed to make targeted post hoc model edits.
% While there is a quickly-growing set of methods to correct model errors post hoc, they heavily rely on . %To correct a model's error for a given query, we need a target answer.
% 
However, they depend on settings where there is one correct answer, such as factual recall.
% However, in practice, a single target answer is commonly unavailable due to ambiguity in the data or variability in the data-collection process.
%
Many real tasks, in contrast, entail multiple plausible answers with varying levels of uncertainty.
For example, doctors often provide multiple plausible diagnoses, and different annotators often disagree on subjective tasks. %, clinicians typically provide multiple plausible diagnoses.
% 
% In another common example, distributions over multiple plausible answers can naturally arise from disagreement among multiple users. 
% This is the key:
% Such user feedback not only contains multiple plausible target answers to correct model mistakes, but also provides a natural informative target distribution to correct \textit{miscalibrated model generations}.
To quickly repair models on \textbf{inherently uncertain tasks}, we need efficient model updates that handle multiple plausible answers and uncertainty between them.

%\textcolor{blue}{(1) Do we want to address why it's importance to correct miscalibrated model generations in the introduction? (2) if so, where to put this paragraph?-----}
%Large pretrained models can overconfidently favor single responses, failing to capture uncertainty over multiple plausible answers~\citep{ma2025large, ni2026can}. Further, a globally-calibrated model can still be poorly calibrated in the local neighborhood of individual samples~\citep{luo2022local}. In practice, however, calibrating a model’s predictive distribution is important to faithfully capture the correct data generation process when simulating human responses~\citep{fleisig2024perspectivist, meister2025benchmarking, hu2026simbench}. It is also especially important in high-stakes domains such as medicine, where uncertainty over diagnostic predictions provides crucial information for reliable and trustworthy decision support~\citep{jung2025large, singhal2025toward}.  Therefore, beyond correcting errors, it is also important to correct miscalibrated generations after deployment.

\begin{figure}
    \centering
    \includegraphics[width=1\linewidth]{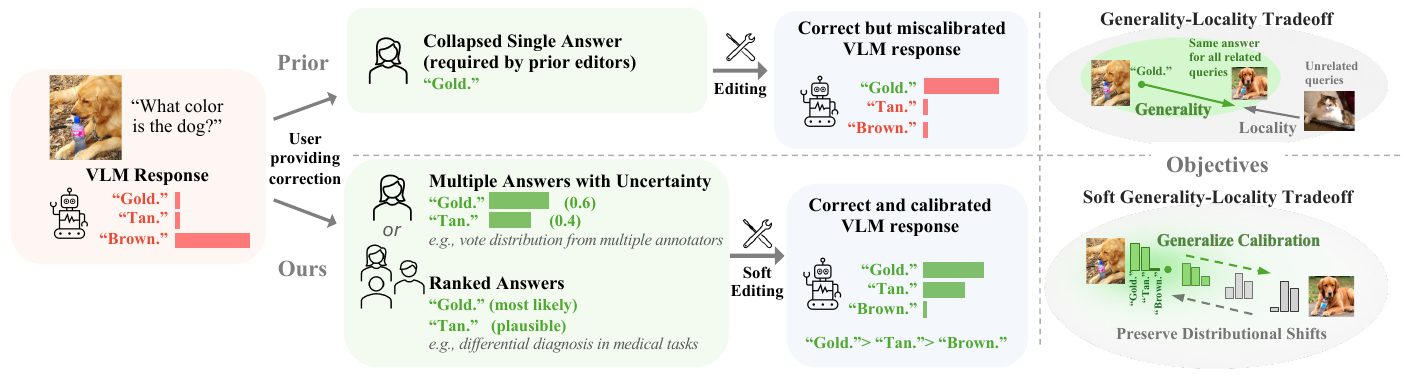}
    \caption{ Soft model editing edits inherently uncertain tasks and performs targeted calibration beyond single-answer correction, introducing a new distributional generality–locality tradeoff.}
    \label{fig:intro} 
    % \vspace{-6mm}
\end{figure}

% Gap
Model editing aims to efficiently make targeted, one-shot updates to large pretrained models using user feedback, while generalizing to unseen samples and preserving unrelated model behaviors~\citep{wang2024knowledge}.
Recent work has extended from editing factual knowledge in large language models (LLMs)~\citep{mitchell2021fast, meng2022locating, hartvigsen2023aging, gu2024model, zhang2025mc} to editing visually grounded information in vision-language models (VLMs)~\citep{cheng2023can, huang2024vlkeb, guo2025balancedit, qiu2026reasonedit}. 
However, as shown in Fig.~\ref{fig:intro}, existing model editors assume that each edit is associated with a single target answer,
and cannot directly leverage multi-answer feedback and its distributional information to calibrate generation beyond single-answer correction.
% When the target contains multiple plausible answers with varying uncertainty, no existing editors can directly incorporate this multi-answer feedback and leverage its distributional information to calibrate model generation beyond single-answer correction.
%
One way to adapt existing editors would be to collapse distributional feedback into a single target answer. However, this induces miscalibration because maximizing the likelihood of one answer suppresses other plausible answers. 
Even worse, generalizing this single-answer correction to unseen samples propagates the miscalibration.

% Our problem
We propose soft model editing: given a query, a model, and multiple plausible natural-language answers with uncertainty estimates for each, a model editor produces an updated model that (1) outputs an answer distribution calibrated with the given uncertainty estimates for each plausible answer, (2) generalizes the calibrated distribution to unseen semantically- and distributionally-similar queries, and (3) preserves distributional variations for unseen queries. 
% By calibrating the answer-level distribution and learning to generate each target answer in natural language, the edited model generates target plausible answers by construction.
% Intuitively, one-shot calibration corrects answer-specific over- and under-confidence, benefiting unseen samples with similar miscalibration.

% Challenges
Soft model editing poses new challenges. 
% (1)
The first challenge is a new soft generality--locality tradeoff: 
%the calibrated distribution should generalize to semantically and distributionally equivalent queries, but not to similar queries with genuine distributional shifts, despite similarity being unlabeled. 
matching the target distribution at the edited query does not ensure calibration on related unseen queries, while propagating the same distribution across its semantic neighborhood can overwrite meaningful distributional shifts (Fig.~\ref{fig:intro}).
% On the one hand, matching the target distribution at the edited query does not ensure that calibration generalizes to related unseen queries. On the other hand, semantically similar queries can have different plausible-answer distributions, so propagating the same distribution across the semantic neighborhood can overwrite meaningful distributional shifts.
%
% On the one hand, matching the target distribution at the edited query does not ensure that calibration generalizes to related unseen queries; by moving an edit near a decision boundary, small perturbations can induce unstable changes in the predicted distribution~\citep{he2018decision, pearce2021understanding}.
% On the other hand, semantically similar queries can differ in their distributions over plausible answers, so propagating the same distribution to the edited sample's semantic neighborhood can overwrite meaningful distributional shifts.
%
Therefore, there must be a balance between under- and over-generalization.
The second challenge is that plausible answers can span multiple tokens. Aligning first-token probabilities~\citep{lee2023can, ma2025large} calibrates answer selection but does not ensure complete answer generation. Thus, an editor must also preserve accurate generation.% after calibration.

%Reducing answer-level distribution alignment to single-entry-token alignment does not ensure correct continuation of the full answer and can degrade generation by perturbing the full-vocabulary distribution. Thus, an editor must still generate each answer accurately after editing.

% Our solution
We introduce SoftEdit, the first uncertainty-aware model editor, which calibrates and corrects models using one sample while generalizing the calibration to unseen samples. %targeting one-shot model calibration beyond single-answer correction. 
% (0) retrieve a local weight update.
To enable local calibration from a single query, we adopt a retrieval-based editing framework~\citep{zheng2023can, guo2025balancedit, qiu2026reasonedit}: for each edit, SoftEdit learns and caches a row-sparse weight perturbation alongside a retrieval key. Then if a new query falls close enough to a cached key, the corresponding perturbation is retrieved and temporarily added to the base model weight to generate the answer.
We optimize each weight perturbation with three objectives. 
% (1) l_gen: generalize calibration
(1) To generalize calibration beyond an edit, we construct a pairwise co-plausibility graph over the plausible answers and use it to regularize the local weight Jacobian at the edit. This encourages the probabilities of co-plausible answers to change coherently rather than independently as the query moves away from the edit.
% (2) l_loc: preserve genuine distribution shifts
(2) To prevent overwriting desirable distributional shifts, we regularize the weight perturbation with simple synthetic text-only in-context locality samples constructed from each edit’s target answers.
% (3) L_align: distribution alignment for multiple natural language answers. 
(3) To align the answer-level distribution while preserving text generation, we match the model's first-token probabilities to the plausible answers' while optimizing full-text generation of each answer. This encourages stochastic generation of each plausible answer at the target probability, while enabling fast deterministic readout of multiple answers with their estimated probabilities.
%by combining first-token probabilities with teacher-forced continuation validation, without Monte Carlo sampling.

% Results
In experiments with three LLMs and three VLMs on two large datasets with multi-answer distributions, SoftEdit achieves state-of-the-art performance in one-shot calibration as well as error correction for model generation. Compared with existing model editors adapted to the same objective, SoftEdit better generalizes calibration to unseen samples, preserves genuine distributional variation, and maintains higher computational efficiency. %\tom{Maybe add a point that one-hot editing is worse, suggesting that information's successfully extracted from soft labels?}

% contributions
In summary, our contributions are as follows:
\begin{enumerate}[topsep=0pt,itemsep=3pt,parsep=0pt]
    \item We introduce soft model editing, generalizing post-deployment model editing from single-answer correction to local calibration of large models via multi-answer human feedback.
    
    \item We introduce SoftEdit, a model editing method that performs local calibration while balancing the inherent tradeoff between generalizing and localizing distributional changes.
    
    \item Our extensive experiments across VLMs and LLMs show that SoftEdit achieves state-of-the-art performances with high computational efficiency.
\end{enumerate}

\section{Related Work}

% post training
\xhdr{Post-training Model Editing} 
Model editing enables efficient, targeted corrections to large pretrained models using individual human feedback, aiming to generalize these updates to unseen samples (generality) while preserving unrelated behaviors (locality).
Balancing generality and locality becomes harder as edits accumulate and interfere sequentially. 
Early editors rely on fine-tuning, which can cause model collapse~\citep{sinitsin2020editable}, while later methods such as MEND~\citep{mitchell2021fast}, ROME~\citep{meng2022locating}, and AlphaEdit~\citep{fang2025alphaedit} use more targeted weight updates to better preserve model behaviors.
Retrieval-based editors such as GRACE~\citep{hartvigsen2023aging} and IKE~\citep{zheng2023can} instead perform edits without modifying pretrained model weights, often achieving strong locality but conservative generalization.
Some editors leverage model-induced distributions to regularize editing: ICE~\citep{qi2024context} regularizes single-fact editing using contextual distributions; APP/PEAK~\citep{ma2024neighboring} preserves existing plausible answers through probability margins, but neither targets human-provided answer rankings or distributions.
Recent work has extended model editing to VLMs~\citep{cheng2023can, huang2024vlkeb, zhang2025mc}, aiming to edit visually grounded information while balancing multimodal generality and locality~\citep{zeng2025visual, guo2025balancedit, qiu2026reasonedit}.
However, existing model editors require users to provide a single ground-truth answer for each model error. They neither account for uncertainty over multiple plausible answers in realistic human feedback nor leverage this distributional information to calibrate model generation beyond single-answer correction. 
% Collapsing to a single answer can further propagate miscalibration to related inputs.  
Our work addresses this gap by allowing for uncertainty over multiple plausible answers and targets distributional calibration beyond single-answer correction.

\xhdr{Post-training Model Calibration}
Large pretrained models can be miscalibrated at inference, often overconfidently favoring a single response and failing to capture realistic uncertainty over multiple plausible answers~\citep{lee2023can, cheng2024soft, ma2025large, inoshita2026llms, jiang2026capturing, ni2026can, wu2026shala}. Calibrating model uncertainty helps capture real-world ambiguity~\citep{hullermeier2021aleatoric, fleisig2024perspectivist}, reduce systematic biases in generated outputs~\citep{smith2025calibrating}, and support high-stakes domains, where uncertainty over predictions provides crucial information for decision-support~\citep{mimori2021diagnostic, singhal2025toward}. 
Traditional calibration methods apply global post-hoc transformations using a held-out dataset of single-labeled samples drawn from the target distribution to improve global calibration~\citep{platt1999probabilistic, zadrozny2001obtaining, guo2017calibration, zhou2024batch}. 
However, global calibration can hide local miscalibration for individual samples~\citep{luo2022local, joy2023sample, xie2024calibrating}.
More flexible approaches allow adaptive temperature scaling~\citep{joy2023sample,xie2024calibrating}, but they still require curated calibration datasets and cannot perform targeted updates. They also assume a shared, coherent miscalibration pattern, which may not hold when individual edits require heterogeneous calibrations across the full vocabulary.
% More flexible approaches allow adaptive temperature scaling~\citep{joy2023sample, xie2024calibrating}, but they also require a curated held-out calibration dataset with a coherent target distribution and cannot use diverse individual human corrections to update a pretrained model.
%
Our work targets these gaps by introducing generalizable one-shot calibration using individual human corrections in free-from text.

\section{One-Shot Calibration via Model Editing}

\subsection{Problem Definition}

Let $e=(x,\mathcal{Y}_{K},\mathbf{p}_{K})$ denote an edit, where $x$ is an input query to a pretrained model and $\mathcal{Y}_{K}=(y_1,\ldots,y_K)$ is the corresponding set of $K$ plausible human-provided answers.
The $K$ answers follow an answer-level target distribution $\mathbf{p}_{K}(\cdot\mid x)=(p_1,\ldots,p_K)$ estimated from human feedback, where $p_k$ denotes the probability that a human provides $y_k$ as the answer.
For a language model, $x$ is a text question.
For a vision-language model, $x$ pairs an image with a text question.

We aim to develop a model editor $g$ that takes in an edit $e$ and a pretrained model $f$, and outputs a locally calibrated model $f_{\text{new}}=g(e,f)$. 
Let $\widehat{\mathbf{p}}_{K}(\cdot\mid x;f_{\text{new}})$ denote the model's distribution over the $K$ human-provided answers under stochastic generation. 
After editing, $f_\text{new}$ should 1) align $\widehat{\mathbf{p}}_{K}$ with the human target distribution $\mathbf{p}_{K}$, such that each answer $y_k$ is generated in full text with target probability $p_k$, 2) generalize the calibrated distribution to unseen queries that are semantically and distributionally similar to the edit, and 3) preserve real distributional shifts.
% \tom{These criteria are a little informal. Is it possible to replace any of this language to be precise with what "align," "generalize," and "preserve" mean? We have all the notation now so could use it}.
%
Note that conventional model editing is a special case of our one-shot calibration problem: when the target distribution assigns probability one to a single ground-truth answer, one-shot local calibration reduces to standard single-answer correction.

\subsection{SoftEdit}

% ---- overview ----
We propose SoftEdit, the first model editor targeting one-shot model calibration beyond single-answer correction.
To enable localized editing, SoftEdit adopts a retrieval-based framework~\citep{zheng2023can, guo2025balancedit, qiu2026reasonedit} and stores a local weight perturbation matrix $\Delta W_e$ for each edit.
Rather than storing a fixed answer or target distribution, $\Delta W_e$ captures how the target distribution should vary across the edit's semantic neighborhood.
At inference, when a new query falls within the semantic neighborhood of an edit, SoftEdit retrieves $\Delta W_e$ and temporarily applies to original model weight to generate an updated answer using a calibrated distribution for this query.

\subsubsection{Learning $\Delta W_e$: Soft Distributional Generality--Locality Tradeoff}
\label{sec:tradeoff}

For each edit $e=(x,\mathcal{Y}_{K},\mathbf{p}_{K})$, we learn a local weight perturbation $\Delta W_e$ with the objective:
\[
\mathcal{L} = \mathcal{L}_{\text{align}} + \lambda_{\text{gen}}\,( \mathcal{L}_{\text{gen}} + \lambda_{\text{loc}}\, \mathcal{L}_{\text{loc}}),
\]
where $\mathcal{L}_{\text{align}}$ aligns the model's answer-level distribution with the target distribution $\mathbf{p}_{K}$ for the edit $e$. $\mathcal{L}_{\text{gen}}$ encourages this calibration to generalize to semantically similar unseen queries that share the target distribution $\mathbf{p}_{K}$, while $\mathcal{L}_{\text{loc}}$ preserves real distributional shifts.  $\lambda_{\text{gen}}$ and $\lambda_{\text{loc}}$ balance the loss terms using standard loss-scale normalization~\citep{kendall2018multi}.

We only edit the final vocabulary projection layer, where $W\in\mathbb{R}^{|\mathcal{V}|\times d}$ maps the last hidden state $h\in\mathbb{R}^{d}$ to logits $z\in\mathbb{R}^{|\mathcal{V}|}$. 
% Note that $h$ remains fixed for a given input because all preceding layers are frozen.
For edit $e$, let $\mathcal{S}_e$ be the set of unique token indices that appear in the $K$ target answers, including the end-of-sequence token. SoftEdit learns a row-sparse perturbation $\Delta W_e\in\mathbb{R}^{|\mathcal{S}_e|\times d}$ only for these rows by optimizing $\mathcal{L}$, and stores it with their token indices and the retrieval key. 
Let $\oplus_{\mathcal{S}_e}$ denote $\mathcal{S}_e$-indexed addition. The updated layer weight is $W' = W \oplus_{\mathcal{S}_e}\Delta W_e$. 
%
% benefits of editing the final vocabulary projection layer
Editing the final projection weight keeps $\Delta W_e$ compact and fast to optimize: only $W$'s rows indexed by $\mathcal{S}_e$ are modified, and $|\mathcal{S}_e| \ll |\mathcal{V}|$. The hidden states required by $\mathcal{L}$ can therefore be computed once and reused across all optimization steps, avoiding repeated forward passes.

\begin{figure}
    \centering
    \includegraphics[width=1\linewidth]{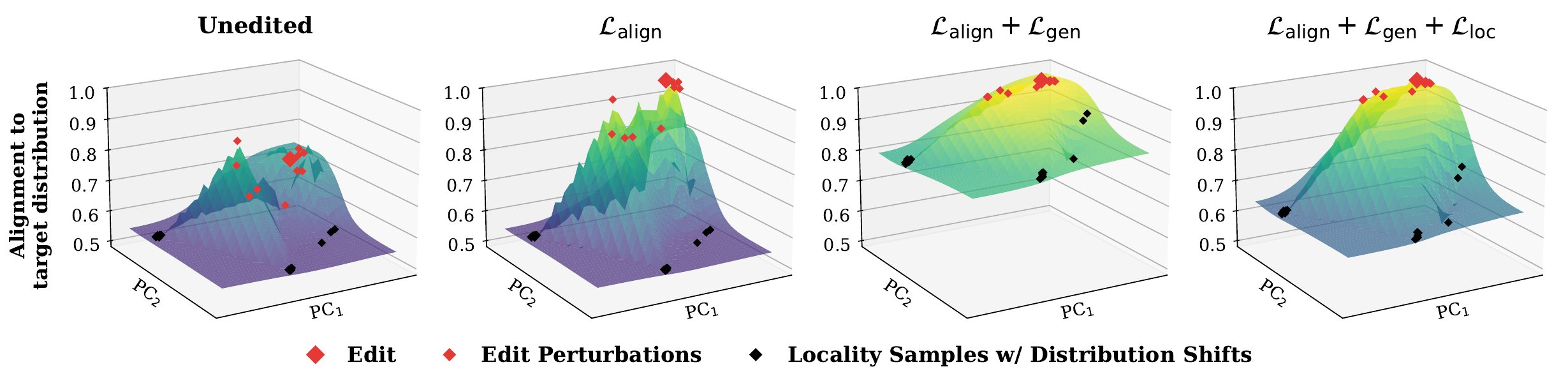}
    \caption{\small Visualization of changes in the local distribution field of a real edit from VQAv2 as SoftEdit objectives are progressively applied to edit \vlmllava. The \(xy\)-plane is a PCA projection of neighbors’ last-token hidden states; the $z$-axis shows the distributional alignment towards the edit's target distribution.  SoftEdit ($\mathcal{L}_{\text{align}} + \mathcal{L}_{\text{gen}} + \mathcal{L}_{\text{loc}}$) calibrates the edit ($\text{alignment}\!\rightarrow\!1$ at the large red point), generalizes calibration to distributionally equivalent neighbors ($\text{alignment}\!\rightarrow\!1$ at other red points) without overwriting distributional shifts in locality samples (low alignment to edit target at black points).
    }
    \vspace{-2mm}
    \label{fig:tradeoff}
\end{figure}

% ===== (1) L_align =====
\xhdr{$\mathcal{L}_{\text{align}}$: Local targeted calibration}
% -- goal + idea --
To align the model distribution $\widehat{\mathbf{p}}_K$ with the human-provided distribution $\mathbf{p}_K$, while preserving full-text generation, we supervise both answer selection and answer completion.
Each answer $y_k$ consists of a token sequence $(y_{k,1}, \ldots, y_{k,L_k})$ of length $L_k$, where $y_{k,1}$ is its first token and $y_{k,L_k}$ is the appended end-of-sentence token. 
We use its first-token probability to represent how likely the model is to select that answer, and optimize a continuation term---the length-normalized negative log-likelihood of the remaining tokens---to ensure that the complete answer is generated correctly.
Let $h_0$ denote the edit query $x$'s final hidden state before the final projection layer. 
The alignment loss is given by:
\[
\mathcal{L}_{\text{align}}
= \underbrace{D_{\mathrm{KL}}\!\left(\mathbf{p}_K \,\big\|\, \widehat{\mathbf{p}}_K(h_0)\right)}_{\text{first-token alignment}}
+ \underbrace{\frac{1}{K}\sum_{k=1}^{K}\left[-\frac{1}{L_k - 1}\sum_{t=2}^{L_k}\log p_\theta\!\left(y_{k,t} \mid x,\, y_{k,<t}\right)\right]}_{\text{full-answer continuation}},
\]
where $p_\theta(y_{k,t} \mid x, y_{k,<t})$ is the model's probability of generating the $t$-th token conditioned on all preceding tokens. 
% Since the relative answer probabilities are already captured by the first-token distribution alignment, we average the continuation loss uniformly across the $K$ answers to avoid weaker full-text generation of lower-probability answers.
%
This loss factorizes full-sequence answer probability calibration into first-token probability alignment ($p_\theta(y_{k,1}\mid x)\approx p_k$) and full-answer continuation (encouraging the remaining conditional token probabilities toward $1$), so that the probability of generating each complete answer approaches its target.
At inference, this enables (1) fast deterministic readout of multiple complete answers with their estimated probabilities, (2) exact autoregressive full-sequence probabilities for specified answers, and (3) stochastic sampling of complete answers according to the calibrated distribution. We discuss these inference approaches and their empirical validation in Appendix~\ref{app:calib_def}.
As shown by a real edit example in Fig.~\ref{fig:tradeoff}, $\mathcal{L}_{\text{align}}$ calibrates the model to the target distribution at the edit. However, $\mathcal{L}_{\text{align}}$ alone does not ensure that this calibration generalizes to semantically and distributionally equivalent samples. High-entropy target distributions can also place edits near decision boundaries, where small perturbations may induce unstable changes in the predicted distribution~\citep{he2018decision, pearce2021understanding}.

% ===== (2) L_gen =====
\xhdr{$\mathcal{L}_{\text{gen}}$: Generalizing calibration}
To generalize calibration beyond the single edit query $x$, we leverage the co-plausibility relationships between answers. Specifically, we construct a pairwise co-plausibility graph over the $K$ answers and apply graph-Laplacian regularization~\citep{belkin2006manifold} to the Jacobian of their edited first-token probabilities with respect to the hidden state $h$, evaluated at $h_0$.
The edge weight between answers $y_i$ and $y_j$ is $A_{ij}=p_i\,p_j$ ($i\neq j$, $A_{ii}=0$), so co-plausible answers with higher probabilities are more strongly connected.
Letting $J_i=\partial\widehat{p}_i/\partial h\big|_{h_0}$ be the gradient of the edited readout for answer $y_i$, the generalization loss is
\[
\mathcal{L}_{\text{gen}} = \sum_{i,j} A_{ij}\,\|J_i - J_j\|_2^2.
\]
This graph-Laplacian regularization encourages co-plausible answers' probabilities to change coherently as the input moves away from the edit, helping preserve the target answers' relative ordering and distribution in a neighborhood of $h_0$. 
% We adopt probability-weighted edges as a useful and overall optimal design choice that emphasizes relationships between more plausible answers, while alternative regularization formulations can also encourage coherent changes in answer probabilities, as detailed in Appendix~\ref{app:additional_ablation}.
%
As shown in Fig.~\ref{fig:tradeoff}, adding $\mathcal{L}_{\text{gen}}$ generalizes the calibrated distribution to samples that are semantically and distributionally equivalent to the edit, however, imposing the same distribution across the entire semantic neighborhood can erase genuine distributional shifts in semantically similar but distributionally different samples.

\begin{figure}
    \centering
    \includegraphics[width=1\linewidth]{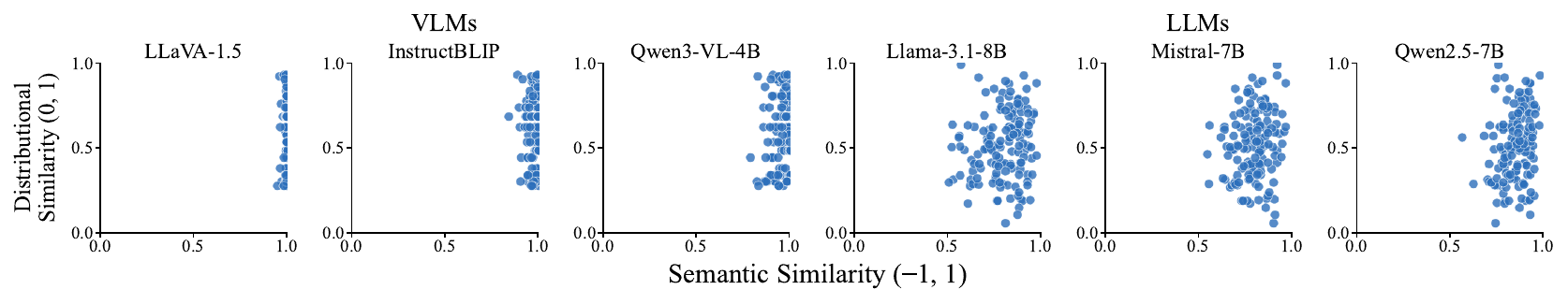}
    \caption{Semantically similar queries can have largely different target distributions, in VLMs (on VQAv2) and LLMs (on ProtoQA). Each point is a pair of semantically related queries, with cosine similarity of their last-layer hidden states (x-axis) and answer distribution alignment ($1-\mathrm{JS}$, y-axis).}
    \label{fig:motivation_sims}
    \vspace{-5mm}
\end{figure}

% ===== (3) L_loc =====
\xhdr{$\mathcal{L}_{\text{loc}}$: Preserving distributional shifts}
Within the semantic neighborhood of an edit $e$, unseen queries may share high semantic similarity due to closely related context or content, but semantic similarity does not imply distributional similarity: small differences in answer-relevant details can substantially shift the probabilities of the target answers, as shown in Fig.~\ref{fig:motivation_sims}.
To prevent SoftEdit from overwriting distributional shifts, we regularize the learning of $\Delta W_e$ using synthetic text-only locality samples constructed from each edit's target answers through a very simple procedure.

Specifically, for each answer $y_k$, we synthesize an unambiguous declarative prompt $x_k^{\text{loc}}$ by prepending ``The answer to [question $x$] is $y_k$.'' to the original question $x$, and pairing it with a one-hot target distribution where we assign probability 1 to $y_k$. Each locality sample $(x_k^{\text{loc}}, y_k; (p_k=1))$ therefore represents a semantically similar query where one target answer should clearly dominate. Optimizing distributional alignment on these locality samples encourages the edited model to preserve genuine distributional shifts.
We use the same first-token alignment and continuation structure as $\mathcal{L}_{\text{align}}$, applied to each locality sample:
\[
\mathcal{L}_{\text{loc}} = \frac{1}{K}\sum_{k=1}^{K} \left[
  \underbrace{D_{\mathrm{KL}}\!\left(\mathbf{p}_k^{\text{loc}} \,\big\|\, \widehat{\mathbf{p}}_K(h_k^{\text{loc}})\right)}_{\text{first-token alignment}}
  + \underbrace{\frac{1}{K}\sum_{j=1}^{K}\left[-\frac{1}{L_j - 1}\sum_{t=2}^{L_j}\log p_\theta\!\left(y_{j,t} \mid x_k^{\text{loc}},\, y_{j,<t}\right)\right]}_{\text{full-answer continuation}}
\right],
\]
where $\mathbf{p}_k^{\text{loc}}$ is the one-hot target on $y_k$ and $h_k^{\text{loc}}$ is the hidden state at $x_k^{\text{loc}}$.
Because the anchors are purely textual, this approach applies directly to not only LLMs but also VLMs without requiring answer-specific counterfactual images. As shown in Fig.~\ref{fig:tradeoff}, by incorporating both $\mathcal{L}_{\text{gen}}$ and $\mathcal{L}_{\text{loc}}$, SoftEdit generalizes the calibrated distribution while preserving genuine distributional shifts. 
More examples are provided in Appendix Fig.~\ref{fig:local_field_vlms}.

% ===== (4) automatic lambda_loc and lambda_gen =====
\xhdr{Balancing weights}
$\lambda_{\text{loc}}$ and $\lambda_{\text{gen}}$ are set automatically via loss-scale normalization~\citep{kendall2018multi}, computed once on the first 100 accumulated edits using the unedited model.
Specifically, $\lambda_{\text{loc}}=\bar{\mathcal{L}}_{\text{gen}}/\bar{\mathcal{L}}_{\text{loc}}$ equalizes the scales of the two regularization terms, while $\lambda_{\text{gen}}$ scales the sum of them relative to $\mathcal{L}_{\text{align}}$ to prevent negligible regularization.
Details are provided in Appendix~\ref{app:loss_balancing}.

\vspace{-5mm}

\subsubsection{Storing and Retrieving $\Delta W_e$}

Weight-update editors~\citep{mitchell2021fast, meng2022locating, fang2025alphaedit} capture nuanced output distribution changes as inputs change, but shared parameter updates cause interference and catastrophic forgetting as edits accumulate.
Retrieval-based editors~\citep{zheng2023can, hartvigsen2023aging, qiu2026reasonedit} improve local preservation by restricting each edit to its semantic neighborhood, but store a fixed answer or target distribution that cannot adapt to distributional shifts within that neighborhood.
% Therefore, SoftEdit stores a local weight perturbation $\Delta W_e$ for each edit. Retrieval determines whether a query falls within the edit's semantic neighborhood, while $\Delta W_e$ maps the query to a calibrated distribution that preserves distributional shifts as answer-relevant details change.
SoftEdit leverages both strategies by storing the learned $\Delta W_e$ for each edit and retrieving it to provide an updated distributional mapping for its semantically similar queries.

\xhdr{Storing $\Delta W_e$}
After learning $\Delta W_e$, we store it in a codebook together with a retrieval key constructed from the embedding representation $\mathcal{E}(x)$ of the edit query $x$. At inference, this key is used to retrieve the relevant $\Delta W_e$ when a new query's embedding falls within the edit's semantic neighborhood.
We follow prior retrieval-based editors for constructing the retrieval key. For VLMs,
\begin{wrapfigure}[25]{r}{0.38\textwidth}
\begin{minipage}{0.38\textwidth}

\hrule
\captionsetup{
    font=small,
    labelfont=bf,
    labelsep=space,
    justification=raggedright,
    singlelinecheck=false,
    aboveskip=0pt,
    belowskip=1pt
}
\captionof{algorithm}{SoftEdit}
\label{alg:softedit}
\hrule

\small

% ==================== Editing ====================
\begin{algorithmic}
\setlength{\itemsep}{0pt}

    \STATE \hfill === \textit{Editing} === \hfill
    \STATE {\bfseries Input:} $f(\cdot;W)$, pretrained model
    \STATE {\bfseries Input:} $e=(x,\mathcal{Y}_K,\mathbf{p}_K)$, edit
    \STATE {\bfseries Input:} $\mathcal{C}$, codebook

    \STATE \textcolor{blue}{\textit{Step 1: Learn $\Delta W_e$}}
    \STATE Learn row-sparse $\Delta W_e$ by minimizing
    \STATE \hspace{\algorithmicindent}
    \footnotesize $\mathcal{L}_{\mathrm{align}}
    +\lambda_{\mathrm{gen}}(\mathcal{L}_{\mathrm{gen}}
    +\lambda_{\mathrm{loc}}\mathcal{L}_{\mathrm{loc}})$

    \STATE \textcolor{blue}{\textit{Step 2: Store edit}}
    \STATE $\mathrm{K}_e \gets \mathcal{E}(x)$
    \STATE $\mathcal{C} \gets
    \mathcal{C}\cup\{(\mathrm{K}_e,\Delta W_e)\}$
    \STATE {\bfseries Output:} updated codebook $\mathcal{C}$

\end{algorithmic}

% \vspace{1pt}
\hrule
% \vspace{1pt}

% ==================== Inference ====================
\begin{algorithmic}
\setlength{\itemsep}{0pt}

    \STATE \hfill === \textit{Inference} === \hfill
    \STATE {\bfseries Input:} $x'$, new query
    % \STATE {\bfseries Input:} $\mathcal{C}$, codebook

    \STATE \textcolor{blue}{\textit{Step 1: Retrieve edit}}
    \STATE $\mathrm{K}' \gets \mathcal{E}(x')$
    \STATE Calculate $\|\mathrm{K}'-\mathrm{K}_e\|_2$
    \STATE Retrieve nearest
    $(\mathrm{K}_e,\Delta W_e) \in \mathcal{C}$

    \STATE \textcolor{blue}{\textit{Step 2: Apply or reject edit}}
    {\footnotesize
    \IF{$\|\mathrm{K}'-\mathrm{K}_e\|_2 \leq \gamma$}
        \STATE $f_{\text{new}}(\cdot)
        \gets f(\cdot;W+\Delta W_e)$
    \ELSE
        \STATE $f_{\text{new}}(\cdot)
        \gets f(\cdot;W)$
    \ENDIF
    }

    \STATE {\bfseries Output:} $y' = f_{\text{new}}(x')$

\end{algorithmic}
\hrule
\end{minipage}
\end{wrapfigure}
we adopt the dual embedding for multimodal queries from ReasonEdit~\citep{qiu2026reasonedit}, combining the vision encoder's embedding of the image and a sentence encoder's embedding of the question using topology-balanced multimodal weighting. For LLMs, following IKE~\citep{zheng2023can, wang2024easyedit, guo2025balancedit}, we embed questions with sentence encoder \texttt{\small all-MiniLM-L6-v2}. 

\xhdr{Gated retrieval of $\Delta W_e$ at inference}
At inference, a new query is encoded into the same embedding space and matched to the nearest codebook key by $L_2$ distance.
Following prior retrieval-based editors~\citep{qiu2026reasonedit}, we use the distances among codebook keys to estimate how far semantically unrelated queries typically lie in the embedding space. Retrieval is rejected when the distance from a new query to its nearest key exceeds a percentile $\gamma$ of this distance distribution.
We automatically determine this rejection percentile using a small batch of accumulated edits: we generate semantics-preserving augmentations of these edits and sweep candidate percentiles of the key-distance distribution to choose the lowest percentile that preserves accurate retrieval of each edit for its augmentations. To further distinguish subtle semantic differences between queries with similar embeddings, we use a lightweight LLM judge (\texttt{\small Qwen2.5-1.5B-Instruct}) to verify whether a new query is a paraphrase of its matched edit's query. More details are provided in Appendix~\ref{app:reject_threshold}.

\section{Experiments}

\subsection{Evaluation Metrics}

To evaluate distributional calibration for an input query $x$, we compare the edited model's answer-level distribution $\widehat{\mathbf{p}}_K(\cdot\mid x;\,f_{\mathrm{new}})$ with the target distribution $\mathbf{p}_K$.
A standard Monte Carlo estimate samples many full generations and computes answer frequencies~\citep{sorensen2022information, lee2023can}, but is prohibitively expensive at scale.
We therefore introduce a fast answer-level probability estimator using the model's first-token probability $p_\theta(t_{k,1}\mid x)$ combined with a binary continuation gate $c_k(x)\!\in\!\{0,1\}$, which verifies whether the model correctly completes the full answer via teacher forcing:
$\widehat{p}(y_k\mid x)
=
p_\theta(t_{k,1}\mid x)\cdot c_k(x).$
Full details are provided in Appendix~\ref{app:gated_readout}.
We then measure a gated Jensen--Shannon distance:
$\mathrm{JS_{gate}}
=
\sqrt{\mathrm{JSD}\!\bigl(\widehat{\mathbf{p}}_K\|\mathbf{p}_K\bigr)}
\;\in\;[0,1].$
Lower values indicate closer distributional alignment.
We validate the answer probabilities estimated by this fast readout against autoregressive full-sequence probabilities and the empirical answer frequencies from Monte Carlo sampling for SoftEdit in Appendix~\ref{app:calib_def}. 
% Additionally, we detail three approaches to calibrated answer generation enabled by SoftEdit: (1) fast deterministic readout of multiple answers with their probabilities in \ref{app:calib_def_readout}, (2) estimating autoregressive full-sequence probabilities given user-specified answers in \ref{app:calib_def_full_seq}, (3) stochastic sampling of complete answers in \ref{app:calib_def_MC}, as well as how our
% one-shot local calibration relates to conventional calibration
% in \ref{app:calib_def_compare}.

%
Based on the problem definition, we evaluate an editor's one-shot calibration performance on three properties: \textit{alignment, soft generality, and soft locality}, as well as standard model editing performance: \textit{reliability, generality, and locality}.
Unseen samples are used to evaluate generality and locality metrics, and construction details are provided in Appendix~\ref{app:eval_metrics}.

\xhdr{Alignment (Align)}
The edited model should match the human-provided target distribution $\mathbf{p}_{K}$ on each edit input.
$\small
\mathrm{Align}
= \mathbb{E}_{e\in D_{\mathrm{edit}}}
\left[
1-\mathrm{JS_{gate}}\!\left(
\mathbf{p}_{K},\,
\widehat{\mathbf{p}}_{K}(\cdot\mid x;\,f_{\mathrm{new}})
\right)
\right].$

\xhdr{Soft Generality (SGen)}
The edited model should generalize the calibrated distribution to unseen semantically and distributionally similar queries.
Let $R(x)$ denote variants of $x$ that preserve the target distribution $\mathbf{p}_{K}$.
Then,
$\small
\mathrm{SGen}
= \mathbb{E}_{e\in D_{\mathrm{edit}}}
\left[
1-\mathrm{JS_{gate}}\!\left(
\mathbf{p}_{K},\,
\widehat{\mathbf{p}}_{K}(\cdot\mid R(x);\,f_{\mathrm{new}})
\right)
\right].$
We evaluate text variants (SGen-T) using question rephrasing and, for VLMs, image variants (SGen-I) using generated semantically and distributionally similar images following prior studies (Appendix~\ref{app:eval_metrics}).

\xhdr{Soft Locality (SLoc)}
The edited model should preserve genuine distribution shifts in unseen samples that are semantically similar to an edit query $x$, rather than overwriting them with $\mathbf{p}_{K}$.
Let $R_{\mathrm{loc}}(x)$ denote related inputs whose target distribution $\mathbf{p}_{K}^{\mathrm{loc}}\neq\mathbf{p}_{K}$.
$\small
\mathrm{SLoc}
= \mathbb{E}_{e\in D_{\mathrm{edit}}}
\left[
1-\mathrm{JS_{gate}}\!\left(
\mathbf{p}_{K}^{\mathrm{loc}},\,
\widehat{\mathbf{p}}_{K}(\cdot\mid R_{\mathrm{loc}}(x);\,f_{\mathrm{new}})
\right)
\right].$ 
Note that higher SLoc indicates better preservation of distributional shifts, and improvements over the unedited baseline are meaningful without reaching 1, because one-shot calibration does not directly supervise the unseen shifted distributions.

% \xhdr{(4) Reliability (Acc)}
% The edited model should generate a correct answer from the target plausible answers set $\mathcal{Y}_{K}$ on the edit input.
% $\small
% \mathrm{Acc}
% = \mathbb{E}_{e\in D_{\mathrm{edit}}}
% \mathbf{1}\!\left\{
% f_{\mathrm{new}}(x)\in\mathcal{Y}_{K}
% \right\}.$

% \xhdr{(5) Generality (Gen)}
% The edited model should generate a correct answer on unseen semantically equivalent inputs.
% $\small
% \mathrm{Gen}
% = \mathbb{E}_{e\in D_{\mathrm{edit}}}
% \mathbf{1}\!\left\{
% f_{\mathrm{new}}(R(x))\in\mathcal{Y}_{K}
% \right\}.$

% \xhdr{(6) Locality (Loc)}
% The edited model should preserve the original prediction on unrelated samples $D_u$.
% $\small
% \mathrm{Loc}
% = \mathbb{E}_{x\in D_u}
% \mathbf{1}\!\left\{
% f_{\mathrm{new}}(x)=f(x)
% \right\}.$

\xhdr{Model Editing Metrics}
Given model editing is a special case of our setup, we also measure traditional model editing metrics:
Reliability measures whether the edited model generates a correct answer from $\mathcal{Y}_{K}$ on the edit input:
$\small \mathrm{Acc} = \mathbb{E}_{e\in D_{\mathrm{edit}}} \mathbf{1}\{f_{\mathrm{new}}(x)\in\mathcal{Y}_{K}\}.$
Generality measures correctness on unseen semantically equivalent inputs:
$\small \mathbb{E}_{e\in D_{\mathrm{edit}}} \mathbf{1}\{f_{\mathrm{new}}(R(x))\in\mathcal{Y}_{K}\}.$
Locality measures whether the original prediction is preserved on unrelated samples $D_u$:
$\small \mathbb{E}_{x\in D_u} \mathbf{1}\{f_{\mathrm{new}}(x)=f(x)\}.$
% Details are provided in Appendix~\ref{app:eval_metrics}

\vspace{-2mm}
\subsection{Experimental Setup}

% \vspace{-2mm}
\xhdr{Datasets}
We evaluate editors on two multi-answer question--answering benchmarks with multi-annotator vote distributions.
(1) VQAv2~\citep{goyal2017making} contains visual questions answered by 10 annotators. We use the vote distribution over free-form answers for each sample and retain 19{,}186 samples for VLM editing.
(2) ProtoQA~\citep{boratko2020protoqa} contains commonsense questions with free-form answers clustered into canonical groups. We use vote distribution for each edit and retain 9{,}762 samples for LLM editing.
Using a pretrained model's predictions on each dataset, we construct the edit set (error set) \(D_{\mathrm{edit}}\) from incorrectly answered samples \((f(x)\notin\mathcal{Y}_K)\), and unrelated set \(D_u\) from correctly-answered ones \((f(x)\in\mathcal{Y}_K)\), following prior work~\citep{qiu2026reasonedit}. Data processing details are provided in Appendix~\ref{app:experiment_dataset}.

\xhdr{Edited Models}
We edit three popular VLMs: \vlmllava~\citep{liu2024improved}, \vlmblip~\citep{dai2023instructblip}, and \vlmqwen~\citep{yang2025qwen3} and three popular LLMs: \llmllama~\citep{grattafiori2024llama}, \llmmistral~\citep{jiang2023mistral7b}, and \llmqwen~\citep{qwen2025qwen25technicalreport}.
Details on these models are provided in Appendix~\ref{app:experiment_models}.

\xhdr{Baseline Editors}
We compare SoftEdit with five VLM editors and five LLM editors adapted for one-shot calibration (denoted [editor]-soft).
For weight or activation-updating editors: FT (Fine-tuning), MEND~\citep{mitchell2021fast}, AlphaEdit~\citep{fang2025alphaedit}, GRACE~\citep{hartvigsen2023aging}, and BalancEdit~\citep{guo2025balancedit}, we replace their single-answer cross-entropy loss with \(\mathcal{L}_{\text{align}}\).
For retrieval-based methods, ReasonEdit~\citep{qiu2026reasonedit} and IKE~\citep{zheng2023can}, we store and retrieve the ranked list of plausible answers.
Editor details are provided in Appendix~\ref{app:experiment_baselines}.

\subsection{Main Results}

\begin{table*}[t]
\centering

% (1) VLM Editing on VQAv2 (19,186 samples)
\resizebox{\textwidth}{!}{%
\setlength{\tabcolsep}{3pt}%
\begin{NiceTabular}{l>{\centering\arraybackslash }b{2.0em}>{\centering\arraybackslash }b{2.0em}>{\centering\arraybackslash }b{2.0em}>{\centering\arraybackslash }b{2.0em}>{\centering\arraybackslash }b{2.0em}>{\centering\arraybackslash }b{2.0em}>{\centering\arraybackslash }b{2.0em}>{\centering\arraybackslash }b{2.0em}>{\centering\arraybackslash }b{2.0em}>{\centering\arraybackslash }b{2.0em}>{\centering\arraybackslash }b{2.0em}>{\centering\arraybackslash }b{2.0em}>{\centering\arraybackslash }b{2.0em}>{\centering\arraybackslash }b{2.0em}>{\centering\arraybackslash }b{2.0em}>{\centering\arraybackslash }b{2.0em}>{\centering\arraybackslash }b{2.0em}>{\centering\arraybackslash }b{2.0em}>{\centering\arraybackslash }b{2.0em}>{\centering\arraybackslash }b{2.0em}>{\centering\arraybackslash }b{2.0em}>{\centering\arraybackslash }b{2.0em}>{\centering\arraybackslash }b{2.0em}>{\centering\arraybackslash }b{2.0em}}
\CodeBefore
\begin{tikzpicture}
  \fill[orange!6] (2-|2) rectangle (15-|6) ;
  \fill[orange!6] (2-|10) rectangle (15-|14) ;
  \fill[orange!6] (2-|18) rectangle (15-|22) ;
\end{tikzpicture}
\Body

\toprule
 \Block{2-1}{\textbf{Editor}} & \multicolumn{8}{c}{\textbf{LLaVA-1.5-7B (4,436 errors)}} & \multicolumn{8}{c}{\textbf{InstructBLIP-7B (4,625 errors)}} & \multicolumn{8}{c}{\textbf{Qwen3-4B (4,157 errors)}} \\
\cmidrule(lr){2-5} \cmidrule(lr){6-9} \cmidrule(lr){10-13} \cmidrule(lr){14-17} \cmidrule(lr){18-21} \cmidrule(lr){22-25}
 & \footnotesize Align & \footnotesize SGen-I & \footnotesize SGen-T & \footnotesize SLoc & \footnotesize Acc & \footnotesize Gen-I & \footnotesize Gen-T & \footnotesize Loc & \footnotesize Align & \footnotesize SGen-I & \footnotesize SGen-T & \footnotesize SLoc & \footnotesize Acc & \footnotesize Gen-I & \footnotesize Gen-T & \footnotesize Loc & \footnotesize Align & \footnotesize SGen-I & \footnotesize SGen-T & \footnotesize SLoc & \footnotesize Acc & \footnotesize Gen-I & \footnotesize Gen-T & \footnotesize Loc \\
\midrule
Unedited & 0.08 & 0.16 & 0.08 & 0.27 & 0.02 & 0.38 & 0.12 & 1.00 & 0.17 & 0.23 & 0.17 & 0.31 & 0.02 & 0.35 & 0.12 & 1.00 & 0.11 & 0.22 & 0.12 & 0.51 & 0.01 & 0.39 & 0.11 & 1.00 \\
\midrule
FT & 0.52 & 0.47 & 0.48 & 0.40 & 0.96 & 0.86 & 0.87 & 0.30 & 0.53 & 0.48 & 0.50 & 0.45 & 0.95 & 0.87 & 0.90 & 0.47 & 0.53 & 0.45 & 0.49 & 0.42 & 0.96 & 0.82 & 0.90 & 0.43 \\
MEND & 0.12 & 0.15 & 0.12 & 0.14 & 0.25 & 0.32 & 0.23 & 0.40 & 0.11 & 0.14 & 0.11 & 0.17 & 0.17 & 0.24 & 0.17 & 0.29 & 0.08 & 0.09 & 0.08 & 0.09 & 0.16 & 0.20 & 0.16 & 0.21 \\
GRACE & 0.53 & 0.36 & 0.36 & 0.41 & 0.91 & 0.72 & 0.62 & \textbf{0.97} & 0.53 & 0.37 & 0.40 & 0.45 & 0.87 & 0.64 & 0.61 & \textbf{0.98} & 0.54 & 0.26 & 0.31 & 0.55 & 0.90 & 0.47 & 0.50 & \textbf{0.99} \\
BalancEdit & 0.23 & 0.27 & 0.21 & 0.33 & 0.39 & 0.49 & 0.36 & 0.51 & 0.23 & 0.26 & 0.21 & 0.34 & 0.36 & 0.44 & 0.34 & 0.59 & 0.19 & 0.24 & 0.18 & 0.40 & 0.32 & 0.44 & 0.31 & 0.56 \\
ReasonEdit & 0.40 & 0.40 & 0.37 & 0.34 & 0.97 & 0.96 & 0.93 & 0.54 & 0.40 & 0.42 & 0.39 & 0.35 & 0.87 & 0.88 & 0.84 & 0.71 & 0.53 & 0.53 & 0.51 & 0.53 & \textbf{0.99} & 0.98 & 0.95 & 0.64 \\
\midrule
FT-Soft & 0.74 & 0.59 & 0.61 & 0.43 & 0.98 & 0.89 & 0.91 & 0.30 & 0.74 & 0.58 & 0.64 & 0.47 & 0.98 & 0.90 & 0.94 & 0.48 & 0.79 & 0.55 & 0.66 & 0.42 & \textbf{0.99} & 0.85 & 0.94 & 0.26 \\
MEND-Soft & 0.13 & 0.15 & 0.12 & 0.15 & 0.23 & 0.33 & 0.23 & 0.48 & 0.13 & 0.16 & 0.13 & 0.16 & 0.26 & 0.33 & 0.26 & 0.44 & 0.06 & 0.07 & 0.06 & 0.06 & 0.13 & 0.16 & 0.13 & 0.20 \\
GRACE-Soft & 0.91 & 0.50 & 0.56 & 0.42 & 0.90 & 0.72 & 0.62 & \textbf{0.97} & 0.87 & 0.48 & 0.59 & 0.48 & 0.87 & 0.64 & 0.61 & \textbf{0.98} & \textbf{0.92} & 0.29 & 0.45 & 0.55 & 0.90 & 0.47 & 0.50 & \textbf{0.99} \\
BalancEdit-Soft & 0.22 & 0.26 & 0.20 & 0.32 & 0.36 & 0.45 & 0.33 & 0.44 & 0.23 & 0.26 & 0.21 & 0.32 & 0.33 & 0.43 & 0.33 & 0.46 & 0.21 & 0.26 & 0.20 & 0.42 & 0.32 & 0.46 & 0.31 & 0.66 \\
ReasonEdit-Soft & 0.51 & 0.50 & 0.49 & 0.38 & \textbf{0.99} & \textbf{0.98} & 0.96 & 0.53 & 0.51 & 0.53 & 0.48 & 0.41 & 0.91 & 0.92 & 0.86 & 0.79 & 0.56 & 0.55 & 0.55 & 0.62 & \textbf{0.99} & \textbf{0.99} & \textbf{0.97} & 0.66 \\
\midrule
\textbf{SoftEdit} & \textbf{0.97} & \textbf{0.78} & \textbf{0.92} & \textbf{0.61} & \textbf{0.99} & 0.92 & \textbf{0.98} & 0.92 & \textbf{0.98} & \textbf{0.82} & \textbf{0.88} & \textbf{0.64} & \textbf{1.00} & \textbf{0.98} & \textbf{0.99} & 0.78 & 0.89 & \textbf{0.67} & \textbf{0.82} & \textbf{0.65} & 0.94 & 0.88 & 0.93 & 0.95 \\
\bottomrule
\CodeAfter
\begin{tikzpicture}

\end{tikzpicture}
\end{NiceTabular}
}\par
\vspace{1.5mm}
{\small (1) VLM Editing on VQAv2 (19,186 samples)}\par
\vspace{1mm}

% (2) LLM Editing on ProtoQA (9,762 samples)
\resizebox{0.8000\textwidth}{!}{%
\setlength{\tabcolsep}{3pt}%
\begin{NiceTabular}{l>{\centering\arraybackslash }b{2.0em}>{\centering\arraybackslash }b{2.0em}>{\centering\arraybackslash }b{2.0em}>{\centering\arraybackslash }b{2.0em}>{\centering\arraybackslash }b{2.0em}>{\centering\arraybackslash }b{2.0em}>{\centering\arraybackslash }b{2.0em}>{\centering\arraybackslash }b{2.0em}>{\centering\arraybackslash }b{2.0em}>{\centering\arraybackslash }b{2.0em}>{\centering\arraybackslash }b{2.0em}>{\centering\arraybackslash }b{2.0em}>{\centering\arraybackslash }b{2.0em}>{\centering\arraybackslash }b{2.0em}>{\centering\arraybackslash }b{2.0em}>{\centering\arraybackslash }b{2.0em}>{\centering\arraybackslash }b{2.0em}>{\centering\arraybackslash }b{2.0em}}
\CodeBefore
\begin{tikzpicture}
  \fill[orange!6] (2-|2) rectangle (15-|5) ;
  \fill[orange!6] (2-|8) rectangle (15-|11) ;
  \fill[orange!6] (2-|14) rectangle (15-|17) ;
\end{tikzpicture}
\Body

\toprule
 \Block{2-1}{\textbf{Editor}} & \multicolumn{6}{c}{\textbf{Llama-3.1-8B (8,326 errors)}} & \multicolumn{6}{c}{\textbf{Mistral-7B (7,782 errors)}} & \multicolumn{6}{c}{\textbf{Qwen2.5-7B (7,207 errors)}} \\
\cmidrule(lr){2-4} \cmidrule(lr){5-7} \cmidrule(lr){8-10} \cmidrule(lr){11-13} \cmidrule(lr){14-16} \cmidrule(lr){17-19}
 & \footnotesize Align & \footnotesize SGen & \footnotesize SLoc & \footnotesize Acc & \footnotesize Gen & \footnotesize Loc & \footnotesize Align & \footnotesize SGen & \footnotesize SLoc & \footnotesize Acc & \footnotesize Gen & \footnotesize Loc & \footnotesize Align & \footnotesize SGen & \footnotesize SLoc & \footnotesize Acc & \footnotesize Gen & \footnotesize Loc \\
\midrule
Unedited & 0.04 & 0.04 & 0.07 & 0.01 & 0.03 & 1.00 & 0.03 & 0.03 & 0.11 & 0.02 & 0.07 & 1.00 & 0.04 & 0.06 & 0.17 & 0.00 & 0.10 & 1.00 \\
\midrule
FT & 0.48 & 0.41 & 0.29 & 0.98 & 0.86 & 0.25 & 0.48 & 0.37 & 0.25 & 0.98 & 0.80 & 0.34 & 0.48 & 0.40 & 0.25 & 0.97 & 0.83 & 0.20 \\
MEND & 0.14 & 0.11 & 0.07 & 0.40 & 0.32 & 0.10 & 0.05 & 0.03 & 0.03 & 0.19 & 0.13 & 0.08 & 0.04 & 0.03 & 0.02 & 0.12 & 0.10 & 0.05 \\
GRACE & 0.49 & 0.27 & 0.15 & 0.34 & 0.21 & 0.96 & 0.50 & 0.04 & 0.11 & 0.59 & 0.09 & \textbf{1.00} & 0.48 & 0.26 & 0.20 & 0.58 & 0.35 & \textbf{0.98} \\
IKE & 0.08 & 0.07 & 0.09 & 0.23 & 0.18 & 0.59 & 0.08 & 0.06 & 0.08 & 0.26 & 0.20 & 0.49 & 0.36 & 0.30 & 0.23 & 0.85 & 0.71 & 0.50 \\
AlphaEdit & 0.32 & 0.20 & 0.19 & 0.69 & 0.44 & 0.25 & 0.42 & 0.23 & 0.24 & 0.88 & 0.46 & 0.38 & 0.28 & 0.10 & 0.21 & 0.58 & 0.21 & 0.32 \\
\midrule
FT-Soft & 0.81 & 0.55 & 0.37 & \textbf{1.00} & 0.87 & 0.23 & 0.75 & 0.46 & 0.31 & \textbf{1.00} & 0.83 & 0.33 & 0.80 & 0.52 & 0.31 & \textbf{0.99} & 0.85 & 0.27 \\
MEND-Soft & 0.19 & 0.15 & 0.13 & 0.51 & 0.43 & 0.15 & 0.04 & 0.03 & 0.06 & 0.14 & 0.11 & 0.21 & 0.06 & 0.05 & 0.06 & 0.19 & 0.15 & 0.14 \\
GRACE-Soft & \textbf{0.95} & 0.44 & 0.17 & 0.35 & 0.22 & \textbf{0.97} & 0.92 & 0.05 & 0.12 & 0.57 & 0.09 & \textbf{1.00} & 0.87 & 0.36 & 0.20 & 0.58 & 0.35 & \textbf{0.98} \\
IKE-Soft & 0.07 & 0.07 & 0.10 & 0.20 & 0.17 & 0.58 & 0.08 & 0.08 & 0.07 & 0.25 & 0.23 & 0.49 & 0.34 & 0.30 & 0.35 & 0.80 & 0.71 & 0.50 \\
AlphaEdit-Soft & 0.38 & 0.22 & 0.21 & 0.67 & 0.45 & 0.27 & 0.40 & 0.21 & 0.25 & 0.68 & 0.42 & 0.31 & 0.32 & 0.10 & 0.23 & 0.62 & 0.20 & 0.36 \\
\midrule
\textbf{SoftEdit} & \textbf{0.95} & \textbf{0.72} & \textbf{0.47} & 0.98 & \textbf{0.94} & 0.76 & \textbf{0.93} & \textbf{0.70} & \textbf{0.52} & 0.98 & \textbf{0.95} & 0.69 & \textbf{0.93} & \textbf{0.73} & \textbf{0.52} & 0.98 & \textbf{0.94} & 0.77 \\
\bottomrule
\CodeAfter
\begin{tikzpicture}

\end{tikzpicture}
\end{NiceTabular}
}\par
\vspace{1.5mm}
{\small (2) LLM Editing on ProtoQA (9,762 samples)}\par
\vspace{-2mm}
\caption{Batch editing results on VLMs and LLMs using the VQAv2 and ProtoQA datasets.}
\label{tab:main_results}
\end{table*}

\xhdr{Comparison to prior editors}
We first compare SoftEdit with state-of-the-art editors adapted to model calibration in static editing, where all edits are available for each editor. 
Soft editors use the distribution over all plausible answers as the target, while their standard counterparts use a one-hot target on the majority-vote answer.
% calibration
As shown in Table~\ref{tab:main_results}, all soft editors outperform their non-soft counterparts in calibration (Align, SGen, and SLoc) while maintaining comparable editing performance, suggesting that calibration constitutes a broader objective beyond single-answer correction.
SoftEdit consistently outperforms other soft editors in Align, SGen, and SLoc, showing strongest calibration generalization while preserving genuine distribution shifts. 
Besides calibration, SoftEdit shows comparable model editing performance to state-of-the-art editors, with high Acc, Gen, and Loc indicating accurate answer generation on edits and unseen samples.
GRACE-Soft achieves the second-best alignment by steering each edit's activation to produce the target distribution. Applying the same steered activation to nearby unseen samples yields better SGen than other soft editors. However, assigning these samples the same answer distribution erases genuine distributional shifts, resulting in poor SLoc. 
% but optimizing \(\mathcal{L}_{\text{align}}\) alone only mildly generalize calibration to unseen samples, yielding low SGen. In contrast, SoftEdit’s graph-Laplacian generality term improves SGen.
%
Batch fine-tuning (FT) achieves the second-best SGen--SLoc tradeoff by learning global calibration across thousands of edits, but its lower edit alignment indicates weaker reliability and precision for local targeted calibration.
% Comparison to prior calibration methods.
We also compare SoftEdit with three post-training calibration methods: Temperature Scaling (TS)~\citep{guo2017calibration}, Batch Calibration (BCL)~\citep{zhou2024batch}, and Adaptive Temperature Scaling (ATS)~\citep{xie2024calibrating}, along with their variants adapted for one-shot localized calibration. We find that global calibration methods fail to improve performance because (1) globally averaged temperature or bias adjustments cannot address heterogeneous calibration needs across individual edits, and (2) temperature scaling cannot correct incorrectly ranked answers. Directly retrieving the correct first-token distribution for each edit improves performance by encouraging the model to start with the correct tokens, but (1) suffers from poor soft locality by overwriting genuine distribution shifts across semantic variants with the same distribution and (2) fails to correctly generate full answers. More details on implementation, results, and discussion are provided in Appendix~\ref{app:experiment_baselines_calib}.

\begin{figure}
    \centering
    \includegraphics[width=1\linewidth]{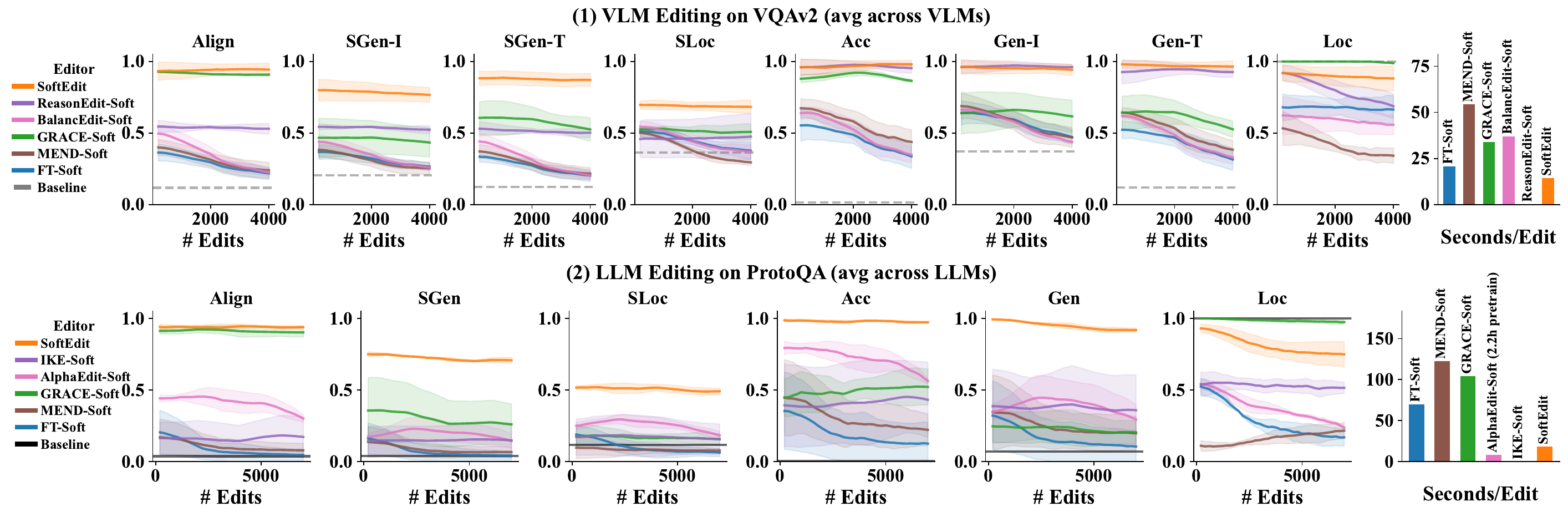}
    \caption{Sequential one-shot calibration performance and computational efficiency across soft VLM and LLM editors. Results are averaged across edited pretrained models.}
    \label{fig:seq_main}
\end{figure}

\begin{table*}[t]
\centering

% (1) VLM Editing on VQAv2 (19,186 samples)
\resizebox{\textwidth}{!}{%
\setlength{\tabcolsep}{3pt}%
\begin{NiceTabular}{l>{\centering\arraybackslash }b{2.0em}>{\centering\arraybackslash }b{2.0em}>{\centering\arraybackslash }b{2.0em}>{\centering\arraybackslash }b{2.0em}>{\centering\arraybackslash }b{2.0em}>{\centering\arraybackslash }b{2.0em}>{\centering\arraybackslash }b{2.0em}>{\centering\arraybackslash }b{2.0em}>{\centering\arraybackslash }b{2.0em}>{\centering\arraybackslash }b{2.0em}>{\centering\arraybackslash }b{2.0em}>{\centering\arraybackslash }b{2.0em}>{\centering\arraybackslash }b{2.0em}>{\centering\arraybackslash }b{2.0em}>{\centering\arraybackslash }b{2.0em}>{\centering\arraybackslash }b{2.0em}>{\centering\arraybackslash }b{2.0em}>{\centering\arraybackslash }b{2.0em}>{\centering\arraybackslash }b{2.0em}>{\centering\arraybackslash }b{2.0em}>{\centering\arraybackslash }b{2.0em}>{\centering\arraybackslash }b{2.0em}>{\centering\arraybackslash }b{2.0em}>{\centering\arraybackslash }b{2.0em}}
\CodeBefore
\begin{tikzpicture}
  \fill[orange!6] (2-|2) rectangle (9-|6) ;
  \fill[orange!6] (2-|10) rectangle (9-|14) ;
  \fill[orange!6] (2-|18) rectangle (9-|22) ;
\end{tikzpicture}
\Body

\toprule
 \Block{2-1}{\textbf{Editor}} & \multicolumn{8}{c}{\textbf{LLaVA-1.5-7B (4,436 errors)}} & \multicolumn{8}{c}{\textbf{InstructBLIP-7B (4,625 errors)}} & \multicolumn{8}{c}{\textbf{Qwen3-4B (4,157 errors)}} \\
\cmidrule(lr){2-9} \cmidrule(lr){10-17} \cmidrule(lr){18-25}
 & \footnotesize \textbf{Rank-Align} & \footnotesize \textbf{Rank-SGen-I} & \footnotesize \textbf{Rank-SGen-T} & \footnotesize \textbf{Rank-SLoc} & \footnotesize Acc & \footnotesize Gen-I & \footnotesize Gen-T & \footnotesize Loc & \footnotesize \textbf{Rank-Align} & \footnotesize \textbf{Rank-SGen-I} & \footnotesize \textbf{Rank-SGen-T} & \footnotesize \textbf{Rank-SLoc} & \footnotesize Acc & \footnotesize Gen-I & \footnotesize Gen-T & \footnotesize Loc & \footnotesize \textbf{Rank-Align} & \footnotesize \textbf{Rank-SGen-I} & \footnotesize \textbf{Rank-SGen-T} & \footnotesize \textbf{Rank-SLoc} & \footnotesize Acc & \footnotesize Gen-I & \footnotesize Gen-T & \footnotesize Loc \\
\midrule
Unedited & 0.15 & 0.26 & 0.17 & 0.35 & 0.02 & 0.38 & 0.12 & 1.00 & 0.15 & 0.27 & 0.18 & 0.38 & 0.02 & 0.35 & 0.12 & 1.00 & 0.15 & 0.29 & 0.18 & 0.50 & 0.01 & 0.39 & 0.11 & 1.00 \\
\midrule
FT-Soft & 0.78 & 0.60 & 0.63 & 0.52 & 0.98 & 0.88 & 0.91 & 0.34 & 0.79 & 0.61 & 0.67 & 0.56 & 0.98 & 0.90 & 0.93 & 0.50 & 0.80 & 0.54 & 0.66 & 0.49 & \textbf{0.99} & 0.85 & \textbf{0.94} & 0.25 \\
MEND-Soft & 0.19 & 0.22 & 0.18 & 0.27 & 0.23 & 0.30 & 0.22 & 0.40 & 0.13 & 0.14 & 0.13 & 0.15 & 0.12 & 0.13 & 0.12 & 0.10 & 0.14 & 0.15 & 0.13 & 0.16 & 0.15 & 0.17 & 0.14 & 0.18 \\
GRACE-Soft & \textbf{0.95} & 0.55 & 0.60 & 0.51 & 0.88 & 0.74 & 0.62 & \textbf{0.98} & 0.83 & 0.48 & 0.56 & 0.54 & 0.86 & 0.64 & 0.61 & \textbf{0.99} & \textbf{0.96} & 0.34 & 0.45 & 0.54 & 0.89 & 0.47 & 0.50 & \textbf{0.99} \\
BalancEdit-Soft & 0.26 & 0.31 & 0.24 & 0.38 & 0.36 & 0.45 & 0.33 & 0.44 & 0.26 & 0.29 & 0.25 & 0.37 & 0.34 & 0.43 & 0.33 & 0.52 & 0.24 & 0.30 & 0.23 & 0.43 & 0.30 & 0.45 & 0.30 & 0.61 \\
\midrule
\textbf{SoftEdit} & 0.94 & \textbf{0.72} & \textbf{0.84} & \textbf{0.69} & \textbf{0.99} & \textbf{0.92} & \textbf{0.98} & 0.92 & \textbf{0.96} & \textbf{0.70} & \textbf{0.78} & \textbf{0.68} & \textbf{0.99} & \textbf{0.97} & \textbf{0.99} & 0.82 & 0.92 & \textbf{0.65} & \textbf{0.78} & \textbf{0.73} & 0.95 & \textbf{0.89} & 0.93 & 0.96 \\
\bottomrule
\CodeAfter
\begin{tikzpicture}

\end{tikzpicture}
\end{NiceTabular}
}\par
\vspace{1.5mm}
{\small (1) VLM Editing on VQAv2 (19,186 samples)}\par
\vspace{1mm}

% (2) LLM Editing on ProtoQA (9,762 samples)
\resizebox{0.8000\textwidth}{!}{%
\setlength{\tabcolsep}{3pt}%
\begin{NiceTabular}{l>{\centering\arraybackslash }b{2.0em}>{\centering\arraybackslash }b{2.0em}>{\centering\arraybackslash }b{2.0em}>{\centering\arraybackslash }b{2.0em}>{\centering\arraybackslash }b{2.0em}>{\centering\arraybackslash }b{2.0em}>{\centering\arraybackslash }b{2.0em}>{\centering\arraybackslash }b{2.0em}>{\centering\arraybackslash }b{2.0em}>{\centering\arraybackslash }b{2.0em}>{\centering\arraybackslash }b{2.0em}>{\centering\arraybackslash }b{2.0em}>{\centering\arraybackslash }b{2.0em}>{\centering\arraybackslash }b{2.0em}>{\centering\arraybackslash }b{2.0em}>{\centering\arraybackslash }b{2.0em}>{\centering\arraybackslash }b{2.0em}>{\centering\arraybackslash }b{2.0em}}
\CodeBefore
\begin{tikzpicture}
  \fill[orange!6] (2-|2) rectangle (9-|5) ;
  \fill[orange!6] (2-|8) rectangle (9-|11) ;
  \fill[orange!6] (2-|14) rectangle (9-|17) ;
\end{tikzpicture}
\Body

\toprule
 \Block{2-1}{\textbf{Editor}} & \multicolumn{6}{c}{\textbf{Llama-3.1-8B (8,326 errors)}} & \multicolumn{6}{c}{\textbf{Mistral-7B (7,782 errors)}} & \multicolumn{6}{c}{\textbf{Qwen2.5-7B (7,207 errors)}} \\
\cmidrule(lr){2-7} \cmidrule(lr){8-13} \cmidrule(lr){14-19}
 & \footnotesize \textbf{Rank-Align} & \footnotesize \textbf{Rank-SGen} & \footnotesize \textbf{Rank-SLoc} & \footnotesize Acc & \footnotesize Gen & \footnotesize Loc & \footnotesize \textbf{Rank-Align} & \footnotesize \textbf{Rank-SGen} & \footnotesize \textbf{Rank-SLoc} & \footnotesize Acc & \footnotesize Gen & \footnotesize Loc & \footnotesize \textbf{Rank-Align} & \footnotesize \textbf{Rank-SGen} & \footnotesize \textbf{Rank-SLoc} & \footnotesize Acc & \footnotesize Gen & \footnotesize Loc \\
\midrule
Unedited & 0.14 & 0.14 & 0.17 & 0.01 & 0.03 & 1.00 & 0.12 & 0.12 & 0.18 & 0.02 & 0.07 & 1.00 & 0.14 & 0.16 & 0.26 & 0.00 & 0.10 & 1.00 \\
\midrule
FT-Soft & 0.95 & 0.67 & 0.54 & \textbf{1.00} & 0.89 & 0.21 & 0.91 & 0.65 & 0.51 & \textbf{1.00} & 0.85 & 0.26 & 0.93 & 0.69 & 0.49 & \textbf{1.00} & 0.87 & 0.24 \\
MEND-Soft & 0.09 & 0.09 & 0.07 & 0.06 & 0.05 & 0.03 & 0.07 & 0.07 & 0.00 & 0.00 & 0.00 & 0.00 & 0.07 & 0.07 & 0.02 & 0.02 & 0.02 & 0.00 \\
GRACE-Soft & \textbf{0.97} & 0.56 & 0.32 & 0.35 & 0.24 & \textbf{0.98} & \textbf{0.96} & 0.15 & 0.16 & 0.57 & 0.09 & \textbf{1.00} & \textbf{0.97} & 0.48 & 0.33 & 0.57 & 0.37 & \textbf{0.99} \\
AlphaEdit-Soft & 0.48 & 0.32 & 0.33 & 0.64 & 0.45 & 0.27 & 0.53 & 0.33 & 0.45 & 0.68 & 0.45 & 0.32 & 0.44 & 0.20 & 0.34 & 0.61 & 0.21 & 0.37 \\
\midrule
\textbf{SoftEdit} & \textbf{0.97} & \textbf{0.70} & \textbf{0.59} & 0.98 & \textbf{0.94} & 0.75 & \textbf{0.96} & \textbf{0.67} & \textbf{0.70} & 0.98 & \textbf{0.95} & 0.69 & 0.94 & \textbf{0.71} & \textbf{0.67} & 0.98 & \textbf{0.95} & 0.79 \\
\bottomrule
\CodeAfter
\begin{tikzpicture}

\end{tikzpicture}
\end{NiceTabular}
}\par
\vspace{1.5mm}
{\small (2) LLM Editing on ProtoQA (9,762 samples)}\par
\vspace{-2mm}
\caption{Batch editing VLMs and LLMs when human feedback only provides ranked answers.}
\label{tab:ablate_rank}
\end{table*}

\xhdr{Sequential one-shot calibration and computing efficiency} We next examine sequential editing, where each editor sequentially calibrates one edit at a time to repeatedly update a pretrained model. We evaluate the model every 200 edits on 50 random previous edits. As shown in Fig.~\ref{fig:seq_main}, SoftEdit maintains consistently better and more stable one-shot calibration performance than other soft editors over thousands of edits. This indicates that $\mathcal{L}_{\text{gen}}$ and $\mathcal{L}_{\text{loc}}$ continuously balance the distributional generality--locality tradeoff as edits arrive in real time. 
% The strong editing performance indicates that the retrieval mechanism effectively localizes each edit, preventing interference. 
%
As shown in right panels, SoftEdit also achieves high computational efficiency by editing only the final projection layer. 
% GRACE-soft shows high locality, but much lower generalities (Gen and SGen), suggesting overly conservative one-shot editing. 
As expected, weight-updating editors exhibit catastrophic forgetting and model degradation after a few hundred edits while FT performs notably worse one-shot than batched.

\xhdr{Calibration with ranked answers}
Finally, we study one-shot calibration when given only a ranking of plausible answers without explicit probability estimates. A practical example of this common setting is clinicians often providing differential diagnoses as ranked conditions without numerical uncertainty.
SoftEdit converts ranked answers into a tie-preserving pseudo-distribution (Appendix~\ref{app:ranked}), which is also used for compatible soft editors.
As shown in Table~\ref{tab:ablate_rank}, SoftEdit outperforms all soft editors in ranking alignment and generalization, measured by gated Kendall's \(\tau\) (Appendix~\ref{app:gated_rank_align}). 
% FT achieves the second-best 
% why softedit work but others dont
These results show that SoftEdit remains effective under weaker, ordinal feedback: generalizing the pseudo-calibrated distribution generalizes the target ranking on related inputs, while locality preserves genuine rank changes in semantically similar inputs. 

\subsection{Ablation Studies}

\begin{figure}
    \centering
    \includegraphics[width=1\linewidth]{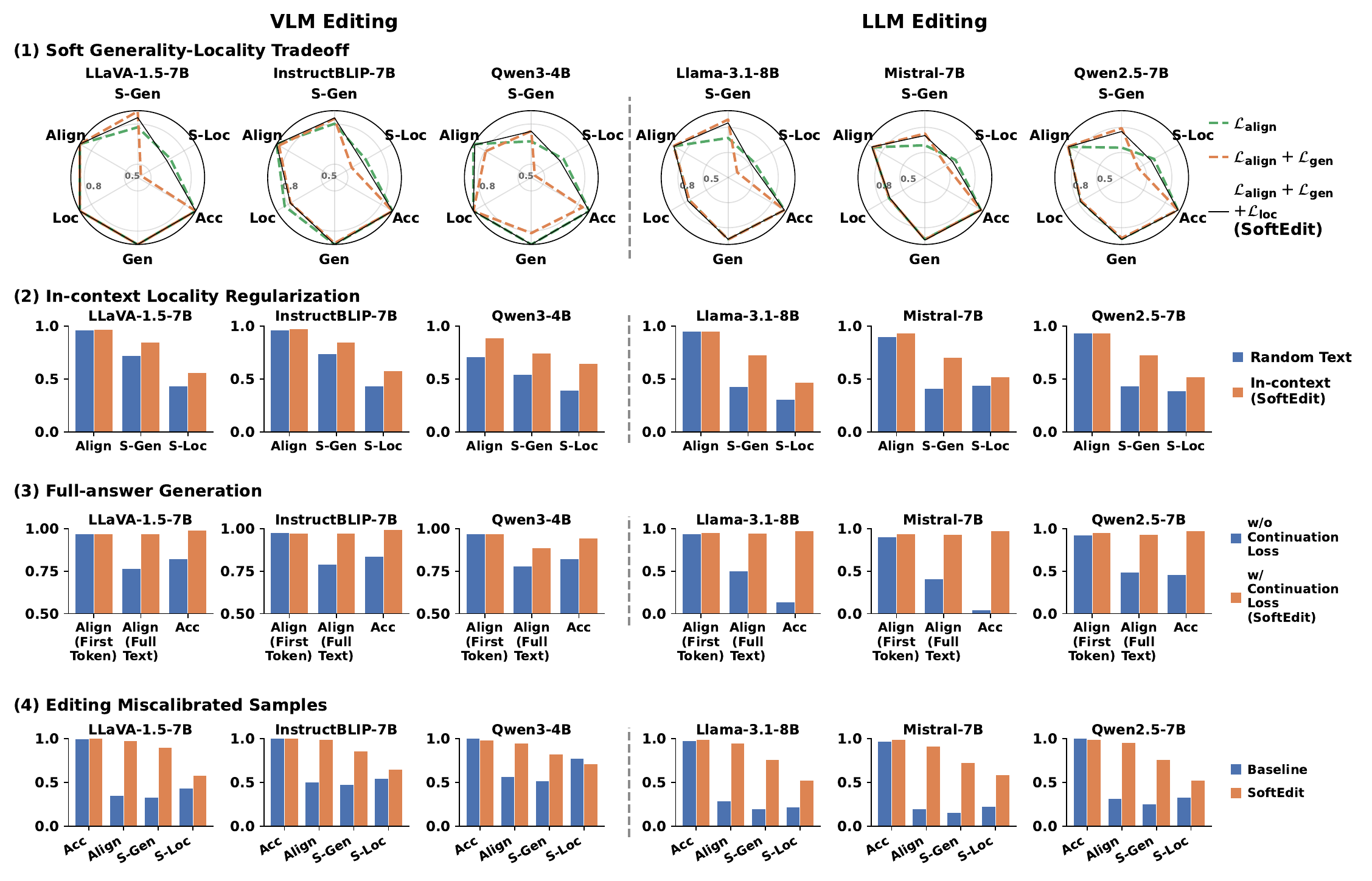}
    \caption{Ablation studies of SoftEdit by varying its objectives, locality samples, and input edits.}
    \label{fig:ablations}
\end{figure}

We next systematically ablate SoftEdit, and the main results are shown in Fig.~\ref{fig:ablations}. 

% \vspace{-1mm}
\xhdr{(1) Soft generality--locality tradeoff} We first ablate SoftEdit by removing the distributional generality and locality terms, $\mathcal{L}_{\text{gen}}$ and $\mathcal{L}_{\text{loc}}$, one at a time. As shown in panel (1), optimizing $\mathcal{L}_{\text{align}}$ alone preserves real distribution shifts (high SLoc) but generalizes calibration poorly (low SGen), whereas adding $\mathcal{L}_{\text{gen}}$ alone improves SGen but decreases SLoc. By including both terms, SoftEdit optimizes both SGen and SLoc, effectively balancing the distributional generality and locality.

\vspace{-1mm}
\xhdr{(2) In-context locality regularization} In panel (2), replacing answer-derived in-context locality text with random sentences substantially degrades calibration, showing that targeted local regularization is more effective than commonly used random locality samples.

\vspace{-1mm}
\xhdr{(3) Full-answer continuation} In panel (3), we remove the continuation term from \(\mathcal{L}_{\text{align}}\) and calibrate models using only first-token alignment. Without the continuation loss, the edited model fails to generate complete answers, leading to poor accuracy and answer-level distribution alignment.

\vspace{-1mm}
\xhdr{(4) Correctly-answered but miscalibrated edits} In panel (4), we use 2,000 random samples from \(D_u\) where the original model generates answers correctly but remains distributionally miscalibrated. SoftEdit effectively calibrates these samples while preserving the original answer accuracy.

In Appendix~\ref{app:additional_ablation}, we conduct additional ablation and sensitivity analyses on SoftEdit:
(1) The auto-tuned retrieval rejection threshold achieves a better generality–locality tradeoff than fixed distance percentiles.
(2) The lightweight LLM paraphrase judge improves locality for LLM editing by distinguishing subtle semantic differences missed by embedding-only retrieval.
(3) Probability-weighted $A_{ij}$ in graph-Laplacian regularization achieves the best overall performance.
(4) The auto-tuned balancing weights $\lambda_{\mathrm{loc}}$ and $\lambda_{\mathrm{gen}}$ remain stable when estimated using less than 100 initial edits.
%  In Appendix~\ref{app:additional_results}, we provide additional results showing that SoftEdit consistently improves image soft generality on both photometric perturbations and generated semantic variants, its advantages in sequential editing hold across individual VLMs and LLMs, and adding $\mathcal{L}_{\mathrm{loc}}$ alone yields similar performance to optimizing $\mathcal{L}_{\mathrm{align}}$ alone.

% We also ablate the retrieval-key embedding for LLM edits by replacing sentence encoder \texttt{\small all-MiniLM-L6-v2} with the last-token hidden state from the final layer of each LLM. This improves locality by around $0.07$ on \llmmistral, while all other performance metrics remain comparable. A comparison table is provided in Table~\ref{tab:ablate_key}.

\section{Conclusion}
We introduce soft model editing, extending model editing from single-answer correction to targeted one-shot calibration using uncertainty in multi-answer human feedback. 
We propose SoftEdit, an uncertainty-aware editor that generalizes calibration to unseen related samples while preserving genuine distributional shifts. 
Across LLMs and VLMs, SoftEdit achieves state-of-the-art performance and efficiency, particularly for real-time sequential editing.
% and efficient sequential editing, showing that uncertainty in user feedback enables generalizable model calibration beyond single-error correction.
%
Interesting future directions include detecting labeling errors from aleatoric uncertainty and extending soft editing to multi-output settings, such as differential diagnosis tasks for medical language models.

\subsubsection*{Acknowledgments}
This work was sponsored in part by NIH Award \#1OT2OD038079-01.
We thank University of Virginia's High Performance Computing team for computing resources. 
We thank Marco Gutierrez, Yash Kumar Atri, Arinbjörn Kolbeinsson, Moeen Mostafavi, and Zoë Gray for their helpful feedback on this project and manuscript.

\bibliography{reference}
\bibliographystyle{iclr2027_conference}

\appendix
\newpage
\section{Evaluation Details}

Let $e=(x,\mathcal{Y}_{K},\mathbf{p}_{K})$ denote an edit, where $x$ is an input query to a pretrained model, $\mathcal{Y}_{K}=(y_1,\ldots,y_K)$ is the corresponding set of $K$ plausible human-provided answers.
For a language model, $x$ is a text question; for a vision-language model, $x$ consists of an image and a text question.
The $K$ answers follow an answer-level target distribution $\mathbf{p}_{K}(\cdot\mid x)=(p_1,\ldots,p_K)$ estimated from human feedback, where $p_k$ denotes the probability that a human provides $y_k$ as the answer.

We aim to develop a model editor $g(\cdot)$ that takes in an edit $e$ and a pretrained model $f$, and outputs a locally calibrated model $f_{\text{new}}=g(e,f)$ that achieves the following properties:  Given the edit query $x$, let $\widehat{\mathbf{p}}_{K}(\cdot\mid x;f_{\text{new}})$ denote the model's distribution over the $K$ human-provided answers under stochastic generation. $f_{\text{new}}$ should align $\widehat{\mathbf{p}}_{K}$ with the human target distribution $\mathbf{p}_{K}$, such that each answer $y_k$ is generated in full text with target probability $p_k$. Further, $f_{\text{new}}$ should generalize the calibrated distribution to unseen queries that are semantically and distributionally similar to the edit, while preserving genuine distributional shifts in other unseen queries.

Evaluating $f_{\text{new}}$ requires comparing its answer-level distribution $\widehat{\mathbf{p}}_{K}$ against the human target $\mathbf{p}_{K}$.
Because Monte Carlo estimation of $\widehat{\mathbf{p}}_{K}$ requires expensive sampling and decoding full responses per query, we first introduce a fast, deterministic readout that approximates the answer-level distribution using a single forward pass for first-token probability combined with teacher-forced continuation validation for text generation (Sec.~\ref{app:gated_readout}).
Building on this readout, we define two alignment metrics: the \emph{gated Jensen--Shannon distance} $\mathrm{JS_{gate}}$, which measures distributional closeness between $\widehat{\mathbf{p}}_{K}$ and $\mathbf{p}_{K}$ (Sec.~\ref{app:gated_align}), and \emph{gated Kendall's~$\tau_{gate}$}, which measures rank-order agreement between the top-K model-generated answers and the K human-provided answers (Sec.~\ref{app:gated_rank_align}).

We then detail the evaluation procedure for the edited model $f_{\text{new}}$ in Sec.~\ref{app:eval_metrics} from two complementary perspectives.
First, we evaluate the editor's conventional \emph{\textbf{one-shot model editing}} performance using accuracy-based metrics---whether $f_{\text{new}}$ generates a correct answer from $\mathcal{Y}^{H}_{K}$ on the edit input (\emph{reliability}) and on semantically equivalent inputs (\emph{generality}), and whether it preserves correct predictions on unrelated inputs (\emph{locality})---and describe the corresponding dataset construction in Sec.~\ref{app:eval_model_editing}.
Second, we evaluate the editor's \emph{\textbf{one-shot calibration}} performance on the model---whether $f_{\text{new}}$ produces a well-calibrated answer distribution on the edit input (\emph{alignment}), generalizes this calibration to semantically and distributionally similar inputs (\emph{soft generality}), and preserves genuine distributional shifts (\emph{soft locality})---and describe the construction of evaluation data for each property in Sec.~\ref{app:eval_model_calibration}.

Lastly, we provide details of the LLM-as-a-judge validation of generated images used for VLM editing evaluation in Sec.~\ref{app:experiment_gen_image_validation}.

\subsection{Fast Answer-Level Distribution Estimation without Monte Carlo Sampling}\label{app:gated_readout}

We define a model's \emph{answer-level} distribution by the probability that a stochastic generation from the model would produce each answer $y_k$ as a complete text response.
The standard approach is Monte Carlo estimation~\citep{lee2023can, sorensen2022information}: sample $N$ full generations $\{\tilde{y}^{(1)},\ldots,\tilde{y}^{(N)}\}$ and estimate the answer-level probability by the empirical frequency of exact text matches,
$p_{\mathrm{MC}}(y_k\mid x)=\frac{1}{N}\sum_{n=1}^{N}\mathbf{1}\!\bigl[\tilde{y}^{(n)}=y_k\bigr].$
However, this requires sampling and decoding $N$ full responses per query, and is prohibitively expensive when the model is evaluated at scale.

A common alternative is to use the model's first-token probability as a direct proxy for the answer-level probability~\citep{lee2023can, ma2025large}. This requires only a single forward pass, but the first-token probability is only an upper bound on the probability of the complete answer: it does not verify whether the model would continue along that selected first token to generate the intended multi-token answer, and multiple answers may share the same first token.

We therefore introduce a fast answer-level probability estimation method by combining the first-token distribution with a teacher-forced continuation gate to validate complete answer generation.

\paragraph{Setup.}
Let the human answers be
$$
\mathcal{Y}_K=\{y_1,\ldots,y_K\},
\qquad
y_k=(t_{k,1},t_{k,2},\ldots,t_{k,L_k}),
$$
with human first-token set
$$
\mathcal{T}=\{t_{1,1},\ldots,t_{K,1}\}.
$$
Let the model readout space be
$$
\widehat{\mathcal{Y}}
=
\{\widehat{y}_1,\ldots,\widehat{y}_K,\widehat{y}_{K+1}\},
\qquad
\widehat{y}_{K+1}\equiv\mathrm{OTHER},
$$
where $\widehat{y}_1,\ldots,\widehat{y}_K$ are the model's top-$K$ most probable first-token answers, with corresponding first tokens
$$
\widehat{\mathcal{T}}
=
\{\widehat{t}_{1,1},\ldots,\widehat{t}_{K,1}\},
$$
and exact first-step probabilities $\widehat{p}(\widehat{t}_{j,1}\mid x)$. The $\mathrm{OTHER}$ category collects all remaining probability mass not assigned to the top-$K$ answers.

\paragraph{Continuation gate.}
For each model answer $\widehat{y}_j$ whose first token matches a human answer's first token, $\widehat{t}_{j,1}=t_{k,1}\in\mathcal{T}$, we check whether the model would complete the full answer $y_k$ under teacher forcing accuracy:
$$
c_j(x)
=
\mathbf{1}\!\bigl[\operatorname{TFAcc}(t_{k,2:L_k}\mid x,\,t_{k,1})=1\bigr].
$$
Specifically, the prompt $x$ and the full answer tokens $(t_{k,1},\ldots,t_{k,L_k})$ are concatenated and fed to the model in a single forward pass. 
At each continuation position $\ell=2,\ldots,L_k$, the model's most likely prediction is
$\widehat{t}_{k,\ell} = \arg\max\, \widehat{p}(\cdot\mid x,t_{k,1},\ldots,t_{k,\ell-1})$.
If $\widehat{t}_{k,\ell}=t_{k,\ell}$ for every position $\ell$, that is, if the argmax prediction matches the target token at every continuation position, the gate passes and $c_j=1$; otherwise $c_j=0$.

\paragraph{Gated readout.}
For $j=1,\ldots,K$, we estimate answer-level probability as
$$
\widehat{p}(\widehat{y}_j\mid x)
=
\begin{cases}
p_\theta(\widehat{t}_{j,1}\mid x)\cdot c_j(x),
& \widehat{t}_{j,1}=t_{k,1}\in\mathcal{T},\\[2mm]
p_\theta(\widehat{t}_{j,1}\mid x),
& \widehat{t}_{j,1}\notin\mathcal{T}.
\end{cases}
$$
That is, if a model answer's first token matches a human answer's first token and the continuation validation gate passes ($c_j=1$), the first-token probability is retained as the answer-level estimate. If the continuation validation fails ($c_j=0$), the probability is zeroed because the model fails to completely generate the answer.

All probability mass not ultimately assigned to the $K$ human-provided answers is collected into a residual category $\mathrm{OTHER}$. This includes both model answers whose first token does not match any human answer ($\widehat{t}_{j,1}\notin\mathcal{T}$) and any mass zeroed by a failed continuation gate:
$$
\widehat{p}(\widehat{y}_{K+1}\mid x)
=
\widehat{p}(\mathrm{OTHER}\mid x)
=
1-\sum_{j=1}^{K}\widehat{p}(\widehat{y}_j\mid x).
$$
Hence, the complete readout of the model's probability over the $K$ human-provided answers is the $(K\!+\!1)$-dimensional vector
$\widehat{\mathbf{p}}(\cdot\mid x)
=
\bigl[\widehat{p}(\widehat{y}_1\mid x),\ldots,\widehat{p}(\widehat{y}_K\mid x),\;\widehat{p}(\mathrm{OTHER}\mid x)\bigr].$

\paragraph{Discussion.}
This approach requires only one forward pass for the full-vocabulary first-token distribution plus at most $K$ teacher-forced continuation checks, providing a fast approximation to the Monte Carlo answer-level distribution.
It improves upon a first-token-only probability readout~\citep{lee2023can, ma2025large} in two ways: (1)~the continuation gate validates that the model would complete the full answer, zeroing the probability when it would not; (2)~probability mass is retained in a residual $\mathrm{OTHER}$ category rather than renormalized via softmax over the human answer set, so model predictions outside the human answer set are penalized explicitly.

We note an inherent limitation shared by first-token-based readout methods~\citep{lee2023can, ma2025large}: when multiple human answers share the same first token (e.g., ``red car'' and ``red bus''), the readout cannot distinguish them at the first-token level. In this case, we sum the probabilities of answers that share the same first token into a single probability score for that first token in both the human-target and model distributions, so that the distributional alignment metrics (e.g., $\mathrm{JS_{gate}}$) correctly compare the probability associated with each first token. Then, the continuation gate checks each colliding answer, and the gate passes if any of them passes, validating that the model would complete at least one of the colliding answers.

\subsubsection{Distributional Alignment: Gated Jensen--Shannon Distance $\mathrm{JS_{gate}}$}\label{app:gated_align}

We measure distributional closeness between the human target $\mathbf{p}_{K}(\cdot\mid x)$ and the model gated readout $\widehat{\mathbf{p}}_K(\cdot\mid x)$ using the Jensen--Shannon (JS) distance.
Both distributions are first projected onto the shared $(K\!+\!1)$-dimensional readout space $\widehat{\mathcal{Y}}=\{\widehat{y}_1,\ldots,\widehat{y}_K,\mathrm{OTHER}\}$ defined above. For the human-provided target answers, probabilities sharing the same first token are summed, and $p_k(\mathrm{OTHER}\mid x)=0$. For the model, the gated readout $\widehat{\mathbf{p}}(\cdot\mid x)$ is defined as above. The JS divergence is
$$
\mathrm{JSD}\!\bigl(\widehat{\mathbf{p}}_K\|\mathbf{p}_{K}\bigr)
=
\frac{1}{2}\sum_{j=1}^{K+1}
\left[
\widehat{p}_j\log_2\frac{\widehat{p}_j}{m_j}
+
p_j\log_2\frac{p_j}{m_j}
\right],
\qquad
m_j=\tfrac{1}{2}\bigl(\widehat{p}_j+p_j\bigr),
$$
We define the gated JS distance (lower is better) based on the gated readout as
\begin{equation}
    \mathrm{JS_{gate}} = \sqrt{\mathrm{JSD}\!\bigl(\widehat{\mathbf{p}}_K\|\mathbf{p}_{K}\bigr)} \;\in\; [0,1].
    \label{eq:js_gate}
\end{equation}

\subsubsection{Ranking Alignment: Gated Kendall's $\tau_{\mathrm{gate}}$}\label{app:gated_rank_align}

When only the ranking of the $K$ human answers is available (without explicit probability estimates), we measure rank-order agreement using Kendall's $\tau$ adapted for ties.

Both the human and model sides are represented as \emph{tier-lists}: ordered groups of answers, best first, where answers within the same group are tied. On the human side, answers with equal vote counts form a tier and are mapped to first-token keys. On the model side, gated readout probabilities $\widehat{p}(\widehat{y}_j\mid x)$ with equal values form a tier (the $\mathrm{OTHER}$ entry is excluded). We take the union of all human and model first-token keys as the item set $\mathcal{U}$; any item missing from one side is placed in a worst-rank tier on that side.

We then compare every pair of items $(a,b)\in\mathcal{U}$. A pair is \emph{concordant} if the human and model sides agree---either both rank $a$ above $b$ (or vice versa), or both tie them. A pair is \emph{discordant} in all other cases: the two sides reverse the order, or one side ties them while the other does not. The metric is the fraction of concordant pairs:
\begin{equation}
\tau_{\mathrm{gate}}
=
\frac{|\{\text{concordant pairs}\}|}{\binom{|\mathcal{U}|}{2}}
\;\in\;[0,1].
\end{equation}
Higher is better. Every pair is weighted equally, making $\tau_{\mathrm{gate}}$ an ordinal complement to $\mathrm{JS_{gate}}$.

\subsection{Evaluation Metrics}\label{app:eval_metrics}

We evaluate the editor's performance from two perspectives.
First, the editor should perform the new one-shot local calibration on the pretrained model, thus we evaluate performances based on distributional alignment---whether $f_{\text{new}}$ produces a well-calibrated answer distribution on the edit input (\emph{alignment}), generalizes this calibration to semantically equivalent inputs (\emph{soft generality}), and preserves the model's sensitivity to genuine distributional shifts (\emph{soft locality}) (Sec.~\ref{app:eval_model_calibration}).
Second, as a model editor, the one-shot calibration-enabled editor should still be able to perform conventional model editing, thus we evaluate performances based on answer generation accuracy---whether $f_{\text{new}}$ generates a correct answer on the edit input (\emph{reliability}) and on semantically equivalent inputs (\emph{generality}), and whether it preserves correct predictions on unrelated inputs (\emph{locality}) (Sec.~\ref{app:eval_model_editing}).

\subsubsection{One-shot Model Calibration Performances}\label{app:eval_model_calibration}

Given a pretrained model's predictions on a multi-answer dataset, the edit set $D_{\text{edit}}$ contains errors made by the model ($f(x)\notin\mathcal{Y}_{K}$); these are the wrong model generations to be edited. The remaining samples $D{u}$, on which the model was already correct ($f(x)\in\mathcal{Y}_{K}$), are the unrelated samples used to preserve the model's original predictions.

\paragraph{Alignment (Align).}
For each edit $e=(x,\mathcal{Y}_{K},\mathbf{p}_{K})$ in $D_{\text{edit}}$, the edited model should align its answer-level distribution with the human target distribution $\mathbf{p}_{K}$. We measure alignment using one minus the gated Jensen--Shannon distance $\mathrm{JS_{gate}}$ (Eq.~\ref{eq:js_gate}) between the human target distribution and the edited model's gated readout:
\[
\mathrm{Align}
=
\mathbb{E}_{e\in D_{\text{edit}}}
\left[
1-\mathrm{JS_{gate}}\!\left(
\mathbf{p}_{K}(\cdot\mid x),
\widehat{\mathbf{p}}_{K}(\cdot\mid x;\,f_{\text{new}})
\right)
\right].
\]
Higher is better, with $\mathrm{Align}=1$ indicating perfect distributional alignment.

\paragraph{Soft Generality (SGen).}
For an edit $e=(x,\mathcal{Y}_{K},\mathbf{p}_{K})$, let $R(x)$ denote semantically similar variants of the input $x$ that preserve the target distribution $\mathbf{p}_{K}$. The edited model should preserve the calibrated answer distribution on these related inputs:
\[
\mathrm{SGen}
=
\mathbb{E}_{e\in D_{\mathrm{edit}}}
\left[
1-\mathrm{JS_{gate}}\!\left(
\mathbf{p}_{K}(\cdot\mid x),
\widehat{\mathbf{p}}_{K}(\cdot\mid R(x);\,f_{\mathrm{new}})
\right)
\right].
\]
Higher is better, with $\mathrm{SGen}=1$ indicating perfect generalization of the target distribution.

\begin{figure}[t]
    \centering
    \includegraphics[width=1\linewidth]{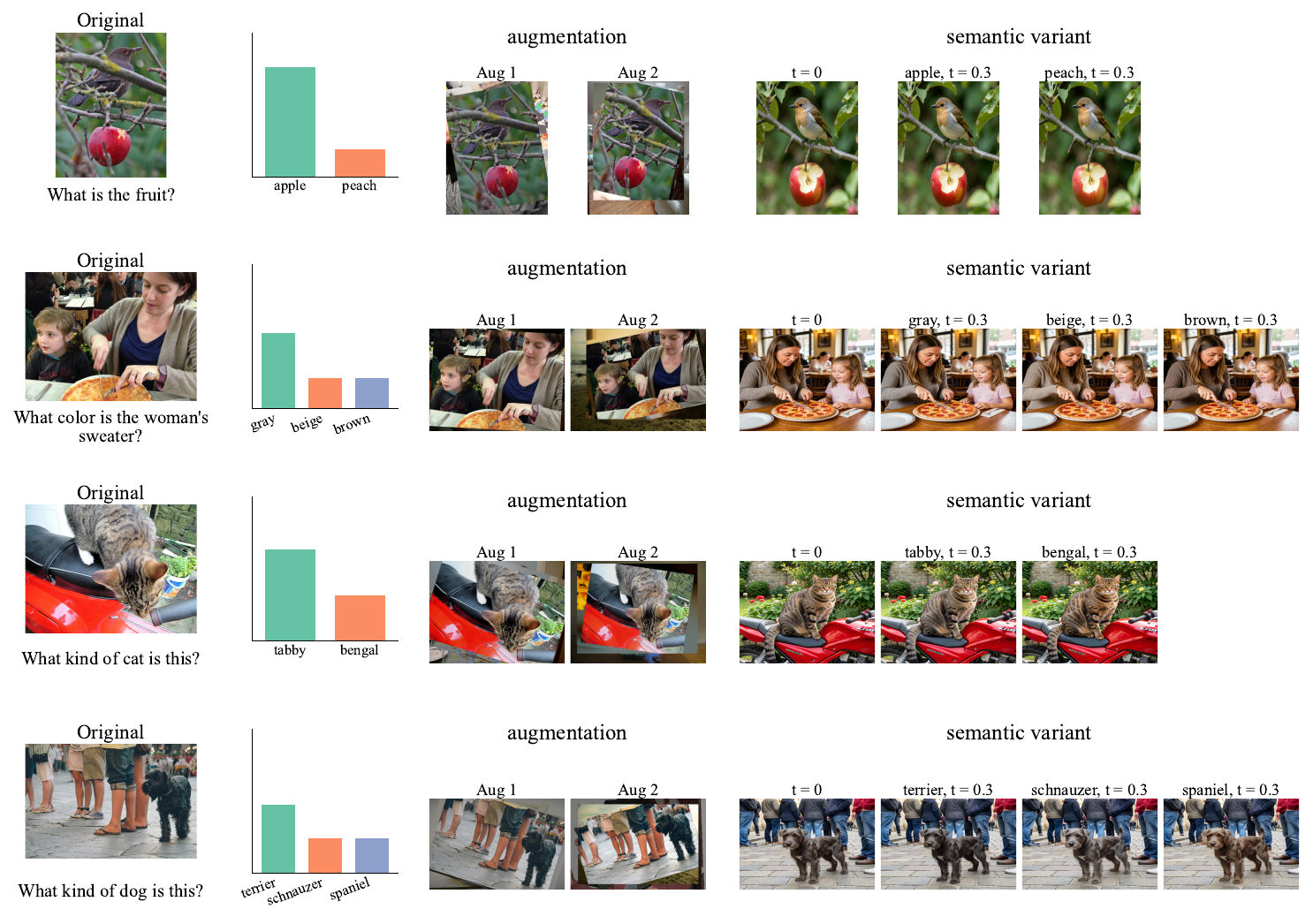}
    \caption{VQAv2 example edits and their SGen-I evaluation samples.}
    \label{fig:sgen_i}
\end{figure}

\begin{itemize}[leftmargin=*,itemsep=2pt]
    \item \textit{\textbf{Text Soft Generality (SGen-T).}}
    
    For an edit $e=(x,\mathcal{Y}_{K},\mathbf{p}_{K})$, let $x_t$ denote the text question in $x$. We evaluate whether the edited model follows the calibrated answer distribution for semantically equivalent rephrasings $R(x_t)$:
    \[
    \mathrm{SGen\text{-}T}
    =
    \mathbb{E}_{e\in D_{\mathrm{edit}}}
    \left[
    1-\mathrm{JS_{gate}}\!\left(
    \mathbf{p}_{K}(\cdot\mid x),
    \widehat{\mathbf{p}}_{K}(\cdot\mid R(x_t);\,f_{\mathrm{new}})
    \right)
    \right].
    \]
    
    For each edit, we use GPT-4o to generate 4 rephrased questions with the prompt following prior works~\citep{guo2025balancedit, qiu2026reasonedit}: \texttt{\footnotesize Rephrase the following question in 4 different ways while keeping the meaning. Question: {question}.} Each rephrased question is used directly (for LLMs) or paired with the original image (for VLMs). Because the semantic meaning is preserved under rephrasing, the original target distribution is also preserved, i.e., $\mathbf{p}_{K}(\cdot\mid R(x_t))=\mathbf{p}_{K}(\cdot\mid x)$.
    Examples of SGen-T samples are provided in Table~\ref{tab:sgen-t}.

    \item \textit{\textbf{Image Soft Generality (SGen-I, VLMs only).}}
    
    For an edit $e=(x,\mathcal{Y}_{K},\mathbf{p}_{K})$, let $x_i$ denote the image and $x_t$ its associated text question. We evaluate whether the edited model preserves the calibrated answer distribution for semantically and distributionally close image variants $R(x_i)$ while keeping the original text question unchanged:
    \[
    \mathrm{SGen\text{-}I}
    =
    \mathbb{E}_{e\in D_{\mathrm{edit}}}
    \left[
    1-\mathrm{JS_{gate}}\!\left(
    \mathbf{p}_{K}(\cdot\mid x),
    \widehat{\mathbf{p}}_{K}(\cdot\mid R(x_i),x_t;\,f_{\mathrm{new}})
    \right)
    \right].
    \]
    
    For each edit, we construct a set of distribution-preserving image variants in two ways:
    \begin{enumerate}
        \item Photometric perturbations (2 images). We apply a joint combination of Gaussian noise, brightness and contrast jitter, aspect-ratio stretching, and rotation to the original image. Exposed regions from stretching and rotation are filled with a randomly sampled real background scene from VQAv2~\citep{goyal2017making}. These photometric augmentations preserve the semantic content of the original image and are therefore associated with the same target distribution $\mathbf{p}_{K}$ as the edit.

        \item Generated semantically and distributionally similar variants.
        To evaluate generalization to visually novel but semantically and distributionally similar images, we generate new images that approximate the edit's target distribution $\mathbf{p}_{K}$, following the generated-image generality evaluation setting of prior works~\citep{cheng2023can, guo2025balancedit, qiu2026reasonedit}.
        For each edit, we use FLUX.2-klein-9B~\citep{flux-2-2025} to first generate a top-1 anchor image from a GPT-4o-generated caption of the original image, and then edit this anchor toward each of the top-$K$ answers, producing answer-specific images $\mathrm{img}_j,\;j=1,\ldots,K$ that share the same underlying visual context while depicting different answers.
        Let $\mathbf{z}_j$ denote the VAE latent embedding of $\mathrm{img}_j$. We construct a target-distribution-weighted blend
        \[
        \mathbf{z}_{\mathrm{blend}}
        =
        \sum_{j=1}^{K} p_j\,\mathbf{z}_j,
        \]
        and decode $\mathbf{z}_{\mathrm{blend}}$ to obtain a visually novel image whose target distribution approximates the original edit distribution $\mathbf{p}_{K}$.
        To reduce approximation error from using a single blended image, we additionally interpolate from $\mathbf{z}_{\mathrm{blend}}$ toward each answer-specific embedding $\mathbf{z}_j$ at mild strengths $t\in\{0.0,0.3\}$, and average distributional alignment across the resulting variants.
        We repeat the entire procedure with two random seeds to obtain two visually distinct sets of generated variants.
    \end{enumerate}
    Examples of SGen-I samples are provided in Fig.~\ref{fig:sgen_i}.

\end{itemize}

\begin{table*}[t]
\centering
\scriptsize
\setlength{\tabcolsep}{3pt}
\renewcommand{\arraystretch}{1.05}
\begin{tabular}{@{}lll@{}}
\hline
Question & Target & SGen \\
\hline
\begin{minipage}[t]{0.20\textwidth}
\raggedright\setlength{\parindent}{0pt}
Tell me something that's hard to do while driving in the snow.
\end{minipage} & \begin{minipage}[t]{0.30\textwidth}
\raggedright\setlength{\parindent}{0pt}
(stop~0.51, turn~0.49)
\end{minipage} & \begin{minipage}[t]{0.40\textwidth}
\raggedright\setlength{\parindent}{0pt}
What is something challenging to do when driving on snowy roads?
\end{minipage} \\
 &  & \begin{minipage}[t]{0.40\textwidth}
\raggedright\setlength{\parindent}{0pt}
Can you name a difficult task to perform while driving in snowy conditions?
\end{minipage} \\
 &  & \begin{minipage}[t]{0.40\textwidth}
\raggedright\setlength{\parindent}{0pt}
What's a hard thing to manage when operating a vehicle in the snow?
\end{minipage} \\
 &  & \begin{minipage}[t]{0.40\textwidth}
\raggedright\setlength{\parindent}{0pt}
What is one thing that's tricky to handle while driving in snowy weather?
\end{minipage} \\
\hline
\begin{minipage}[t]{0.20\textwidth}
\raggedright\setlength{\parindent}{0pt}
Name a holiday song with a fictional character.
\end{minipage} & \begin{minipage}[t]{0.30\textwidth}
\raggedright\setlength{\parindent}{0pt}
(rudolph~0.51, frosty the snowman~0.49)
\end{minipage} & \begin{minipage}[t]{0.40\textwidth}
\raggedright\setlength{\parindent}{0pt}
What is a holiday song that features a fictional character?
\end{minipage} \\
 &  & \begin{minipage}[t]{0.40\textwidth}
\raggedright\setlength{\parindent}{0pt}
Can you identify a holiday song about a fictional character?
\end{minipage} \\
 &  & \begin{minipage}[t]{0.40\textwidth}
\raggedright\setlength{\parindent}{0pt}
Which holiday song includes a fictional character?
\end{minipage} \\
 &  & \begin{minipage}[t]{0.40\textwidth}
\raggedright\setlength{\parindent}{0pt}
Provide the name of a holiday song that mentions a fictional character.
\end{minipage} \\
\hline
\begin{minipage}[t]{0.20\textwidth}
\raggedright\setlength{\parindent}{0pt}
Name a fruit that you can buy dried
\end{minipage} & \begin{minipage}[t]{0.30\textwidth}
\raggedright\setlength{\parindent}{0pt}
(grape~0.34, banana~0.33, apricot~0.33)
\end{minipage} & \begin{minipage}[t]{0.40\textwidth}
\raggedright\setlength{\parindent}{0pt}
What is a fruit that is available in dried form?
\end{minipage} \\
 &  & \begin{minipage}[t]{0.40\textwidth}
\raggedright\setlength{\parindent}{0pt}
Can you name a fruit that is sold dried?
\end{minipage} \\
 &  & \begin{minipage}[t]{0.40\textwidth}
\raggedright\setlength{\parindent}{0pt}
Which fruit can be purchased in a dried state?
\end{minipage} \\
 &  & \begin{minipage}[t]{0.40\textwidth}
\raggedright\setlength{\parindent}{0pt}
Identify a fruit that comes in a dried version.
\end{minipage} \\
\hline
\begin{minipage}[t]{0.20\textwidth}
\raggedright\setlength{\parindent}{0pt}
Name a food that it's hard to keep your face clean while eating.
\end{minipage} & \begin{minipage}[t]{0.30\textwidth}
\raggedright\setlength{\parindent}{0pt}
(ice cream~0.34, ribs~0.33, spaghetti~0.33)
\end{minipage} & \begin{minipage}[t]{0.40\textwidth}
\raggedright\setlength{\parindent}{0pt}
What is a food that makes it difficult to keep your face clean while eating?
\end{minipage} \\
 &  & \begin{minipage}[t]{0.40\textwidth}
\raggedright\setlength{\parindent}{0pt}
Can you name a food that's challenging to eat without getting your face messy?
\end{minipage} \\
 &  & \begin{minipage}[t]{0.40\textwidth}
\raggedright\setlength{\parindent}{0pt}
What type of food is hard to eat without messing up your face?
\end{minipage} \\
 &  & \begin{minipage}[t]{0.40\textwidth}
\raggedright\setlength{\parindent}{0pt}
Identify a food that often causes your face to get dirty while eating.
\end{minipage} \\
\hline
\end{tabular}
\caption{\small ProtoQA example edits and their text soft generality (SGen) evaluation samples.}
\label{tab:sgen-t}
\end{table*}

\paragraph{Soft Locality (SLoc).}
For an edit $e=(x,\mathcal{Y}_{K},\mathbf{p}_{K})$, let $R_{\mathrm{loc}}(x)$ denote semantically similar variants of $x$ whose distributions over the target answers differ from the original edit's target distribution $\mathbf{p}_{K}$. The edited model should preserve these genuine distributional variations rather than over-generalizing the edit distribution to all related inputs. We measure soft locality as
\[
\mathrm{SLoc}
=
\mathbb{E}_{e\in D_{\mathrm{edit}}}
\left[
1-\mathrm{JS_{gate}}\!\left(
\mathbf{p}_{K}(\cdot\mid R_{\mathrm{loc}}(x)),
\widehat{\mathbf{p}}_{K}(\cdot\mid R_{\mathrm{loc}}(x);\,f_{\mathrm{new}})
\right)
\right].
\]
Higher is better, with $\mathrm{SLoc}=1$ indicating perfect preservation of the target distributional variations within the semantic neighborhood of an edit.

\begin{figure}[t]
    \centering
    \includegraphics[width=0.8\linewidth]{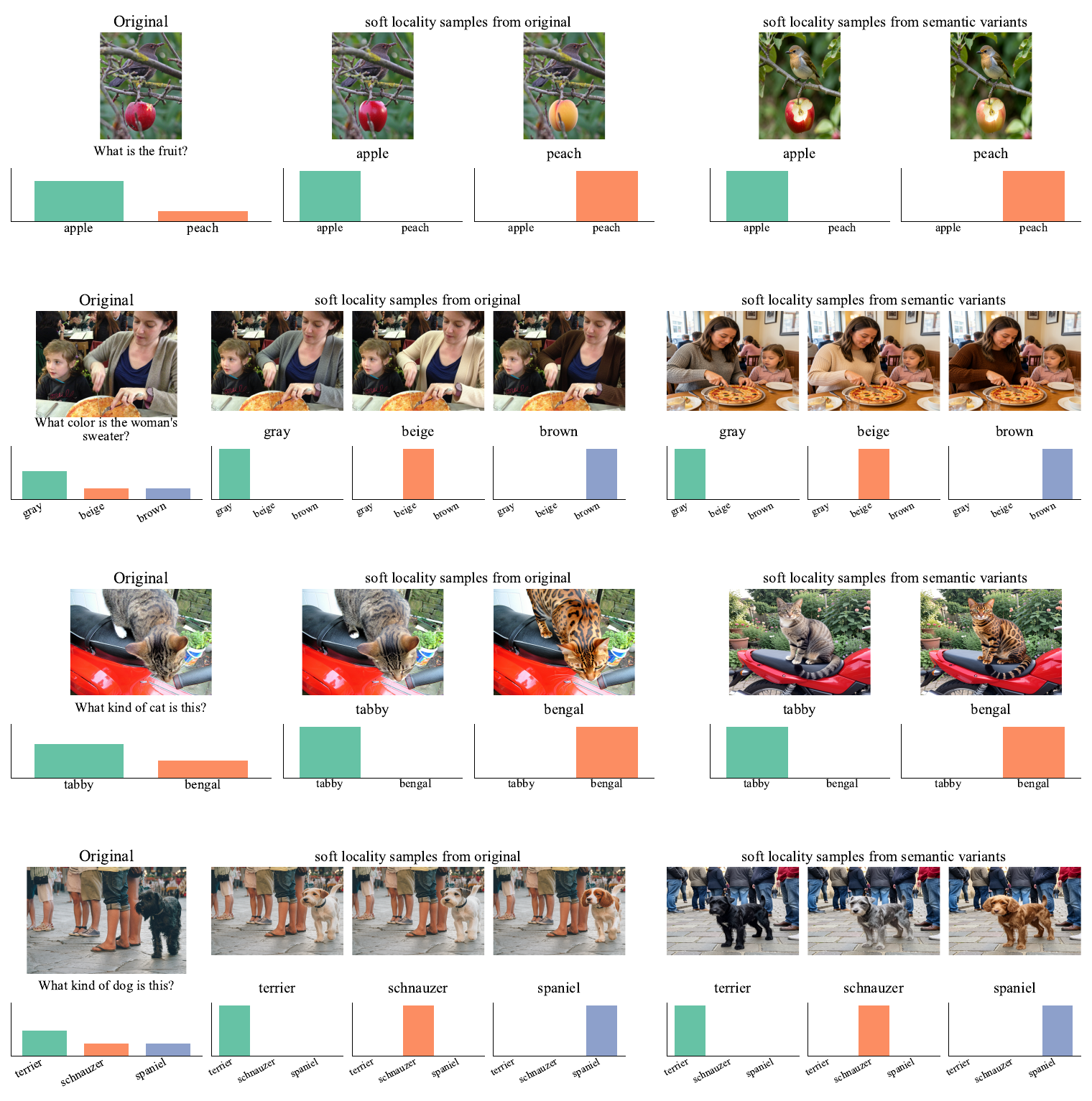}
    \caption{VQAv2 example edits and their SLoc-I evaluation samples.}
    \label{fig:sloc_i}
\end{figure}

\begin{itemize}[leftmargin=*,itemsep=4pt]
    \item \textit{\textbf{Text Soft Locality (SLoc-T, LLMs only).}}

    For an edit $e=(x_t,\mathcal{Y}_{K},\mathbf{p}_{K})$, we construct a semantically related variant $x_k$ for each target answer $y_k\in\mathcal{Y}_{K}$ by adding textual context that unambiguously favors $y_k$. The edited model should therefore shift its answer distribution from the original multi-answer target $\mathbf{p}_{K}$ to the one-hot target $\mathbf{e}_k$. We measure text soft locality as
    \[
    \mathrm{SLoc\text{-}T}
    =
    \mathbb{E}_{e\in D_{\mathrm{edit}}}
    \left[
    \frac{1}{K}
    \sum_{k=1}^{K}
    \left(
    1-\mathrm{JS_{gate}}\!\left(
    \mathbf{e}_k,
    \widehat{\mathbf{p}}_{K}(\cdot\mid x_k;\,f_{\mathrm{new}})
    \right)
    \right)
    \right].
    \]
    
    For each edit, we construct the distributional text locality samples by adding short textual context to the original question. Specifically, for each edit $e=(x_t,\mathcal{Y}_{K},\mathbf{p}_{K})$ and each target answer $y_k\in\mathcal{Y}_{K}$, we use GPT-4o to generate a brief premise $c_k$ that makes $y_k$ unambiguously the most plausible answer when prepended to the question, without explicitly mentioning the answer itself.
    The generation prompt follows the general form:
    \begin{quote}\small
    \texttt{Given the question, target answer, and other plausible answers, write a brief natural premise that makes the target answer clearly the most likely answer without directly revealing it.}
    \end{quote}
    
    The resulting input $x_k=c_k\oplus x_t$ is assigned the one-hot target distribution $\mathbf{e}_k$. Thus, each premise induces a controlled distributional shift from the original multi-answer distribution $\mathbf{p}_{K}$ toward a single target answer $y_k$. 
    % Unlike the image modality, where we construct intermediate shifts at multiple strengths, the text field uses a single fully disambiguated variant for each answer. 
    We generate one such variant for each of the $K$ target answers.
    Examples of SLoc-T samples are provided in Table~\ref{tab:sloc-t}.

    \item \textit{\textbf{Image Soft Locality (SLoc-I, VLMs only).}}

    For a VLM edit $e=(x,\mathcal{Y}_{K},\mathbf{p}_{K})$, let $\mathrm{img}_{k}$ denote an image variant whose visual evidence is changed toward answer $y_k$. We assign it a one-hot target distribution $\mathbf{e}_k$, where $y_k$ receives probability 1. We measure image soft locality by how well the edited model still follows these distributional shifts:
    \[
    \mathrm{SLoc\text{-}I}
    =
    \mathbb{E}_{e\in D_{\mathrm{edit}}}
    \left[
    \frac{1}{K}
    \sum_{k=1}^{K}
    \frac{1}{|\mathcal{I}_{e,k}|}
    \sum_{\mathrm{img}\in\mathcal{I}_{e,k}}
    \left[
    1-\mathrm{JS_{gate}}\!\left(
    \mathbf{e}_k,
    \widehat{\mathbf{p}}_{K}
    (\cdot\mid \mathrm{img},x_t;\,f_{\mathrm{new}})
    \right)
    \right]
    \right],
    \]
    where $\mathcal{I}_{e,k}$ is the set of validated image variants depicting $y_k$, and $x_t$ is the original text question. Higher is better. We construct the image variants in two complementary ways.
    
    \begin{enumerate}
        \item \textit{Editing the original image toward each answer.}
        We first construct distributionally shifted variants while preserving the original scene and visual context. For each edit, we use FLUX.2-klein-9B~\citep{flux-2-2025} to edit the original image toward each target answer $y_k$, producing an unambiguous image $\mathrm{img}_k$ in which $y_k$ is clearly visually present.
        
        \item \textit{Editing generated semantically similar images toward each answer.}
        We additionally construct the same distributional shifts on unseen but semantically related images. For each edit, we first use FLUX.2-klein-9B to generate a top-1 anchor image from a GPT-4o-generated caption of the original image, and then edit this anchor toward each of the top-$K$ answers, producing answer-specific images $\mathrm{img}_j,\;j=1,\ldots,K$ that share the same underlying visual structure. We repeat the procedure with two random seeds to obtain two visually distinct sets of variants.
    \end{enumerate}
    
    For both procedures, we use GPT-4o to verify that each target label is visually recognizable as detailed in Appendix~\ref{app:experiment_gen_image_validation}. We assign a one-hot target distribution to each answer-specific image, following prior VLM editing works that assign an unambiguous target label to each generated image depicting that label~\citep{cheng2023can, zhang2025mc, guo2025balancedit, qiu2026reasonedit}.
    Examples of SLoc-I samples are provided in Fig.~\ref{fig:sloc_i}.

\end{itemize}

\begin{table*}[t]
\centering
\scriptsize
\setlength{\tabcolsep}{3pt}
\renewcommand{\arraystretch}{1.05}
\begin{tabular}{@{}lll@{}}
\hline
Question & Target & SLoc: Answer-specific premises (italic) prepended to question \\
\hline
\begin{minipage}[t]{0.20\textwidth}
\raggedright\setlength{\parindent}{0pt}
Tell me something that's hard to do while driving in the snow.
\end{minipage} & \begin{minipage}[t]{0.30\textwidth}
\raggedright\setlength{\parindent}{0pt}
(stop~0.51, turn~0.49)
\end{minipage} &  \\
 & \begin{minipage}[t]{0.30\textwidth}
\raggedright\setlength{\parindent}{0pt}
stop
\end{minipage} & \begin{minipage}[t]{0.40\textwidth}
\raggedright\setlength{\parindent}{0pt}
\textit{In snowy conditions, vehicles often lose traction, making it difficult to avoid sliding forward when attempting to slow down.} Tell me something that's hard to do while driving in the snow.
\end{minipage} \\
 & \begin{minipage}[t]{0.30\textwidth}
\raggedright\setlength{\parindent}{0pt}
turn
\end{minipage} & \begin{minipage}[t]{0.40\textwidth}
\raggedright\setlength{\parindent}{0pt}
\textit{When roads are icy, maintaining control during directional changes becomes particularly challenging.} Tell me something that's hard to do while driving in the snow.
\end{minipage} \\
\hline
\begin{minipage}[t]{0.20\textwidth}
\raggedright\setlength{\parindent}{0pt}
Name a holiday song with a fictional character.
\end{minipage} & \begin{minipage}[t]{0.30\textwidth}
\raggedright\setlength{\parindent}{0pt}
(rudolph~0.51, frosty the snowman~0.49)
\end{minipage} &  \\
 & \begin{minipage}[t]{0.30\textwidth}
\raggedright\setlength{\parindent}{0pt}
rudolph
\end{minipage} & \begin{minipage}[t]{0.40\textwidth}
\raggedright\setlength{\parindent}{0pt}
\textit{This song tells the story of a reindeer who helps guide Santa's sleigh on a foggy Christmas Eve.} Name a holiday song with a fictional character.
\end{minipage} \\
 & \begin{minipage}[t]{0.30\textwidth}
\raggedright\setlength{\parindent}{0pt}
frosty the snowman
\end{minipage} & \begin{minipage}[t]{0.40\textwidth}
\raggedright\setlength{\parindent}{0pt}
\textit{This song features a magical figure who comes to life with the help of a special hat.} Name a holiday song with a fictional character.
\end{minipage} \\
\hline
\begin{minipage}[t]{0.20\textwidth}
\raggedright\setlength{\parindent}{0pt}
Name a fruit that you can buy dried
\end{minipage} & \begin{minipage}[t]{0.30\textwidth}
\raggedright\setlength{\parindent}{0pt}
(grape~0.34, banana~0.33, apricot~0.33)
\end{minipage} &  \\
 & \begin{minipage}[t]{0.30\textwidth}
\raggedright\setlength{\parindent}{0pt}
grape
\end{minipage} & \begin{minipage}[t]{0.40\textwidth}
\raggedright\setlength{\parindent}{0pt}
\textit{This fruit is commonly dried to make one of the most popular snack items worldwide.} Name a fruit that you can buy dried
\end{minipage} \\
 & \begin{minipage}[t]{0.30\textwidth}
\raggedright\setlength{\parindent}{0pt}
banana
\end{minipage} & \begin{minipage}[t]{0.40\textwidth}
\raggedright\setlength{\parindent}{0pt}
\textit{This fruit is commonly sliced into thin pieces and dried into chips, often used as a snack or topping.} Name a fruit that you can buy dried
\end{minipage} \\
 & \begin{minipage}[t]{0.30\textwidth}
\raggedright\setlength{\parindent}{0pt}
apricot
\end{minipage} & \begin{minipage}[t]{0.40\textwidth}
\raggedright\setlength{\parindent}{0pt}
\textit{This fruit is commonly dried and retains a sweet, tangy flavor, often used in trail mixes and cooking.} Name a fruit that you can buy dried
\end{minipage} \\
\hline
\begin{minipage}[t]{0.20\textwidth}
\raggedright\setlength{\parindent}{0pt}
Name a food that it's hard to keep your face clean while eating.
\end{minipage} & \begin{minipage}[t]{0.30\textwidth}
\raggedright\setlength{\parindent}{0pt}
(ice cream~0.34, ribs~0.33, spaghetti~0.33)
\end{minipage} &  \\
 & \begin{minipage}[t]{0.30\textwidth}
\raggedright\setlength{\parindent}{0pt}
ice cream
\end{minipage} & \begin{minipage}[t]{0.40\textwidth}
\raggedright\setlength{\parindent}{0pt}
\textit{This food is often eaten on a cone and tends to melt quickly, especially in warm weather.} Name a food that it's hard to keep your face clean while eating.
\end{minipage} \\
 & \begin{minipage}[t]{0.30\textwidth}
\raggedright\setlength{\parindent}{0pt}
ribs
\end{minipage} & \begin{minipage}[t]{0.40\textwidth}
\raggedright\setlength{\parindent}{0pt}
\textit{This food is often coated in a sticky sauce that gets all over your hands and face while eating.} Name a food that it's hard to keep your face clean while eating.
\end{minipage} \\
 & \begin{minipage}[t]{0.30\textwidth}
\raggedright\setlength{\parindent}{0pt}
spaghetti
\end{minipage} & \begin{minipage}[t]{0.40\textwidth}
\raggedright\setlength{\parindent}{0pt}
\textit{This dish is known for its long strands often coated in a saucy topping that can easily splatter.} Name a food that it's hard to keep your face clean while eating.
\end{minipage} \\
\hline
\end{tabular}
\caption{\small ProtoQA example edits and their text soft locality (SLoc) evaluation samples.}
\label{tab:sloc-t}
\end{table*}

\subsubsection{One-shot Model Editing Performances}\label{app:eval_model_editing}

\paragraph{Reliability (Acc).}
The edited model should output a correct answer from the target answers for each edit $e=(x,\mathcal{Y}_{K},\mathbf{p}_{K})$ in $D_{\text{edit}}$:
\[
\mathrm{Acc} = \mathbb{E}_{e\in D_{\text{edit}}}\,\mathbf{1}\!\bigl\{f_{\text{new}}(x)\in\mathcal{Y}_{K}\bigr\}.
\]

\paragraph{Locality (Loc).}
The edited model should preserve its original correct prediction on unrelated samples $D_u$:
\[
\mathrm{Loc} = \mathbb{E}_{x\in D_{u}}\,\mathbf{1}\!\bigl\{f_{\text{new}}(x)=f(x)\bigr\}.
\]

\paragraph{Generality (Gen).}
For an edit $e=(x,\mathcal{Y}_{K},\mathbf{p}_{K})$, let $R(x)$ denote semantically similar variants of the input $x$ for which the edited model should generate the same target answers $\mathcal{Y}_{K}$ according to the target distribution $\mathbf{p}_{K}$. The edited model should output a correct answer from the target set:
\[
\mathrm{Gen} = \mathbb{E}_{e\in D_{\text{edit}}}\mathbf{1}\!\bigl\{f_{\text{new}}(R(x))\in\mathcal{Y}_{K}\bigr\}.
\]

\begin{itemize}[leftmargin=*,itemsep=2pt]
    \item \emph{\textbf{Text generality (Gen-T).}} 
    For an edit $e=(x,\mathcal{Y}_{K},\mathbf{p}_{K})$, let $x_t$ denote a text question in $x$. We evaluate whether the edited model generates a correct answer for rephrased questions $R(x_t)$:
    \[
    \mathrm{Gen\text{-}T} 
    = 
    \mathbb{E}_{e\in D_{\text{edit}}}
    \mathbf{1}\!\bigl\{f_{\text{new}}(R(x_t))\in\mathcal{Y}_{K}\bigr\}.
    \]
    We use the same rephrased questions $R(x_t)$ constructed for SGen-T, but evaluate answer-generation accuracy rather than distributional alignment.

    \item \emph{\textbf{Image generality (Gen-I, VLMs only).}} For an edit $e=(x,\mathcal{Y}_{K},\mathbf{p}_{K})$, let $x_i$ denote an image and $x_t$ denote its associated text question in the input $x$. We evaluate whether the edited model generates a correct answer for related semantic-preserving images $R(x_i)$ that correspond to the same target answers $\mathcal{Y}_{K}$ following $\mathbf{p}_{K}$, while keeping the original text question unchanged:
    \[
    \mathrm{Gen\text{-}I}
    =
    \mathbb{E}_{e\in D_{\text{edit}}}
    \mathbf{1}\!\bigl\{f_{\text{new}}(R(x_i),x_t)\in\mathcal{Y}_{K}\bigr\}.
    \]
    We use the same image variants $R(x_i)$ constructed for SGen-I, but evaluate answer-generation accuracy rather than distributional alignment.

\end{itemize}

\subsection{Validation of Generated Images}\label{app:experiment_gen_image_validation}

We follow prior work~\citep{cheng2023can, zhang2025mc, guo2025balancedit, qiu2026reasonedit} to generate images for VLM-editing evaluation metrics.
We validate every generated answer-specific image with GPT-4o.
% , before interpolating answer-specific images to obtain intermediate-strength evaluation images.
%
Specifically, for each image we pose two binary questions to GPT-4o:
(1)~whether the image depicts the intended answer label (\emph{e.g.}, ``Does this image show a \{label\}?''), and
(2)~whether the image contains any stamped or watermark-style text.
An image is marked as failed if the target label is not depicted in the image, or if the answer label is leaked as watermark text onto the image.
We repeat this generation-and-validation cycle up to three times with different random seeds, and discard samples that have any image that still fails after three attempts.
For VQAv2, out of 78{,}872 generated answer-specific images, 16{,}992 failed the content check (target label not shown) and 12{,}207 failed the watermark check (answer text leaked onto the image).
After discarding all samples with at least one failed image, a total of 19{,}186 VQAv2 samples are retained with valid evaluation images.
This constitutes a sufficiently large evaluation set for benchmarking editing performance across all VLM editing methods.

\newpage
\section{One-shot Calibrated Generation}\label{app:calib_def}

One-shot calibration aims to align a pretrained model's answer-level distribution with a human-provided target distribution using a single edit, while generalizing this calibration to related inputs and preserving genuine distributional shifts. SoftEdit enables calibrated answer generation through three complementary approaches: (1) fast deterministic readout of complete answers and their estimated probabilities (Sec.~\ref{app:calib_def_readout}), (2) autoregressive full-sequence probabilities (Sec.~\ref{app:calib_def_full_seq}), and (3) stochastic sampling of complete answers (Sec.~\ref{app:calib_def_MC}). In the following subsections, we examine these approaches, their practical applications, and empirical evaluations, followed by the connection between one-shot calibration and conventional model calibration.

\subsection{Fast Deterministic Answer Distribution Readout}
\label{app:calib_def_readout}

The edited model enables fast deterministic readout of multiple complete answers and their estimated probabilities. Specifically, we obtain the first-token probability distribution and use greedy decoding to complete answers from the user-specified top-$K$ most probable candidate first tokens. The corresponding first-token probabilities serve as approximations of complete-answer probabilities, allowing users to directly obtain multiple answers with their estimated likelihoods or present them in a desired format using a simple text template. Our fast answer-level probability estimator (Appendix Sec.~\ref{app:gated_readout}) follows this deterministic generation approach, combining first-token probabilities with continuation validation to ensure that each evaluated answer can be generated completely. We adopt this readout in our main experiments to consistently compare the one-shot calibration performance of SoftEdit and other adapted editors under the same generation and evaluation procedure.

\subsection{Full-Sequence Answer Probability}
\label{app:calib_def_full_seq}

For an autoregressive model, the probability of generating a complete answer is the product of its conditional token probabilities. Given an answer $y_k=(y_{k,1},\ldots,y_{k,L_k})$, including the end-of-sequence token, its full-sequence probability is
\begin{equation*}
p_\theta(y_k\mid x)
=
p_\theta(y_{k,1}\mid x)
\prod_{t=2}^{L_k}
p_\theta(y_{k,t}\mid x,y_{k,<t}).
\end{equation*}
Under ancestral sampling at temperature 1, this represents the exact probability of generating the complete answer $y_k$. % One-shot calibration aims to align this probability with the human-provided target probability $p_k$.
SoftEdit encourages full-sequence probability calibration through the two components of its alignment loss:
\begin{equation*}
\mathcal{L}_{\mathrm{align}}
=
\underbrace{
D_{\mathrm{KL}}\left(
\mathbf{p}_K \,\|\, \widehat{\mathbf{p}}_K(h_0)
\right)
}_{\text{first-token alignment}}
+
\underbrace{
\frac{1}{K}\sum_{k=1}^{K}
\left[
-\frac{1}{L_k-1}
\sum_{t=2}^{L_k}
\log p_\theta(y_{k,t}\mid x,y_{k,<t})
\right]
}_{\text{full-answer continuation}}.
\end{equation*}
The first-token alignment term encourages
$p_\theta(y_{k,1}\mid x)\approx p_k$,
while the continuation term maximizes the likelihood of the remaining tokens, encouraging their conditional probabilities toward 1. Therefore, these objectives directly encourage
\begin{equation*}
p_\theta(y_k\mid x)
=
\underbrace{p_\theta(y_{k,1}\mid x)}_{\approx p_k}
\underbrace{
\prod_{t=2}^{L_k}
p_\theta(y_{k,t}\mid x,y_{k,<t})
}_{\approx 1}
\approx p_k.
\end{equation*}
% This factorization applies directly when the target answers have distinct first tokens; answers sharing a first token are aggregated for first-token alignment.

In practice however, unlike the fast deterministic readout, computing full-sequence probabilities requires users to provide the complete text of each candidate answer to evaluate its probability. This is less practical for obtaining a distribution over multiple answers in open-ended generation, as all candidate answers must be specified in advance. Therefore, we adopt the fast deterministic readout for our main evaluation and benchmarking of model editors.

Empirically, full-sequence probabilities provide a direct evaluation of the edited model's autoregressive answer probabilities. In addition to the fast readout, we evaluate all editors using full-sequence probabilities to measure alignment, soft generality, and soft locality on the same evaluation samples as the main experiments (Table~\ref{tab:main_results}). As shown in Table~\ref{tab:seq_prob}, SoftEdit achieves full-sequence calibration performance closely matching its fast readout results (a median[min, max] absolute score difference of 0.01[0.00, 0.04]), while consistently outperforming the adapted editors across all six backbone models when evaluated using full-sequence probabilities.

\begin{table*}[t]
\centering

% (1) VLM Editing on VQAv2 (19,186 samples)
\resizebox{\textwidth}{!}{%
\setlength{\tabcolsep}{3pt}%
\begin{tabular}{l>{\centering\arraybackslash }b{4.0em}>{\centering\arraybackslash }b{4.0em}>{\centering\arraybackslash }b{4.0em}>{\centering\arraybackslash }b{4.0em}>{\centering\arraybackslash }b{4.0em}>{\centering\arraybackslash }b{4.0em}>{\centering\arraybackslash }b{4.0em}>{\centering\arraybackslash }b{4.0em}>{\centering\arraybackslash }b{4.0em}>{\centering\arraybackslash }b{4.0em}>{\centering\arraybackslash }b{4.0em}>{\centering\arraybackslash }b{4.0em}}
\toprule
 \multirow{3.5}{*}{\textbf{Editor}} & \multicolumn{4}{c}{\textbf{LLaVA-1.5-7B (4,436 errors)}} & \multicolumn{4}{c}{\textbf{InstructBLIP-7B (4,625 errors)}} & \multicolumn{4}{c}{\textbf{Qwen3-4B (4,157 errors)}} \\
\cmidrule(lr){2-5} \cmidrule(lr){6-9} \cmidrule(lr){10-13}
 & \footnotesize Align & \footnotesize SGen-I & \footnotesize SGen-T & \footnotesize SLoc & \footnotesize Align & \footnotesize SGen-I & \footnotesize SGen-T & \footnotesize SLoc & \footnotesize Align & \footnotesize SGen-I & \footnotesize SGen-T & \footnotesize SLoc \\
\midrule
Unedited & 0.08 & 0.16 & 0.08 & 0.27 & 0.17 & 0.23 & 0.17 & 0.30 & 0.11 & 0.22 & 0.12 & 0.49 \\
\midrule
FT & 0.51 & 0.47 & 0.47 & 0.40 & 0.51 & 0.47 & 0.48 & 0.43 & 0.52 & 0.45 & 0.49 & 0.41 \\
MEND & 0.11 & 0.14 & 0.11 & 0.15 & 0.07 & 0.08 & 0.07 & 0.07 & 0.08 & 0.09 & 0.08 & 0.09 \\
GRACE & 0.52 & 0.24 & 0.32 & 0.17 & 0.52 & 0.37 & 0.40 & 0.44 & 0.53 & 0.26 & 0.31 & 0.53 \\
BalancEdit & 0.23 & 0.19 & 0.18 & 0.28 & 0.22 & 0.26 & 0.20 & 0.33 & 0.18 & 0.23 & 0.16 & 0.40 \\
ReasonEdit & 0.40 & 0.40 & 0.37 & 0.34 & 0.39 & 0.41 & 0.38 & 0.34 & 0.53 & 0.52 & 0.50 & 0.53 \\
\midrule
FT-Soft & 0.73 & 0.58 & 0.61 & 0.42 & 0.74 & 0.58 & 0.64 & 0.47 & 0.78 & 0.55 & 0.66 & 0.41 \\
MEND-Soft & 0.11 & 0.13 & 0.11 & 0.13 & 0.13 & 0.16 & 0.13 & 0.16 & 0.06 & 0.07 & 0.06 & 0.06 \\
GRACE-Soft & 0.90 & 0.38 & 0.53 & 0.19 & 0.87 & 0.48 & 0.59 & 0.47 & \textbf{0.90} & 0.28 & 0.45 & 0.53 \\
BalancEdit-Soft & 0.22 & 0.19 & 0.17 & 0.27 & 0.23 & 0.25 & 0.20 & 0.32 & 0.20 & 0.25 & 0.17 & 0.43 \\
ReasonEdit-Soft & 0.51 & 0.50 & 0.49 & 0.38 & 0.50 & 0.52 & 0.47 & 0.40 & 0.56 & 0.54 & 0.54 & 0.62 \\
\midrule
\textbf{SoftEdit} & \textbf{0.97} & \textbf{0.74} & \textbf{0.92} & \textbf{0.61} & \textbf{0.97} & \textbf{0.80} & \textbf{0.87} & \textbf{0.64} & 0.88 & \textbf{0.67} & \textbf{0.81} & \textbf{0.64} \\
\bottomrule
\end{tabular}
}\par
\vspace{1.5mm}
{\small (1) VLM Editing on VQAv2 (19,186 samples)}\par
\vspace{1mm}

% (2) LLM Editing on ProtoQA (9,762 samples)
\resizebox{0.7500\textwidth}{!}{%
\setlength{\tabcolsep}{3pt}%
\begin{tabular}{l>{\centering\arraybackslash }b{4.0em}>{\centering\arraybackslash }b{4.0em}>{\centering\arraybackslash }b{4.0em}>{\centering\arraybackslash }b{4.0em}>{\centering\arraybackslash }b{4.0em}>{\centering\arraybackslash }b{4.0em}>{\centering\arraybackslash }b{4.0em}>{\centering\arraybackslash }b{4.0em}>{\centering\arraybackslash }b{4.0em}}
\toprule
 \multirow{3.5}{*}{\textbf{Editor}} & \multicolumn{3}{c}{\textbf{Llama-3.1-8B (8,326 errors)}} & \multicolumn{3}{c}{\textbf{Mistral-7B (7,782 errors)}} & \multicolumn{3}{c}{\textbf{Qwen2.5-7B (7,207 errors)}} \\
\cmidrule(lr){2-4} \cmidrule(lr){5-7} \cmidrule(lr){8-10}
 & \footnotesize Align & \footnotesize SGen & \footnotesize SLoc & \footnotesize Align & \footnotesize SGen & \footnotesize SLoc & \footnotesize Align & \footnotesize SGen & \footnotesize SLoc \\
\midrule
Unedited & 0.00 & 0.00 & 0.00 & 0.00 & 0.00 & 0.00 & 0.01 & 0.00 & 0.02 \\
\midrule
FT & 0.46 & 0.39 & 0.27 & 0.45 & 0.33 & 0.23 & 0.46 & 0.38 & 0.24 \\
MEND & 0.11 & 0.09 & 0.07 & 0.03 & 0.03 & 0.02 & 0.03 & 0.03 & 0.02 \\
GRACE & 0.48 & 0.24 & 0.06 & 0.48 & 0.02 & 0.00 & 0.47 & 0.20 & 0.04 \\
IKE & 0.08 & 0.07 & 0.09 & 0.08 & 0.06 & 0.08 & 0.36 & 0.30 & 0.23 \\
AlphaEdit & 0.26 & 0.16 & 0.17 & 0.38 & 0.21 & 0.22 & 0.21 & 0.07 & 0.15 \\
\midrule
FT-Soft & 0.78 & 0.53 & 0.36 & 0.73 & 0.44 & 0.30 & 0.78 & 0.50 & 0.31 \\
MEND-Soft & 0.15 & 0.12 & 0.10 & 0.03 & 0.03 & 0.06 & 0.04 & 0.04 & 0.06 \\
GRACE-Soft & 0.91 & 0.39 & 0.07 & 0.86 & 0.03 & 0.00 & 0.80 & 0.26 & 0.06 \\
IKE-Soft & 0.07 & 0.07 & 0.10 & 0.08 & 0.08 & 0.07 & 0.34 & 0.30 & 0.35 \\
AlphaEdit-Soft & 0.22 & 0.15 & 0.11 & 0.38 & 0.18 & 0.19 & 0.28 & 0.08 & 0.18 \\
\midrule
\textbf{SoftEdit} & \textbf{0.94} & \textbf{0.72} & \textbf{0.47} & \textbf{0.92} & \textbf{0.68} & \textbf{0.51} & \textbf{0.95} & \textbf{0.71} & \textbf{0.48} \\
\bottomrule
\end{tabular}
}\par
\vspace{1.5mm}
{\small (2) LLM Editing on ProtoQA (9,762 samples)}\par
\vspace{-2mm}
\caption{Full-sequence answer probability ($1-\mathrm{JS}$) on the same edits as Table~\ref{tab:main_results}. SoftEdit's full-sequence probability calibration is nearly identical to its fast readout probability calibration.}
\label{tab:seq_prob}
\end{table*}

\subsection{Monte Carlo Sampling of Calibrated Answers}
\label{app:calib_def_MC}

As discussed in Sec.~\ref{app:gated_readout}, a third and gold-standard approach to estimating an edited model's answer-level distribution is Monte Carlo sampling~\citep{lee2023can,sorensen2022information}. However, this requires sampling and decoding $N$ complete responses per query, where larger $N$ yields more reliable probability estimates but incurs greater computational cost, making it prohibitively expensive for large-scale inference and evaluation. 

\begin{figure}[t]
    \centering
    \includegraphics[width=0.8\linewidth]{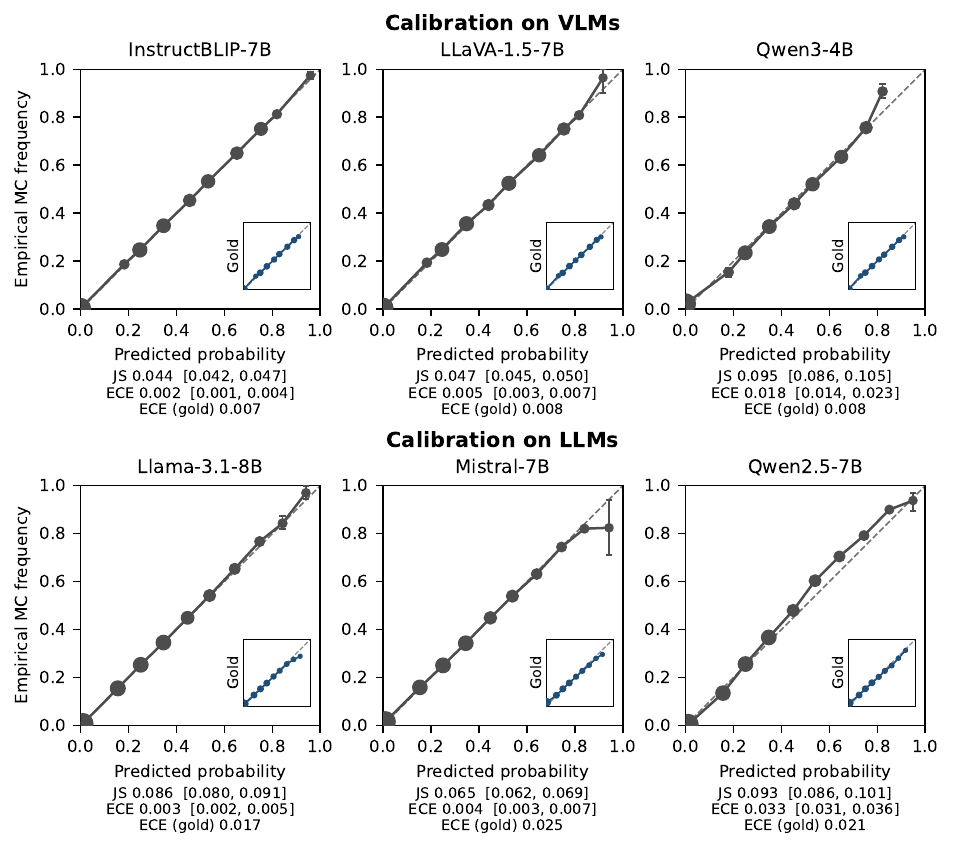}
    \caption{Monte Carlo validation of fast readout probabilities across VLMs on VQAv2 (top) and LLMs on ProtoQA (bottom), using 1,000 edits and 100 stochastic generations per edit. Main plots compare fast readout probabilities with empirical Monte Carlo frequencies, while insets compare them with human target probabilities.}
    \label{fig:mc_calib}
\end{figure}

To further validate the agreement between SoftEdit's fast readout probabilities and stochastic generation, we conduct a Monte Carlo sampling study. For each backbone model, we use 1,000 edits, perform one-shot calibration using SoftEdit, and compute the resulting fast readout distribution over the answer support. We then stochastically generate 100 answers per edit using \textit{ancestral sampling} at temperature 1, yielding an empirical answer-frequency distribution for each edit. We quantify agreement using the mean per-edit Jensen--Shannon (JS) distance (lower is better) between the predicted and Monte Carlo distributions. We also report the expected calibration error (ECE), the weighted average absolute difference between predicted probabilities and empirical frequencies across probability bins. For both metrics, we report 95\% confidence intervals obtained by bootstrapping edits. In the calibration diagrams (Fig.~\ref{fig:mc_calib}), we pool all edit--answer pairs into 10 equal-width bins by predicted probability, weighting each edit equally. The insets additionally compare predicted probabilities with human target distributions. Across all six models, the empirical Monte Carlo frequencies closely follow the predicted probabilities, with most calibration curves near the diagonal and low JS distances and ECE scores.

We further evaluate how well our deterministic gated readout approximates answer distributions under stochastic generation. As shown in Table~\ref{tab:mc_align}, we compare their alignment with human target distributions on 1,000 edits per backbone, using 100 sampled generations per edit. The two approaches show close agreement across all six models, with absolute differences of 0.002--0.024 overall and 0.001--0.018 for multi-token answers. For answers sharing the same first token, both approaches show lower alignment. These cases represent only 11--61 edits per backbone among the 1000 samples, and we further discuss this limitation in the Limitations section.

\begin{table*}[t]
\centering
\resizebox{\textwidth}{!}{%
\setlength{\tabcolsep}{3pt}%
\begin{tabular}{l>{\centering\arraybackslash}b{4.6em}>{\centering\arraybackslash}b{4.6em}>{\centering\arraybackslash}b{4.6em}>{\centering\arraybackslash}b{5.2em}>{\centering\arraybackslash}b{5.2em}>{\centering\arraybackslash}b{5.2em}>{\centering\arraybackslash}b{5.2em}>{\centering\arraybackslash}b{5.2em}>{\centering\arraybackslash}b{5.2em}}
\toprule
 & \footnotesize\shortstack{N\\all} & \footnotesize\shortstack{N\\multi-token} & \footnotesize\shortstack{N\\shared-prefix} & \footnotesize\shortstack{Align\\(gated)} & \footnotesize\shortstack{Align (gated)\\multi-token} & \footnotesize\shortstack{Align (gated)\\shared-prefix} & \footnotesize\shortstack{Align\\(MC)} & \footnotesize\shortstack{Align (MC)\\multi-token} & \footnotesize\shortstack{Align (MC)\\shared-prefix} \\
\midrule
\multicolumn{10}{l}{\textit{VQAv2}} \\
LLaVA-1.5-7B & 1000 & 541 & 13 & 0.975 & 0.970 & 0.878 & 0.973 & 0.973 & 0.930 \\
InstructBLIP-7B & 1000 & 536 & 11 & 0.978 & 0.973 & 0.869 & 0.975 & 0.974 & 0.910 \\
Qwen3-4B & 1000 & 572 & 12 & 0.972 & 0.967 & 0.857 & 0.964 & 0.960 & 0.862 \\
\midrule
\multicolumn{10}{l}{\textit{ProtoQA}} \\
Llama-3.1-8B & 1000 & 924 & 47 & 0.950 & 0.948 & 0.844 & 0.939 & 0.943 & 0.844 \\
Mistral-7B & 1000 & 847 & 61 & 0.928 & 0.921 & 0.808 & 0.918 & 0.924 & 0.820 \\
Qwen2.5-7B & 1000 & 934 & 51 & 0.937 & 0.935 & 0.819 & 0.913 & 0.917 & 0.817 \\
\bottomrule
\end{tabular}
}
\vspace{-2mm}
\caption{Alignment of gated readout and Monte Carlo (MC) sampling distributions with human answer distributions ($1-\mathrm{JS}$).}
\label{tab:mc_align}
\end{table*}

\subsection{Comparison to Conventional Calibration}
\label{app:calib_def_compare}

Conventional confidence calibration requires that predictions made with confidence $p$ are correct with probability $p$. For a classifier with predicted label $\hat{Y}$ and confidence $\hat{P}$, this is defined as
\[
\Pr(Y=\hat{Y}\mid\hat{P}=p)=p.
\]
Because each input typically has only one observed outcome, calibration is estimated by grouping predictions with similar confidence and comparing their average confidence with empirical accuracy~\citep{guo2017calibration}. Local calibration further restricts this comparison to neighboring inputs~\citep{luo2022local}.

Our one-shot calibration setting follows the same principle but uses multiple human responses to estimate the answer distribution for an individual input $x$. Given $M$ human responses $y^{(1)},\ldots,y^{(M)}$, the empirical human distribution is
\[
\mathbf{p}_H(x)
=
\frac{1}{M}\sum_{m=1}^{M}\mathbf{e}_{y^{(m)}},
\]
where $\mathbf{e}_y$ denotes the one-hot vector for answer $y$. Instead of estimating outcome frequencies across different inputs, we directly estimate a $K$-answer distribution from repeated human responses to the same input and aim to align the edited model's distribution with this target:
\[
\widehat{\mathbf{p}}_K(x)\approx\mathbf{p}_H(x).
\]

Thus, both conventional and one-shot calibration aim to match predicted probabilities with empirical outcome frequencies, but at different levels of granularity. One-shot calibration applies this principle to individual inputs with multiple plausible answers and further requires the calibrated distribution to generalize to semantically and distributionally equivalent inputs: $\widehat{\mathbf{p}}_K(x') \approx\mathbf{p}_H(x),\,x'\in\mathcal{R}(x).$

\newpage

\section{SoftEdit Details}

\subsection{Automatic Loss Balancing}\label{app:loss_balancing}

Both $\lambda_{\text{loc}}$ and $\lambda_{\text{gen}}$ are set automatically from a pilot set of $N_{\text{pilot}}$ edits evaluated at $\Delta W_e = 0$ (a single forward pass per edit, no optimization).

\paragraph{$\lambda_{\text{loc}}$: Equalizing the regularization terms.}
The natural scales of $\mathcal{L}_{\text{gen}}$ and $\mathcal{L}_{\text{loc}}$ differ across models and datasets.
To bring them to comparable magnitude, we compute the pilot-set means:
\[
\bar{\mathcal{L}}_{\text{gen}} = \frac{1}{N_{\text{pilot}}} \sum_{n=1}^{N_{\text{pilot}}} \mathcal{L}_{\text{gen}}^{(n)}, \qquad
\bar{\mathcal{L}}_{\text{loc}} = \frac{1}{N_{\text{pilot}}} \sum_{n=1}^{N_{\text{pilot}}} \mathcal{L}_{\text{loc}}^{(n)},
\]
and set $\lambda_{\text{loc}}$ to their ratio, rounded to the nearest half-decade grid point:
\[
\lambda_{\text{loc}} = \operatorname{round}\!\left(\frac{\bar{\mathcal{L}}_{\text{gen}}}{\bar{\mathcal{L}}_{\text{loc}}}\right).
\]
This ensures that $\lambda_{\text{loc}} \cdot \mathcal{L}_{\text{loc}}$ is approximately the same scale as $\mathcal{L}_{\text{gen}}$, so the two regularization terms contribute equally regardless of model or dataset.

\paragraph{$\lambda_{\text{gen}}$: Scaling regularization relative to alignment.}
$\lambda_{\text{gen}}$ controls the overall weight of the combined regularization block $\lambda_{\text{gen}}(\mathcal{L}_{\text{gen}} + \lambda_{\text{loc}}\,\mathcal{L}_{\text{loc}})$ relative to the alignment term $\mathcal{L}_{\text{align}}$.
The alignment term is the primary objective and should dominate; $\lambda_{\text{gen}}$ ensures this while allowing the regularization to be promoted when alignment is already well-scaled.

Using the same pilot set, we also compute:
\[
\bar{\mathcal{L}}_{\text{align}} = \frac{1}{N_{\text{pilot}}} \sum_{n=1}^{N_{\text{pilot}}} \mathcal{L}_{\text{align}}^{(n)}.
\]
We set $\lambda_{\text{gen}}$ so that the alignment loss is approximately $\rho$ times larger than the total regularization:
\[
\lambda_{\text{gen}} = \operatorname{clip}\!\left(\operatorname{round}\!\left(\frac{\bar{\mathcal{L}}_{\text{align}}}{\rho \cdot (\bar{\mathcal{L}}_{\text{gen}} + \lambda_{\text{loc}}\,\bar{\mathcal{L}}_{\text{loc}})}\right),\; 1,\; 10\right),
\]
where $\rho$ is a fixed target dominance ratio (we use $\rho = 2$ in all experiments).
The $\operatorname{round}(\cdot)$ function snaps to the same half-decade grid as above.
The $\operatorname{clip}(\cdot,\, 1,\, 10)$ clamp has the following effect:
Lower bound ($\lambda_{\text{gen}} \geq 1$): If alignment is already much larger than the regularization (ratio $> \rho$), $\lambda_{\text{gen}}$ boosts regularization toward the target ratio. If regularization is already comparable to or larger than alignment, $\lambda_{\text{gen}}$ stays at 1---it never suppresses regularization below its natural scale. 
Upper bound ($\lambda_{\text{gen}} \leq 10$): Caps the boost to prevent regularization from overwhelming alignment in extreme cases.
Both $\lambda_{\text{loc}}$ and $\lambda_{\text{gen}}$ are computed once from the pilot set and held fixed for all edits.
The entire procedure adds negligible cost (one forward pass per pilot edit with no gradient computation). 
%  Although automatic loss balancing selects $\lambda_{\text{gen}}=1.0$ for \vlmqwen, we observe that reducing it to $0.5$ modestly improves performance.

\begin{figure}
    \centering
    \includegraphics[width=0.7\linewidth]{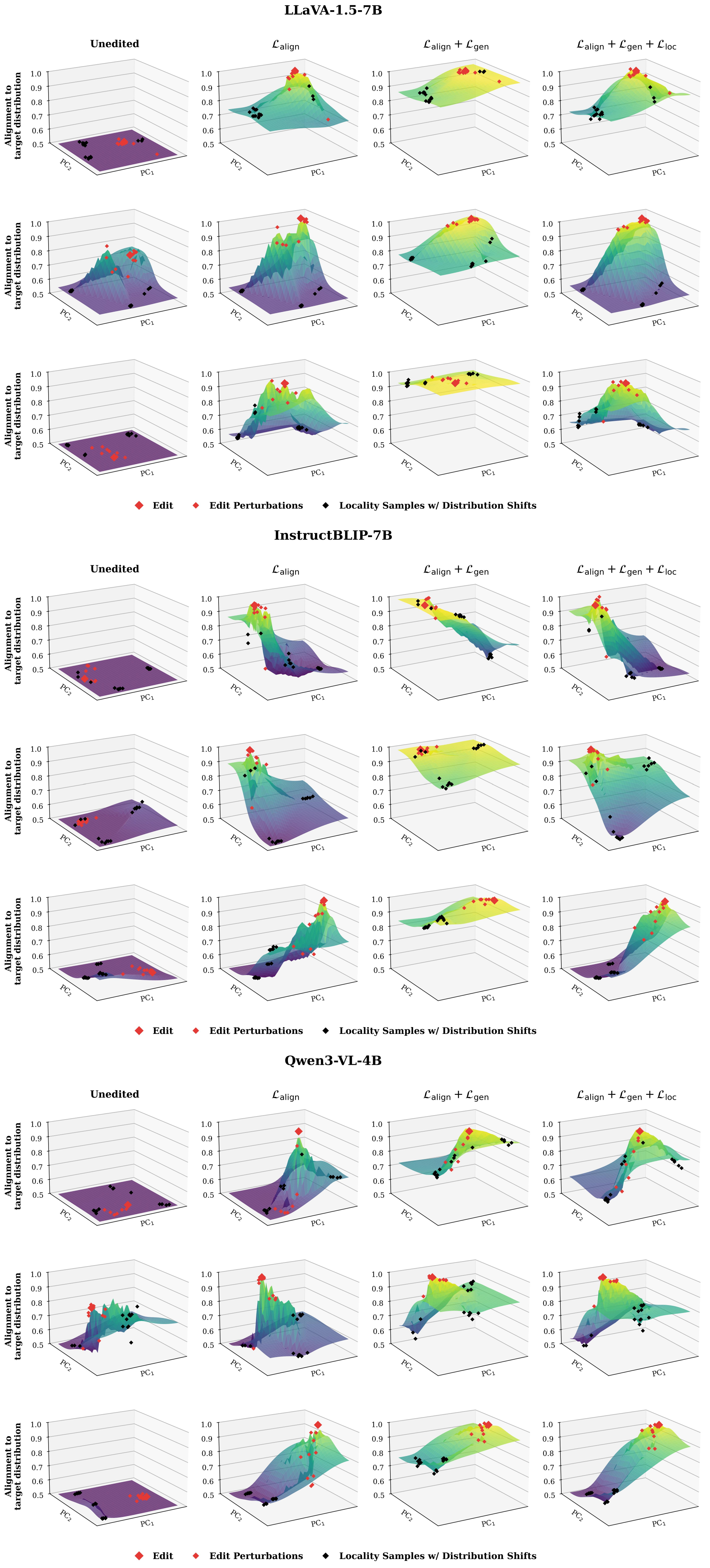}
    \caption{Visualization of changes in the local distribution field of more example edits from VQAv2, when editing three VLMs.}
    \label{fig:local_field_vlms}
\end{figure}

\subsection{Automatic Retrieval Rejection Threshold}\label{app:reject_threshold}

Prior retrieval-based editors~\citep{qiu2026reasonedit} reject out-of-distribution queries using a fixed distance percentile computed from the pairwise distances among codebook keys.
This percentile must be hand-tuned per model and dataset, and its effective meaning shifts as the codebook grows.
We address this with two improvements: a distance threshold estimated from the codebook's own geometry and automatically calibrated, and a paraphrase judge for semantic verification.

\paragraph{Estimating the semantic proximity scale from the codebook.}
The codebook entries are an empirical sample of edit queries from the deployment distribution, where most edits concern different topics.
The nearest-neighbor distances among these entries therefore provide a natural estimate of the distance scale at which ``semantically related'' ends and ``coincidentally nearest'' begins.

Given $N$ codebook key embeddings $\{k_1, \ldots, k_N\}$, we compute each entry's nearest-neighbor distance (excluding itself):
\[
d_i = \min_{j \neq i} \|k_i - k_j\|_2, \qquad i = 1, \ldots, N.
\]
The distribution $\{d_1, \ldots, d_N\}$ characterizes the typical inter-edit spacing in the embedding space.
A new query with embedding $k_q$ is matched to its nearest entry $k_{j^*}$ and rejected if $\|k_q - k_{j^*}\|_2$ exceeds a threshold $\gamma$ set as a percentile of this distribution:
\[
\gamma = \mathrm{percentile}_p(d_1, \ldots, d_N).
\]
Intuitively, a query that is farther from its nearest edit than most edits are from each other is unlikely to be semantically related.
Prior retrieval-based editors~\citep{qiu2026reasonedit} hand-tune this percentile per model and dataset; we instead calibrate it automatically.

\paragraph{Auto-tuning the percentile.}
A lower percentile yields a tighter gate (better locality but risk of missing valid paraphrases); a higher one is more permissive (better generality but risk of false retrievals).
After the first batch of edits, we generate known-related perturbations for a sample of codebook entries by paraphrasing the question and augmenting the image, then encode each perturbation into the key space.
We sweep candidate percentiles and measure retrieval accuracy: whether each perturbation is correctly matched back to its parent edit (passing both the distance gate and the paraphrase judge).
We select the tightest percentile whose accuracy does not drop by more than a tolerance $\delta$ relative to the most lenient gate:
\[
p^* = \min\left\{p : \mathrm{hits}(p) \geq \mathrm{hits}(p_{\max}) - \delta\right\},
\]
where $p_{\max}$ is the loosest candidate and $\delta = 5$ in all experiments.
This adapts the gate to the model's own embedding geometry without manual tuning.

\paragraph{Paraphrase judge.}
Pure distance gating cannot distinguish queries that are close in embedding space but ask about different entities.
As a second filter, a lightweight LLM-based paraphrase judge (\texttt{\small Qwen2.5-1.5B-Instruct}) verifies whether the query and the matched edit's question ask about the same entity.
Given the two questions, the judge produces a Yes/No probability; the query is accepted only if $P(\text{Yes}) \geq 0.5$.
% % 
% When editing VLMs on VQAv2, multimodal embeddings provide better separation, yet the judge catches 19\% (InstructBLIP), 13\% (LLaVA), and 17\% (Qwen3-4B) of distance-matched false retrievals, improving locality by 1.9, 1.4, and 1.4 percentage points respectively.
% %
% When editing LLMs on  ProtoQA, where text-only SBERT embeddings offer less discriminative power, the paraphrase judge rejects 33\%, 36\%, and 39\% of these false retrievals respectively, improving locality by over 21 percentage points.

\subsection{One-Shot Calibration with Ranked Answers}\label{app:ranked}

SoftEdit can also be used when human feedback provides only a \emph{ranking} of plausible answers rather than explicit probability estimates. Such cases reflect realistic scenarios---for example, a clinician may list differential diagnoses in order of likelihood without assigning numeric probabilities to each.

To apply SoftEdit in this setting, we convert the ranked answers into a pseudo-distribution and use it in place of the annotator-frequency target $\mathbf{p}_K$; the SoftEdit objective and all its components ($\mathcal{L}_{\text{align}}$, $\mathcal{L}_{\text{gen}}$, $\mathcal{L}_{\text{loc}}$) remain unchanged.

\paragraph{From rankings to a pseudo-distribution.}
We represent the human ranking as an \emph{ordered partition} of the $K$ answers into $M$ rank groups that preserves ties:
\[
\mathcal{R} = [R_1,\, R_2,\, \ldots,\, R_M],
\]
where $R_1$ contains the most plausible answers, and answers within the same group $R_r$ are tied (unordered). To construct the pseudo-distribution, we first assign an exponential weight to each rank group using a temperature $t> 0$:
\[
w_r = \frac{\exp\!\bigl[-(r-1)/t\bigr]}{\displaystyle\sum_{s=1}^{M} \exp\!\bigl[-(s-1)/t\bigr]},
\]
so that higher-ranked groups receive more mass and the sharpness is controlled by $t$. We then split each group's mass equally among its members. The pseudo-probability of answer $y \in R_r$ is:
\[
\tilde{p}(y) = \frac{w_r}{|R_r|}.
\]
This construction ensures that (i) tied answers always receive equal probability, and (ii) a larger tie group does \emph{not} receive more total mass---rank $r$ always owns $w_r$ regardless of $|R_r|$.
At the extremes, $t \to \infty$ yields a uniform distribution over rank groups (then split within ties), and $t \to 0$ concentrates all mass on the top group $R_1$.
We use $\tilde{p}$ as the target distribution $\mathbf{p}_K$ and apply SoftEdit without modification. Evaluation replaces distributional alignment (JS distance) with ranking alignment (Kendall's $\tau$, Appendix~\ref{app:gated_rank_align}), an ordinal metric that measures pairwise rank concordance between the model's predicted ranking and the human ranking, with proper handling of ties.

\newpage
\section{Experimental Details}

\subsection{Datasets}\label{app:experiment_dataset}

\xhdr{VQAv2} VQAv2~\citep{goyal2017making} is a large-scale visual question answering benchmark built on MS-COCO images, where each question is independently answered by 10 crowd-sourced annotators.
We aggregate the 10 free-form answers per question into a vote-count distribution over unique normalized labels, then apply a multi-stage filter to retain only samples that exhibit genuine multi-annotator uncertainty.
Specifically, we (i)~unify synonymous free-form labels via GPT-4o (\emph{e.g.}\ ``no one'' $\to$ ``nobody''), (ii)~drop samples with non-alphabetic or multi-word labels, and yes/no-only polls, (iii)~merge hyponym--hypernym pairs and remove residual ambiguous samples, and (iv)~retain only single-target questions. 
\textit{To reduce label noise and preserve samples containing high aleatoric uncertainty} due to genuine ambiguity or subjective disagreement, we drop any singleton labels (count\,$=$\,1), and keep only samples in the top 50th percentile of vote-distribution empirical entropy.
After all preprocessing stages, the final VQAv2 benchmark dataset comprises 19{,}186 multi-annotator samples---a large-scale evaluation set with sufficient coverage to benchmark sequential model editing across thousands of edits per backbone.

\xhdr{ProtoQA} ProtoQA~\citep{boratko2020protoqa} is a commonsense question-answering dataset, where human respondents provide free-form answers that are clustered into canonical answer groups with associated vote counts---yielding a distribution over plausible natural language answers per question.
We load both the train and validation splits from HuggingFace and normalize each cluster's canonical answer to lowercase.
\textit{To retain meaningful distributional disagreement among respondents and reduce label noise}, we keep only the top-3 answer clusters by vote count per question as the gold distribution and discard questions with fewer than two surviving clusters after this truncation.
The resulting benchmark dataset contains 9{,}762 multi-answer commonsense questions, providing a large pool for evaluating editing methods on LLMs.

\xhdr{Robustness to unfiltered data}
We further evaluate SoftEdit and the adapted soft editors on 1,000 edits from each dataset with relaxed preprocessing. For VQAv2, we sample 1,000 previously excluded samples that did not pass our filtering criteria. For ProtoQA, we retain all available answer clusters rather than restricting each question to the top 3. As shown in Table~\ref{tab:main_results_left}, SoftEdit continues to achieve the optimal soft generality and soft locality across all backbone models. 

\begin{table*}[t]
\centering

% (1) VLM Editing on VQAv2 (1000 unfiltered samples)
\resizebox{\textwidth}{!}{%
\setlength{\tabcolsep}{3pt}%
\begin{NiceTabular}{l>{\centering\arraybackslash }b{2.0em}>{\centering\arraybackslash }b{2.0em}>{\centering\arraybackslash }b{2.0em}>{\centering\arraybackslash }b{2.0em}>{\centering\arraybackslash }b{2.0em}>{\centering\arraybackslash }b{2.0em}>{\centering\arraybackslash }b{2.0em}>{\centering\arraybackslash }b{2.0em}>{\centering\arraybackslash }b{2.0em}>{\centering\arraybackslash }b{2.0em}>{\centering\arraybackslash }b{2.0em}>{\centering\arraybackslash }b{2.0em}>{\centering\arraybackslash }b{2.0em}>{\centering\arraybackslash }b{2.0em}>{\centering\arraybackslash }b{2.0em}>{\centering\arraybackslash }b{2.0em}>{\centering\arraybackslash }b{2.0em}>{\centering\arraybackslash }b{2.0em}>{\centering\arraybackslash }b{2.0em}>{\centering\arraybackslash }b{2.0em}>{\centering\arraybackslash }b{2.0em}>{\centering\arraybackslash }b{2.0em}>{\centering\arraybackslash }b{2.0em}>{\centering\arraybackslash }b{2.0em}}
\CodeBefore
\begin{tikzpicture}
  \fill[orange!6] (2-|2) rectangle (9-|6) ;
  \fill[orange!6] (2-|10) rectangle (9-|14) ;
  \fill[orange!6] (2-|18) rectangle (9-|22) ;
\end{tikzpicture}
\Body

\toprule
 \Block{2-1}{\textbf{Editor}} & \multicolumn{8}{c}{\textbf{LLaVA-1.5-7B}} & \multicolumn{8}{c}{\textbf{InstructBLIP-7B}} & \multicolumn{8}{c}{\textbf{Qwen3-4B}} \\
\cmidrule(lr){2-5} \cmidrule(lr){6-9} \cmidrule(lr){10-13} \cmidrule(lr){14-17} \cmidrule(lr){18-21} \cmidrule(lr){22-25}
 & \footnotesize Align & \footnotesize SGen-I & \footnotesize SGen-T & \footnotesize SLoc & \footnotesize Acc & \footnotesize Gen-I & \footnotesize Gen-T & \footnotesize Loc & \footnotesize Align & \footnotesize SGen-I & \footnotesize SGen-T & \footnotesize SLoc & \footnotesize Acc & \footnotesize Gen-I & \footnotesize Gen-T & \footnotesize Loc & \footnotesize Align & \footnotesize SGen-I & \footnotesize SGen-T & \footnotesize SLoc & \footnotesize Acc & \footnotesize Gen-I & \footnotesize Gen-T & \footnotesize Loc \\
\midrule
FT-Soft & 0.80 & 0.54 & 0.54 & 0.41 & \textbf{0.98} & \textbf{0.95} & 0.96 & 0.25 & 0.69 & 0.61 & 0.61 & 0.52 & 0.96 & 0.86 & \textbf{0.96} & 0.50 & 0.82 & 0.56 & 0.61 & 0.51 & 0.85 & 0.81 & 0.82 & 0.27 \\
MEND-Soft & 0.06 & 0.21 & 0.18 & 0.17 & 0.15 & 0.30 & 0.24 & 0.39 & 0.10 & 0.21 & 0.11 & 0.09 & 0.34 & 0.38 & 0.22 & 0.47 & 0.12 & 0.06 & 0.08 & 0.00 & 0.21 & 0.19 & 0.18 & 0.21 \\
GRACE-Soft & 0.91 & 0.57 & 0.49 & 0.50 & 0.92 & 0.78 & 0.66 & \textbf{0.99} & 0.86 & 0.55 & 0.52 & 0.50 & 0.84 & 0.56 & 0.63 & \textbf{0.97} & \textbf{0.92} & 0.36 & 0.39 & 0.45 & 0.88 & 0.47 & 0.50 & \textbf{0.92} \\
BalancEdit-Soft & 0.32 & 0.27 & 0.13 & 0.31 & 0.27 & 0.44 & 0.43 & 0.35 & 0.18 & 0.33 & 0.27 & 0.31 & 0.28 & 0.47 & 0.34 & 0.42 & 0.14 & 0.23 & 0.20 & 0.33 & 0.30 & 0.51 & 0.41 & 0.58 \\
ReasonEdit-Soft & 0.54 & 0.41 & 0.40 & 0.38 & 0.90 & 0.92 & 0.94 & 0.58 & 0.60 & 0.51 & 0.53 & 0.38 & 0.96 & 0.89 & 0.83 & 0.87 & 0.62 & \textbf{0.63} & 0.54 & 0.55 & \textbf{0.92} & \textbf{0.89} & 0.86 & 0.69 \\
\midrule
\textbf{SoftEdit} & \textbf{0.94} & \textbf{0.71} & \textbf{0.84} & \textbf{0.54} & \textbf{0.98} & 0.90 & \textbf{0.97} & 0.88 & 0.89 & \textbf{0.81} & \textbf{0.79} & \textbf{0.59} & \textbf{0.97} & \textbf{0.92} & \textbf{0.96} & 0.73 & \textbf{0.92} & \textbf{0.63} & \textbf{0.76} & \textbf{0.56} & 0.89 & 0.88 & \textbf{0.87} & 0.88 \\
\bottomrule
\CodeAfter
\begin{tikzpicture}

\end{tikzpicture}
\end{NiceTabular}
}\par
\vspace{1.5mm}
{\small (1) VLM Editing on VQAv2 (1000 unfiltered samples)}\par
\vspace{1mm}

% (2) LLM editing on protoqa (1000 unfiltered samples)
\resizebox{0.8000\textwidth}{!}{%
\setlength{\tabcolsep}{3pt}%
\begin{NiceTabular}{l>{\centering\arraybackslash }b{2.0em}>{\centering\arraybackslash }b{2.0em}>{\centering\arraybackslash }b{2.0em}>{\centering\arraybackslash }b{2.0em}>{\centering\arraybackslash }b{2.0em}>{\centering\arraybackslash }b{2.0em}>{\centering\arraybackslash }b{2.0em}>{\centering\arraybackslash }b{2.0em}>{\centering\arraybackslash }b{2.0em}>{\centering\arraybackslash }b{2.0em}>{\centering\arraybackslash }b{2.0em}>{\centering\arraybackslash }b{2.0em}>{\centering\arraybackslash }b{2.0em}>{\centering\arraybackslash }b{2.0em}>{\centering\arraybackslash }b{2.0em}>{\centering\arraybackslash }b{2.0em}>{\centering\arraybackslash }b{2.0em}>{\centering\arraybackslash }b{2.0em}}
\CodeBefore
\begin{tikzpicture}
  \fill[orange!6] (2-|2) rectangle (9-|5) ;
  \fill[orange!6] (2-|8) rectangle (9-|11) ;
  \fill[orange!6] (2-|14) rectangle (9-|17) ;
\end{tikzpicture}
\Body

\toprule
 \Block{2-1}{\textbf{Editor}} & \multicolumn{6}{c}{\textbf{Llama-3.1-8B}} & \multicolumn{6}{c}{\textbf{Mistral-7B}} & \multicolumn{6}{c}{\textbf{Qwen2.5-7B}} \\
\cmidrule(lr){2-4} \cmidrule(lr){5-7} \cmidrule(lr){8-10} \cmidrule(lr){11-13} \cmidrule(lr){14-16} \cmidrule(lr){17-19}
 & \footnotesize Align & \footnotesize SGen & \footnotesize SLoc & \footnotesize Acc & \footnotesize Gen & \footnotesize Loc & \footnotesize Align & \footnotesize SGen & \footnotesize SLoc & \footnotesize Acc & \footnotesize Gen & \footnotesize Loc & \footnotesize Align & \footnotesize SGen & \footnotesize SLoc & \footnotesize Acc & \footnotesize Gen & \footnotesize Loc \\
\midrule
FT-Soft & 0.85 & 0.57 & 0.40 & \textbf{0.98} & 0.80 & 0.31 & 0.75 & 0.50 & 0.40 & \textbf{0.97} & 0.80 & 0.35 & 0.89 & 0.50 & 0.37 & \textbf{0.99} & 0.83 & 0.24 \\
MEND-Soft & 0.25 & 0.22 & 0.06 & 0.43 & 0.42 & 0.15 & 0.11 & 0.04 & 0.15 & 0.16 & 0.13 & 0.14 & 0.09 & 0.07 & 0.06 & 0.13 & 0.16 & 0.10 \\
GRACE-Soft & \textbf{0.99} & 0.43 & 0.16 & 0.31 & 0.22 & \textbf{0.94} & 0.88 & 0.13 & 0.13 & 0.52 & 0.05 & \textbf{0.99} & \textbf{0.93} & 0.37 & 0.25 & 0.68 & 0.28 & \textbf{0.99} \\
IKE-Soft & 0.01 & 0.00 & 0.13 & 0.24 & 0.16 & 0.61 & 0.14 & 0.17 & 0.15 & 0.24 & 0.31 & 0.44 & 0.29 & 0.27 & 0.27 & 0.89 & 0.64 & 0.44 \\
AlphaEdit-Soft & 0.35 & 0.25 & 0.20 & 0.60 & 0.35 & 0.36 & 0.41 & 0.13 & 0.21 & 0.72 & 0.49 & 0.28 & 0.34 & 0.07 & 0.20 & 0.56 & 0.16 & 0.33 \\
\midrule
\textbf{SoftEdit} & \textbf{0.99} & \textbf{0.71} & \textbf{0.43} & 0.96 & \textbf{0.84} & 0.66 & \textbf{0.92} & \textbf{0.63} & \textbf{0.47} & 0.95 & \textbf{0.94} & 0.61 & \textbf{0.93} & \textbf{0.63} & \textbf{0.45} & 0.95 & \textbf{0.92} & 0.69 \\
\bottomrule
\CodeAfter
\begin{tikzpicture}

\end{tikzpicture}
\end{NiceTabular}
}\par
\vspace{1.5mm}
{\small (2) LLM editing on protoqa (1000 unfiltered samples)}\par
\vspace{-2mm}
\caption{Batch editing results on 1,000 unfiltered raw samples from each of VQAv2 and ProtoQA.}
\label{tab:main_results_left}
\end{table*}

\subsection{Backbone VLMs and LLMs}\label{app:experiment_models}

\xhdr{Edited VLMs}
\vlmllava~\citep{liu2024improved} connects a CLIP ViT-L/14 vision encoder to a Vicuna-7B language model through a two-layer MLP projector and is trained using a two-stage visual instruction-tuning pipeline.
\vlmblip~\citep{dai2023instructblip} is an instruction-tuned BLIP-2 variant that employs a frozen image encoder and an instruction-aware Q-Former to extract visual features conditioned on the input instruction, which are projected into a frozen Vicuna-7B language model for generation.
\vlmqwen~\citep{yang2025qwen3} employs a Vision Transformer for variable-resolution image encoding and integrates multi-level visual features into a 4B-parameter transformer decoder through DeepStack, with Interleaved-MRoPE for unified spatial--temporal positional modeling.

\xhdr{Edited LLMs}
\llmllama~\citep{grattafiori2024llama} is an 8B-parameter decoder-only transformer with grouped query attention and rotary position embeddings, supporting a 128K-token context window and post-trained for instruction following.
\llmmistral~\mbox{\citep{jiang2023mistral7b}} is a 7B-parameter decoder-only transformer that employs sliding window attention and grouped query attention for efficient long-context inference, with an instruction-tuned variant for instruction following.
\llmqwen~\citep{qwen2025qwen25technicalreport} is a 7B-parameter decoder-only transformer with grouped query attention, rotary position embeddings, and SwiGLU activations, post-trained for instruction following and general-purpose language understanding and generation.

\subsection{Baseline Model Editing Methods}\label{app:experiment_baselines}

We compare SoftEdit against state-of-the-art model editing methods adapted for one-shot calibration of pretrained large language models.
Three baseline editing methods---Fine-tuning, MEND, and GRACE---have been adapted to edit both LLMs and VLMs. We additionally include the state-of-the-art VLM editors BalancEdit and ReasonEdit, and the text-only editors IKE and AlphaEdit for LLMs.
All baselines are adapted to a \emph{soft-label} variant that uses the multi-answer distribution instead of a single target answer to supervise the editing, ensuring each method has access to the same distributional target as SoftEdit.
Specifically, (1) Fine-tuning (FT) has been adopted to update the weight of a selected layer of the pretrained model on each edit via gradient descent.
It is the simplest weight-editing baseline but is prone to catastrophic collapse as edits accumulate.
(2) MEND~\citep{mitchell2021fast} trains a lightweight meta-network that decomposes and transforms the raw fine-tuning gradient into a low-rank parameter update, enabling fast edits without per-edit iterative optimization.
(3) GRACE~\citep{hartvigsen2023aging} maintains a discrete key--value codebook alongside the frozen model, where each edit stores a hidden activation as a key and learns a corresponding replacement activation as its value; at inference time, the value is retrieved only when an input falls within the associated local influence radius.
(4) In-context Knowledge Editing (IKE)\citep{zheng2023can, guo2025balancedit} is a retrieval-based, training-free editor that represents edits as textual demonstrations and retrieves relevant edits at inference time to prepend to the input context, guiding the model toward the updated knowledge without modifying its parameters.
(5) BalancEdit~\citep{guo2025balancedit} extends localized codebook editing to multimodal models by generating positive and negative samples for each edit to estimate its influence scope, which is encoded into a discrete latent-space codebook without modifying the underlying model weights.
(6) ReasonEdit~\citep{qiu2026reasonedit} stores human-provided reasoning in a codebook and uses a topology-balanced multimodal embedding method to retrieve relevant reasoning at inference time to guide generation.
(7) AlphaEdit~\citep{fang2025alphaedit} is a locate-then-edit method that constrains parameter updates by projecting them onto the null space of preserved-knowledge activations, reducing interference with previously stored knowledge during sequential editing.

We preserve the original editing mechanisms of prior editors as closely as possible. In addition, we devoted extra tuning effort to GRACE's retrieval radius using 100 edits for each backbone model to improve its generality on the new dataset. We also used a batch size of 10 instead of 1 for AlphaEdit in sequential editing to improve its performance and ensure competitive and fair comparisons.

\subsection{Comparison to Post-training Model Calibration Methods}\label{app:experiment_baselines_calib}

We compare SoftEdit with five post-training calibration methods: three global calibration methods and two retrieval-based variants we adapt for one-shot localized calibration. 

\xhdr{Global calibrators}
(1) Temperature Scaling (TS)~\citep{guo2017calibration} learns a single global temperature from calibration data to rescale model logits, adjusting prediction confidence without changing the predicted ranking. We adapt it to our distributional target by optimizing the temperature to minimize the KL divergence between the temperature-scaled first-token distribution and the human answer distribution across all edits.
(2) Batch Calibration (BC)~\citep{zhou2024batch} estimates prediction biases from average output scores over a batch of inputs and removes these biases to calibrate predictions. We adopt its labeled variant (BCL), extending the bias correction to the full token vocabulary and selecting the correction strength $\gamma$ via grid search to minimize KL divergence against human answer distributions.
(3) Adaptive Temperature Scaling (ATS)~\citep{xie2024calibrating} trains a lightweight calibration module to predict token-specific temperatures from hidden states, enabling input-dependent calibration of generated token probabilities. We adapt its training objective by replacing the correctness-dependent loss with KL divergence against the target distribution.

\xhdr{Adapted one-shot local calibrators}
Since these methods are originally designed for global calibration and assume a coherent miscalibration pattern, they may struggle with heterogeneous miscalibration across individual samples. For a more direct comparison, we adapt TS (ATS reduces to the same form in the one-shot setting) and BC into one-shot retrieval-based variants, preserving their core calibration ideas while enabling localized calibration for individual edits.
(4) TS-Retrieval fits an independent temperature $T_e$ for each edit by minimizing KL divergence against its target distribution. The temperature is stored alongside the edit's final-layer last-token hidden state as the retrieval key. At inference, the nearest stored key (by L2 distance) provides the temperature for logit scaling.
(5) BC-Retrieval adopts a simple distribution lookup approach, directly storing each edit's human first-token distribution $q_e$ alongside its final-layer last-token hidden state as the retrieval key. At inference, the nearest stored key (by L2 distance) provides \(q_e\) to replace the model's predicted first-token distribution for a nearby sample.

\begin{table*}[t]
\centering

% (1) VLM Calibration on VQAv2 (19,186 samples)
\resizebox{\textwidth}{!}{%
\setlength{\tabcolsep}{3pt}%
\begin{tabular}{l>{\centering\arraybackslash }b{2.0em}>{\centering\arraybackslash }b{2.0em}>{\centering\arraybackslash }b{2.0em}>{\centering\arraybackslash }b{2.0em}>{\centering\arraybackslash }b{2.0em}>{\centering\arraybackslash }b{2.0em}>{\centering\arraybackslash }b{2.0em}>{\centering\arraybackslash }b{2.0em}>{\centering\arraybackslash }b{2.0em}>{\centering\arraybackslash }b{2.0em}>{\centering\arraybackslash }b{2.0em}>{\centering\arraybackslash }b{2.0em}>{\centering\arraybackslash }b{2.0em}>{\centering\arraybackslash }b{2.0em}>{\centering\arraybackslash }b{2.0em}>{\centering\arraybackslash }b{2.0em}>{\centering\arraybackslash }b{2.0em}>{\centering\arraybackslash }b{2.0em}>{\centering\arraybackslash }b{2.0em}>{\centering\arraybackslash }b{2.0em}>{\centering\arraybackslash }b{2.0em}>{\centering\arraybackslash }b{2.0em}>{\centering\arraybackslash }b{2.0em}>{\centering\arraybackslash }b{2.0em}}
\toprule
 \multirow{3.5}{*}{\textbf{Editor}} & \multicolumn{8}{c}{\textbf{LLaVA-1.5-7B (4,436 errors)}} & \multicolumn{8}{c}{\textbf{InstructBLIP-7B (4,625 errors)}} & \multicolumn{8}{c}{\textbf{Qwen3-4B (4,157 errors)}} \\
\cmidrule(lr){2-5} \cmidrule(lr){6-9} \cmidrule(lr){10-13} \cmidrule(lr){14-17} \cmidrule(lr){18-21} \cmidrule(lr){22-25}
 & \footnotesize Align & \footnotesize SGen-I & \footnotesize SGen-T & \footnotesize SLoc & \footnotesize Acc & \footnotesize Gen-I & \footnotesize Gen-T & \footnotesize Loc & \footnotesize Align & \footnotesize SGen-I & \footnotesize SGen-T & \footnotesize SLoc & \footnotesize Acc & \footnotesize Gen-I & \footnotesize Gen-T & \footnotesize Loc & \footnotesize Align & \footnotesize SGen-I & \footnotesize SGen-T & \footnotesize SLoc & \footnotesize Acc & \footnotesize Gen-I & \footnotesize Gen-T & \footnotesize Loc \\
\midrule
Unedited & 0.08 & 0.16 & 0.08 & 0.27 & 0.02 & 0.38 & 0.12 & 1.00 & 0.17 & 0.23 & 0.17 & 0.31 & 0.02 & 0.35 & 0.12 & 1.00 & 0.11 & 0.22 & 0.12 & 0.51 & 0.01 & 0.39 & 0.11 & 1.00 \\
\midrule
TS & 0.03 & 0.05 & 0.03 & 0.06 & 0.02 & 0.38 & 0.12 & 0.98 & 0.17 & 0.22 & 0.16 & 0.27 & 0.00 & 0.35 & 0.12 & 1.00 & 0.13 & 0.21 & 0.13 & 0.31 & 0.01 & 0.39 & 0.11 & 1.00 \\
BCL & 0.00 & 0.00 & 0.00 & 0.00 & 0.03 & 0.38 & 0.13 & 0.98 & 0.17 & 0.23 & 0.17 & 0.29 & 0.02 & 0.36 & 0.12 & 1.00 & 0.09 & 0.17 & 0.10 & 0.26 & 0.01 & 0.39 & 0.11 & 1.00 \\
ATS & 0.03 & 0.02 & 0.03 & 0.03 & 0.02 & 0.38 & 0.12 & 0.98 & 0.18 & 0.23 & 0.17 & 0.28 & 0.02 & 0.36 & 0.12 & 1.00 & 0.14 & 0.21 & 0.14 & 0.29 & 0.01 & 0.39 & 0.11 & 1.00 \\
TS-Retrieval & 0.03 & 0.05 & 0.03 & 0.06 & 0.02 & 0.38 & 0.12 & 1.00 & 0.18 & 0.23 & 0.18 & 0.29 & 0.00 & 0.36 & 0.12 & 1.00 & 0.14 & 0.21 & 0.14 & 0.32 & 0.02 & 0.39 & 0.11 & 1.00 \\
BC-Retrieval & 0.90 & \textbf{0.79} & 0.86 & 0.48 & 0.92 & 0.83 & 0.89 & 0.18 & 0.88 & 0.79 & 0.86 & 0.50 & 0.91 & 0.84 & 0.89 & 0.16 & \textbf{0.89} & \textbf{0.73} & \textbf{0.86} & 0.44 & 0.90 & 0.78 & 0.87 & 0.15 \\
\midrule
\textbf{SoftEdit} & \textbf{0.97} & 0.78 & \textbf{0.92} & \textbf{0.61} & \textbf{0.99} & \textbf{0.92} & \textbf{0.98} & 0.92 & \textbf{0.98} & \textbf{0.82} & \textbf{0.88} & \textbf{0.64} & \textbf{1.00} & \textbf{0.98} & \textbf{0.99} & 0.78 & \textbf{0.89} & 0.67 & 0.82 & \textbf{0.65} & \textbf{0.94} & \textbf{0.88} & \textbf{0.93} & 0.95 \\
\bottomrule
\end{tabular}
}\par
\vspace{1.5mm}
{\small (1) VLM Calibration on VQAv2 (19,186 samples)}\par
\vspace{1mm}

% (2) LLM Calibration on ProtoQA (9,762 samples)
\resizebox{0.8000\textwidth}{!}{%
\setlength{\tabcolsep}{3pt}%
\begin{tabular}{l>{\centering\arraybackslash }b{2.0em}>{\centering\arraybackslash }b{2.0em}>{\centering\arraybackslash }b{2.0em}>{\centering\arraybackslash }b{2.0em}>{\centering\arraybackslash }b{2.0em}>{\centering\arraybackslash }b{2.0em}>{\centering\arraybackslash }b{2.0em}>{\centering\arraybackslash }b{2.0em}>{\centering\arraybackslash }b{2.0em}>{\centering\arraybackslash }b{2.0em}>{\centering\arraybackslash }b{2.0em}>{\centering\arraybackslash }b{2.0em}>{\centering\arraybackslash }b{2.0em}>{\centering\arraybackslash }b{2.0em}>{\centering\arraybackslash }b{2.0em}>{\centering\arraybackslash }b{2.0em}>{\centering\arraybackslash }b{2.0em}>{\centering\arraybackslash }b{2.0em}}
\toprule
 \multirow{3.5}{*}{\textbf{Editor}} & \multicolumn{6}{c}{\textbf{Llama-3.1-8B (8,326 errors)}} & \multicolumn{6}{c}{\textbf{Mistral-7B (7,782 errors)}} & \multicolumn{6}{c}{\textbf{Qwen2.5-7B (7,207 errors)}} \\
\cmidrule(lr){2-4} \cmidrule(lr){5-7} \cmidrule(lr){8-10} \cmidrule(lr){11-13} \cmidrule(lr){14-16} \cmidrule(lr){17-19}
 & \footnotesize Align & \footnotesize SGen & \footnotesize SLoc & \footnotesize Acc & \footnotesize Gen & \footnotesize Loc & \footnotesize Align & \footnotesize SGen & \footnotesize SLoc & \footnotesize Acc & \footnotesize Gen & \footnotesize Loc & \footnotesize Align & \footnotesize SGen & \footnotesize SLoc & \footnotesize Acc & \footnotesize Gen & \footnotesize Loc \\
\midrule
Unedited & 0.04 & 0.04 & 0.07 & 0.01 & 0.03 & 1.00 & 0.03 & 0.03 & 0.11 & 0.02 & 0.07 & 1.00 & 0.04 & 0.06 & 0.17 & 0.00 & 0.10 & 1.00 \\
\midrule
TS & 0.02 & 0.02 & 0.02 & 0.00 & 0.04 & 1.00 & 0.01 & 0.01 & 0.03 & 0.00 & 0.08 & 1.00 & 0.02 & 0.02 & 0.07 & 0.01 & 0.09 & 1.00 \\
BCL & 0.04 & 0.03 & 0.06 & 0.00 & 0.04 & 1.00 & 0.00 & 0.00 & 0.00 & 0.01 & 0.08 & 1.00 & 0.02 & 0.02 & 0.05 & 0.01 & 0.09 & 0.99 \\
ATS & 0.02 & 0.02 & 0.06 & 0.00 & 0.04 & 1.00 & 0.02 & 0.00 & 0.00 & 0.00 & 0.08 & 1.00 & 0.02 & 0.02 & 0.06 & 0.00 & 0.09 & 1.00 \\
TS-Retrieval & 0.02 & 0.02 & 0.02 & 0.00 & 0.03 & 1.00 & 0.02 & 0.01 & 0.04 & 0.01 & 0.07 & 1.00 & 0.03 & 0.03 & 0.08 & 0.00 & 0.10 & 1.00 \\
BC-Retrieval & 0.56 & 0.52 & 0.22 & 0.35 & 0.32 & 0.12 & 0.57 & 0.53 & 0.20 & 0.58 & 0.53 & 0.13 & 0.52 & 0.47 & 0.20 & 0.58 & 0.54 & 0.15 \\
\midrule
\textbf{SoftEdit} & \textbf{0.95} & \textbf{0.72} & \textbf{0.47} & \textbf{0.98} & \textbf{0.94} & 0.76 & \textbf{0.93} & \textbf{0.70} & \textbf{0.52} & \textbf{0.98} & \textbf{0.95} & 0.69 & \textbf{0.93} & \textbf{0.73} & \textbf{0.52} & \textbf{0.98} & \textbf{0.94} & 0.77 \\
\bottomrule
\end{tabular}
}\par
\vspace{1.5mm}
{\small (2) LLM Calibration on ProtoQA (9,762 samples)}\par
\vspace{-2mm}
\caption{Comparison to post-training calibration baselines on VLMs and LLMs using the VQAv2 and ProtoQA datasets. TS, BCL, and ATS fit a global calibrator on the full edit set; TS-Retrieval and BC-Retrieval retrieve a per-edit temperature scalar or first-token distribution.}
\label{tab:calib_results}
\end{table*}

\xhdr{Results} As shown in Table~\ref{tab:calib_results}, SoftEdit is significantly more effective than post-training calibration methods for heterogeneous, localized calibration of individual edits.
First, all three global calibrators fail to improve model generation or calibration over the unedited baseline, even when trained on the provided edit set where the model made errors. We identify two main reasons. (1) These methods assume a coherent global miscalibration pattern, while our edits require heterogeneous, localized corrections. Learning a global temperature, a probability bias vector, or a temperature prediction module can suffer from averaging across diverse calibration needs. (2) These methods adjust token probabilities but do not ensure correct answer ranking and generation. In particular, temperature scaling preserves the original ranking of answers. When the model already ranks an incorrect answer above the human-provided answers, changing the temperature alone cannot correct the ranking.
For the adapted one-shot retrieval-based calibrators, TS-Retrieval still fails due to the second reason above: learning an individual temperature for each edit cannot calibrate original-incorrectly ranked answers.
BC-Retrieval (directly retrieve the correct first-token probability) improves performance but remains limited for two reasons. (1) Overwriting the same distribution across a semantic neighborhood ignores genuine distribution shifts among nearby samples, resulting in inferior soft locality (SLoc). (2) Replacing only the first-token distribution encourages the model to start with the correct tokens but cannot ensure that it continues to generate the complete answer correctly.

\subsection{Additional Ablations} \label{app:additional_ablation}

\xhdr{Ablation of retrieval rejection threshold} The retrieval rejection threshold controls the conventional generality--locality tradeoff in retrieval-based model editing by preventing edits from generalizing to semantically unrelated queries, similar to distance-based regularization in prior work~\citep{hartvigsen2023aging,guo2025balancedit,qiu2026reasonedit}. We compare our automatically selected threshold with fixed distance percentiles of 1, 10, and 50. As shown in Table~\ref{tab:ablate_reject}, higher percentiles allow retrieval over larger distances, improving generality and soft generality but reducing locality. In contrast, lower percentiles restrict retrieval, preserving locality but limiting generality and soft generality.
The auto-tuned threshold improves generality, soft generality and soft locality while preserving relatively high conventional locality.

\begin{table*}[t]
\centering

% (1) VLM Editing on VQAv2 (19,186 samples)
\resizebox{\textwidth}{!}{%
\setlength{\tabcolsep}{3pt}%
\begin{tabular}{l>{\centering\arraybackslash }b{2.0em}>{\centering\arraybackslash }b{2.0em}>{\centering\arraybackslash }b{2.0em}>{\centering\arraybackslash }b{2.0em}>{\centering\arraybackslash }b{2.0em}>{\centering\arraybackslash }b{2.0em}>{\centering\arraybackslash }b{2.0em}>{\centering\arraybackslash }b{2.0em}>{\centering\arraybackslash }b{2.0em}>{\centering\arraybackslash }b{2.0em}>{\centering\arraybackslash }b{2.0em}>{\centering\arraybackslash }b{2.0em}>{\centering\arraybackslash }b{2.0em}>{\centering\arraybackslash }b{2.0em}>{\centering\arraybackslash }b{2.0em}>{\centering\arraybackslash }b{2.0em}>{\centering\arraybackslash }b{2.0em}>{\centering\arraybackslash }b{2.0em}}
\toprule
 \multirow{3.5}{*}{\textbf{Variant}} & \multicolumn{6}{c}{\textbf{LLaVA-1.5-7B (4,436 errors)}} & \multicolumn{6}{c}{\textbf{InstructBLIP-7B (4,625 errors)}} & \multicolumn{6}{c}{\textbf{Qwen3-4B (4,157 errors)}} \\
\cmidrule(lr){2-4} \cmidrule(lr){5-7} \cmidrule(lr){8-10} \cmidrule(lr){11-13} \cmidrule(lr){14-16} \cmidrule(lr){17-19}
 & \footnotesize Align & \footnotesize SGen & \footnotesize SLoc & \footnotesize Acc & \footnotesize Gen & \footnotesize Loc & \footnotesize Align & \footnotesize SGen & \footnotesize SLoc & \footnotesize Acc & \footnotesize Gen & \footnotesize Loc & \footnotesize Align & \footnotesize SGen & \footnotesize SLoc & \footnotesize Acc & \footnotesize Gen & \footnotesize Loc \\
\midrule
pct 1 & \textbf{0.97} & 0.73 & 0.52 & \textbf{0.99} & 0.74 & \textbf{1.00} & 0.97 & 0.79 & 0.60 & \textbf{1.00} & 0.87 & \textbf{0.85} & \textbf{0.89} & 0.68 & 0.62 & \textbf{0.94} & 0.67 & \textbf{1.00} \\
pct 10 & \textbf{0.97} & 0.77 & 0.56 & \textbf{0.99} & 0.90 & 0.90 & 0.97 & 0.79 & 0.62 & \textbf{1.00} & 0.95 & 0.81 & \textbf{0.89} & 0.73 & 0.64 & \textbf{0.94} & 0.88 & 0.95 \\
pct 50 & \textbf{0.97} & \textbf{0.85} & \textbf{0.61} & \textbf{0.99} & \textbf{0.95} & 0.67 & 0.97 & 0.81 & 0.60 & \textbf{1.00} & 0.97 & 0.60 & \textbf{0.89} & 0.74 & \textbf{0.65} & \textbf{0.94} & \textbf{0.90} & 0.79 \\
auto (SoftEdit) & \textbf{0.97} & \textbf{0.85} & \textbf{0.61} & \textbf{0.99} & \textbf{0.95} & 0.90 & \textbf{0.98} & \textbf{0.85} & \textbf{0.64} & \textbf{1.00} & \textbf{0.99} & 0.78 & \textbf{0.89} & \textbf{0.75} & \textbf{0.65} & \textbf{0.94} & \textbf{0.90} & 0.95 \\
\bottomrule
\end{tabular}
}\par
\vspace{1.5mm}
{\small (1) VLM Editing on VQAv2 (19,186 samples)}\par
\vspace{1mm}

% (2) LLM Editing on ProtoQA (9,762 samples)
\resizebox{\textwidth}{!}{%
\setlength{\tabcolsep}{3pt}%
\begin{tabular}{l>{\centering\arraybackslash }b{2.0em}>{\centering\arraybackslash }b{2.0em}>{\centering\arraybackslash }b{2.0em}>{\centering\arraybackslash }b{2.0em}>{\centering\arraybackslash }b{2.0em}>{\centering\arraybackslash }b{2.0em}>{\centering\arraybackslash }b{2.0em}>{\centering\arraybackslash }b{2.0em}>{\centering\arraybackslash }b{2.0em}>{\centering\arraybackslash }b{2.0em}>{\centering\arraybackslash }b{2.0em}>{\centering\arraybackslash }b{2.0em}>{\centering\arraybackslash }b{2.0em}>{\centering\arraybackslash }b{2.0em}>{\centering\arraybackslash }b{2.0em}>{\centering\arraybackslash }b{2.0em}>{\centering\arraybackslash }b{2.0em}>{\centering\arraybackslash }b{2.0em}}
\toprule
 \multirow{3.5}{*}{\textbf{Variant}} & \multicolumn{6}{c}{\textbf{Llama-3.1-8B (8,326 errors)}} & \multicolumn{6}{c}{\textbf{Mistral-7B (7,782 errors)}} & \multicolumn{6}{c}{\textbf{Qwen2.5-7B (7,207 errors)}} \\
\cmidrule(lr){2-4} \cmidrule(lr){5-7} \cmidrule(lr){8-10} \cmidrule(lr){11-13} \cmidrule(lr){14-16} \cmidrule(lr){17-19}
 & \footnotesize Align & \footnotesize SGen & \footnotesize SLoc & \footnotesize Acc & \footnotesize Gen & \footnotesize Loc & \footnotesize Align & \footnotesize SGen & \footnotesize SLoc & \footnotesize Acc & \footnotesize Gen & \footnotesize Loc & \footnotesize Align & \footnotesize SGen & \footnotesize SLoc & \footnotesize Acc & \footnotesize Gen & \footnotesize Loc \\
\midrule
pct 1 & \textbf{0.95} & 0.14 & 0.07 & \textbf{0.98} & 0.17 & \textbf{0.81} & \textbf{0.93} & 0.06 & 0.08 & \textbf{0.98} & 0.10 & \textbf{0.80} & \textbf{0.93} & 0.04 & 0.24 & \textbf{0.98} & 0.08 & \textbf{0.80} \\
pct 10 & 0.94 & 0.54 & 0.30 & \textbf{0.98} & 0.69 & 0.75 & \textbf{0.93} & 0.53 & 0.42 & \textbf{0.98} & 0.71 & 0.77 & \textbf{0.93} & 0.62 & 0.40 & \textbf{0.98} & 0.81 & 0.79 \\
pct 50 & 0.94 & 0.70 & 0.44 & \textbf{0.98} & 0.91 & 0.70 & \textbf{0.93} & 0.68 & 0.51 & \textbf{0.98} & 0.92 & 0.61 & \textbf{0.93} & 0.71 & 0.51 & \textbf{0.98} & 0.92 & 0.71 \\
auto (SoftEdit) & \textbf{0.95} & \textbf{0.72} & \textbf{0.47} & \textbf{0.98} & \textbf{0.94} & 0.76 & \textbf{0.93} & \textbf{0.70} & \textbf{0.52} & \textbf{0.98} & \textbf{0.95} & 0.69 & \textbf{0.93} & \textbf{0.73} & \textbf{0.52} & \textbf{0.98} & \textbf{0.94} & 0.77 \\
\bottomrule
\end{tabular}
}\par
\vspace{1.5mm}
{\small (2) LLM Editing on ProtoQA (9,762 samples)}\par
\vspace{-2mm}
\caption{Retrieval-threshold ablation.}
\label{tab:ablate_reject}
\end{table*}

\xhdr{Ablation of LLM paraphrase judge}
At inference, before applying a retrieved weight perturbation, we use a lightweight LLM judge (\texttt{\small Qwen2.5-1.5B-Instruct}) to verify whether a new query is a paraphrase of its matched edit's query, further distinguishing subtle semantic differences between queries with similar embeddings. As shown in Table~\ref{tab:ablate_judge}, for VLMs, the judge provides limited additional benefits, as multimodal embeddings capture subtle visual information along with textual differences to distinguish queries. However, for LLMs, the judge effectively differentiates subtle semantic differences in texts missed by embedding-only retrieval, notably improving locality (Loc) while maintaining comparable performance on other metrics. 

\xhdr{Inference cost of retrieval and paraphrase judge}
On an A100 GPU, the lightweight \texttt{\small Qwen2.5-1.5B-Instruct} paraphrase judge adds a modest overhead of 16.1 ± 1.0 ms/query. Including key encoding and lookup, the total inference time increases only from 31.1 to 31.2 ms/query as the number of edits grows from 100 to 4,000. Codebook growth has minimal impact on inference cost: the mean nearest-neighbor retrieval time increases only from 0.04 to 0.20 ms/query.

\begin{table*}[t]
\centering

% (1) VLM Editing on VQAv2 (19,186 samples)
\resizebox{\textwidth}{!}{%
\setlength{\tabcolsep}{3pt}%
\begin{tabular}{l>{\centering\arraybackslash }b{2.0em}>{\centering\arraybackslash }b{2.0em}>{\centering\arraybackslash }b{2.0em}>{\centering\arraybackslash }b{2.0em}>{\centering\arraybackslash }b{2.0em}>{\centering\arraybackslash }b{2.0em}>{\centering\arraybackslash }b{2.0em}>{\centering\arraybackslash }b{2.0em}>{\centering\arraybackslash }b{2.0em}>{\centering\arraybackslash }b{2.0em}>{\centering\arraybackslash }b{2.0em}>{\centering\arraybackslash }b{2.0em}>{\centering\arraybackslash }b{2.0em}>{\centering\arraybackslash }b{2.0em}>{\centering\arraybackslash }b{2.0em}>{\centering\arraybackslash }b{2.0em}>{\centering\arraybackslash }b{2.0em}>{\centering\arraybackslash }b{2.0em}}
\toprule
 \multirow{3.5}{*}{\textbf{}} & \multicolumn{6}{c}{\textbf{LLaVA-1.5-7B (4,436 errors)}} & \multicolumn{6}{c}{\textbf{InstructBLIP-7B (4,625 errors)}} & \multicolumn{6}{c}{\textbf{Qwen3-4B (4,157 errors)}} \\
\cmidrule(lr){2-4} \cmidrule(lr){5-7} \cmidrule(lr){8-10} \cmidrule(lr){11-13} \cmidrule(lr){14-16} \cmidrule(lr){17-19}
 & \footnotesize Align & \footnotesize SGen & \footnotesize SLoc & \footnotesize Acc & \footnotesize Gen & \footnotesize Loc & \footnotesize Align & \footnotesize SGen & \footnotesize SLoc & \footnotesize Acc & \footnotesize Gen & \footnotesize Loc & \footnotesize Align & \footnotesize SGen & \footnotesize SLoc & \footnotesize Acc & \footnotesize Gen & \footnotesize Loc \\
\midrule
no judge & 0.98 & 0.85 & 0.61 & 1.00 & 0.96 & 0.93 & 0.97 & 0.85 & 0.64 & 1.00 & 0.99 & 0.75 & 0.89 & 0.75 & 0.65 & 0.95 & 0.91 & 0.95 \\
with judge (SoftEdit) & 0.97 & 0.85 & 0.61 & 0.99 & 0.95 & 0.92 & 0.98 & 0.85 & 0.64 & 1.00 & 0.99 & \textbf{0.78} & 0.89 & 0.75 & 0.65 & 0.94 & 0.90 & 0.95 \\
\bottomrule
\end{tabular}
}\par
\vspace{1.5mm}
{\small (1) VLM Editing on VQAv2 (19,186 samples)}\par
\vspace{1mm}

% (2) LLM Editing on ProtoQA (9,762 samples)
\resizebox{\textwidth}{!}{%
\setlength{\tabcolsep}{3pt}%
\begin{tabular}{l>{\centering\arraybackslash }b{2.0em}>{\centering\arraybackslash }b{2.0em}>{\centering\arraybackslash }b{2.0em}>{\centering\arraybackslash }b{2.0em}>{\centering\arraybackslash }b{2.0em}>{\centering\arraybackslash }b{2.0em}>{\centering\arraybackslash }b{2.0em}>{\centering\arraybackslash }b{2.0em}>{\centering\arraybackslash }b{2.0em}>{\centering\arraybackslash }b{2.0em}>{\centering\arraybackslash }b{2.0em}>{\centering\arraybackslash }b{2.0em}>{\centering\arraybackslash }b{2.0em}>{\centering\arraybackslash }b{2.0em}>{\centering\arraybackslash }b{2.0em}>{\centering\arraybackslash }b{2.0em}>{\centering\arraybackslash }b{2.0em}>{\centering\arraybackslash }b{2.0em}}
\toprule
 \multirow{3.5}{*}{\textbf{}} & \multicolumn{6}{c}{\textbf{Llama-3.1-8B (8,326 errors)}} & \multicolumn{6}{c}{\textbf{Mistral-7B (7,782 errors)}} & \multicolumn{6}{c}{\textbf{Qwen2.5-7B (7,207 errors)}} \\
\cmidrule(lr){2-4} \cmidrule(lr){5-7} \cmidrule(lr){8-10} \cmidrule(lr){11-13} \cmidrule(lr){14-16} \cmidrule(lr){17-19}
 & \footnotesize Align & \footnotesize SGen & \footnotesize SLoc & \footnotesize Acc & \footnotesize Gen & \footnotesize Loc & \footnotesize Align & \footnotesize SGen & \footnotesize SLoc & \footnotesize Acc & \footnotesize Gen & \footnotesize Loc & \footnotesize Align & \footnotesize SGen & \footnotesize SLoc & \footnotesize Acc & \footnotesize Gen & \footnotesize Loc \\
\midrule
no judge & 0.96 & 0.73 & 0.47 & 1.00 & 0.94 & 0.54 & 0.94 & 0.71 & 0.52 & 0.99 & 0.95 & 0.35 & 0.94 & 0.73 & 0.50 & 0.98 & 0.95 & 0.54 \\
with judge (SoftEdit) & 0.95 & 0.72 & 0.47 & 0.98 & 0.94 & \textbf{0.76} & 0.93 & 0.70 & 0.52 & 0.98 & 0.95 & \textbf{0.69} & 0.93 & 0.73 & 0.52 & 0.98 & 0.94 & \textbf{0.77} \\
\bottomrule
\end{tabular}
}\par
\vspace{1.5mm}
{\small (2) LLM Editing on ProtoQA (9,762 samples)}\par
\vspace{-2mm}
\caption{Ablation of the lightweight LLM paraphrase judge (Qwen2.5-1.5B-Instruct).}
\label{tab:ablate_judge}
\end{table*}

\xhdr{Ablation of weighted $A_{ij}$ in graph-Laplacian regularization} We ablate the probability-weighted graph-Laplacian regularization by replacing it with unweighted and Frobenius-norm alternatives. As shown in Table~\ref{tab:ablate_graph}, both alternatives achieve generally competitive performance but exhibit decreases in soft generality (SGen) and soft locality (SLoc), with variations across models. Overall, the probability-weighted graph-Laplacian regularization achieves the optimal performances across calibration and conventional editing metrics. We thus adopt weighted formulation to emphasize relationships between more plausible answers, while alternative regularization formulations can also encourage coherent changes in answer probabilities.

\begin{table*}[t]
\centering

% (1) VLM Editing on VQAv2 (19,186 samples)
\resizebox{\textwidth}{!}{%
\setlength{\tabcolsep}{3pt}%
\begin{tabular}{l>{\centering\arraybackslash }b{2.0em}>{\centering\arraybackslash }b{2.0em}>{\centering\arraybackslash }b{2.0em}>{\centering\arraybackslash }b{2.0em}>{\centering\arraybackslash }b{2.0em}>{\centering\arraybackslash }b{2.0em}>{\centering\arraybackslash }b{2.0em}>{\centering\arraybackslash }b{2.0em}>{\centering\arraybackslash }b{2.0em}>{\centering\arraybackslash }b{2.0em}>{\centering\arraybackslash }b{2.0em}>{\centering\arraybackslash }b{2.0em}>{\centering\arraybackslash }b{2.0em}>{\centering\arraybackslash }b{2.0em}>{\centering\arraybackslash }b{2.0em}>{\centering\arraybackslash }b{2.0em}>{\centering\arraybackslash }b{2.0em}>{\centering\arraybackslash }b{2.0em}}
\toprule
 \multirow{3.5}{*}{\textbf{Variant}} & \multicolumn{6}{c}{\textbf{LLaVA-1.5-7B (4,436 errors)}} & \multicolumn{6}{c}{\textbf{InstructBLIP-7B (4,625 errors)}} & \multicolumn{6}{c}{\textbf{Qwen3-4B (4,157 errors)}} \\
\cmidrule(lr){2-4} \cmidrule(lr){5-7} \cmidrule(lr){8-10} \cmidrule(lr){11-13} \cmidrule(lr){14-16} \cmidrule(lr){17-19}
 & \footnotesize Align & \footnotesize SGen & \footnotesize SLoc & \footnotesize Acc & \footnotesize Gen & \footnotesize Loc & \footnotesize Align & \footnotesize SGen & \footnotesize SLoc & \footnotesize Acc & \footnotesize Gen & \footnotesize Loc & \footnotesize Align & \footnotesize SGen & \footnotesize SLoc & \footnotesize Acc & \footnotesize Gen & \footnotesize Loc \\
\midrule
frobenius & \textbf{0.97} & 0.84 & 0.60 & \textbf{0.99} & \textbf{0.95} & 0.90 & 0.95 & 0.83 & 0.62 & 0.96 & 0.98 & \textbf{0.79} & 0.87 & \textbf{0.75} & \textbf{0.65} & 0.93 & \textbf{0.92} & 0.93 \\
unweighted & 0.96 & 0.83 & 0.60 & 0.98 & 0.94 & 0.91 & 0.96 & \textbf{0.85} & 0.63 & 0.99 & \textbf{0.99} & 0.78 & 0.80 & 0.72 & 0.60 & 0.88 & 0.87 & 0.93 \\
weighted (SoftEdit) & \textbf{0.97} & \textbf{0.85} & \textbf{0.61} & \textbf{0.99} & \textbf{0.95} & \textbf{0.92} & \textbf{0.98} & \textbf{0.85} & \textbf{0.64} & \textbf{1.00} & \textbf{0.99} & 0.78 & \textbf{0.89} & \textbf{0.75} & \textbf{0.65} & \textbf{0.94} & 0.90 & \textbf{0.95} \\
\bottomrule
\end{tabular}
}\par
\vspace{1.5mm}
{\small (1) VLM Editing on VQAv2 (19,186 samples)}\par
\vspace{1mm}

% (2) LLM Editing on ProtoQA (9,762 samples)
\resizebox{\textwidth}{!}{%
\setlength{\tabcolsep}{3pt}%
\begin{tabular}{l>{\centering\arraybackslash }b{2.0em}>{\centering\arraybackslash }b{2.0em}>{\centering\arraybackslash }b{2.0em}>{\centering\arraybackslash }b{2.0em}>{\centering\arraybackslash }b{2.0em}>{\centering\arraybackslash }b{2.0em}>{\centering\arraybackslash }b{2.0em}>{\centering\arraybackslash }b{2.0em}>{\centering\arraybackslash }b{2.0em}>{\centering\arraybackslash }b{2.0em}>{\centering\arraybackslash }b{2.0em}>{\centering\arraybackslash }b{2.0em}>{\centering\arraybackslash }b{2.0em}>{\centering\arraybackslash }b{2.0em}>{\centering\arraybackslash }b{2.0em}>{\centering\arraybackslash }b{2.0em}>{\centering\arraybackslash }b{2.0em}>{\centering\arraybackslash }b{2.0em}}
\toprule
 \multirow{3.5}{*}{\textbf{Variant}} & \multicolumn{6}{c}{\textbf{Llama-3.1-8B (8,326 errors)}} & \multicolumn{6}{c}{\textbf{Mistral-7B (7,782 errors)}} & \multicolumn{6}{c}{\textbf{Qwen2.5-7B (7,207 errors)}} \\
\cmidrule(lr){2-4} \cmidrule(lr){5-7} \cmidrule(lr){8-10} \cmidrule(lr){11-13} \cmidrule(lr){14-16} \cmidrule(lr){17-19}
 & \footnotesize Align & \footnotesize SGen & \footnotesize SLoc & \footnotesize Acc & \footnotesize Gen & \footnotesize Loc & \footnotesize Align & \footnotesize SGen & \footnotesize SLoc & \footnotesize Acc & \footnotesize Gen & \footnotesize Loc & \footnotesize Align & \footnotesize SGen & \footnotesize SLoc & \footnotesize Acc & \footnotesize Gen & \footnotesize Loc \\
\midrule
frobenius & 0.93 & 0.70 & 0.40 & 0.97 & 0.92 & 0.74 & 0.90 & \textbf{0.72} & 0.48 & 0.96 & 0.91 & \textbf{0.69} & 0.91 & \textbf{0.73} & 0.48 & 0.95 & 0.93 & \textbf{0.77} \\
unweighted & 0.92 & 0.69 & 0.40 & \textbf{0.98} & 0.92 & 0.74 & 0.90 & 0.70 & 0.41 & 0.96 & 0.91 & 0.68 & 0.91 & \textbf{0.73} & 0.47 & 0.95 & 0.93 & \textbf{0.77} \\
weighted (SoftEdit) & \textbf{0.95} & \textbf{0.72} & \textbf{0.47} & \textbf{0.98} & \textbf{0.94} & \textbf{0.76} & \textbf{0.93} & 0.70 & \textbf{0.52} & \textbf{0.98} & \textbf{0.95} & \textbf{0.69} & \textbf{0.93} & \textbf{0.73} & \textbf{0.52} & \textbf{0.98} & \textbf{0.94} & \textbf{0.77} \\
\bottomrule
\end{tabular}
}\par
\vspace{1.5mm}
{\small (2) LLM Editing on ProtoQA (9,762 samples)}\par
\vspace{-2mm}
\caption{Ablation of weighted $A_{ij}$ in graph-Laplacian regularization.}
\label{tab:ablate_graph}
\end{table*}

\xhdr{Sensitivity analysis of $\lambda_{\mathrm{loc}}$ and $\lambda_{\mathrm{gen}}$ auto-tuned using the first $k$ edits}
As shown in Table~\ref{tab:ablate_lambda}, the auto-tuned coefficients are robust to the number of edits used for estimation. Across all six models, the coefficients remain unchanged for $k \geq 50$. We also note that auto-tuning is computationally efficient, requiring only a single forward pass per edit through the original unedited model to compute the initial loss terms, without additional optimization. We use $k=100$ in all experiments, while both $k=50$ and $k=100$ serve as reasonable defaults.

\begin{table*}[t]
\centering
\resizebox{0.7\textwidth}{!}{%
\setlength{\tabcolsep}{3pt}%
\begin{tabular}{lcccccccccccc}
\toprule
 \multirow{3.5}{*}{\textbf{$k$}} & \multicolumn{2}{c}{\textbf{LLaVA-1.5-7B}} & \multicolumn{2}{c}{\textbf{InstructBLIP-7B}} & \multicolumn{2}{c}{\textbf{Qwen3-4B}} & \multicolumn{2}{c}{\textbf{Llama-3.1-8B}} & \multicolumn{2}{c}{\textbf{Mistral-7B}} & \multicolumn{2}{c}{\textbf{Qwen2.5-7B}} \\
\cmidrule(lr){2-3} \cmidrule(lr){4-5} \cmidrule(lr){6-7} \cmidrule(lr){8-9} \cmidrule(lr){10-11} \cmidrule(lr){12-13}
 & $\lambda_{\mathrm{loc}}$ & $\lambda_{\mathrm{gen}}$ & $\lambda_{\mathrm{loc}}$ & $\lambda_{\mathrm{gen}}$ & $\lambda_{\mathrm{loc}}$ & $\lambda_{\mathrm{gen}}$ & $\lambda_{\mathrm{loc}}$ & $\lambda_{\mathrm{gen}}$ & $\lambda_{\mathrm{loc}}$ & $\lambda_{\mathrm{gen}}$ & $\lambda_{\mathrm{loc}}$ & $\lambda_{\mathrm{gen}}$ \\
\midrule
10 & 0.005 & 1 & 0.05 & 1 & 0.01 & 1 & 0.001 & 1 & 0.0001 & 10 & 0.0005 & 5 \\
50 & 0.005 & 1 & 0.05 & 1 & 0.05 & 1 & 0.001 & 1 & 0.0001 & 10 & 0.0005 & 5 \\
100 & 0.005 & 1 & 0.05 & 1 & 0.05 & 1 & 0.001 & 1 & 0.0001 & 10 & 0.0005 & 5 \\
1000 & 0.005 & 1 & 0.05 & 1 & 0.05 & 1 & 0.001 & 1 & 0.0001 & 10 & 0.0005 & 5 \\
\bottomrule
\end{tabular}
}
\caption{Robust auto-tuned $\lambda_{\mathrm{loc}}$ and $\lambda_{\mathrm{gen}}$ estimated from the first $k$ edits.}
\label{tab:ablate_lambda}
\end{table*}

% We ablate the retrieval-key embedding for LLM edits by replacing sentence encoder \texttt{\small all-MiniLM-L6-v2} with the last-token hidden state from the final layer. This improves locality by around $0.07$ on \llmmistral, while all other performance metrics remain generally comparable. A comparison table is provided in Table~\ref{tab:ablate_key}.

% \begin{table*}[t]
% \centering
% \resizebox{\textwidth}{!}{%
% \begin{tabular}{lcccccc|cccccc|cccccc}
% \toprule
%  & \multicolumn{6}{c}{Llama-3.1-8B} & \multicolumn{6}{|c}{Mistral-7B} & \multicolumn{6}{|c}{Qwen2.5-7B} \\
%  & \footnotesize Acc & \footnotesize Gen & \footnotesize Loc & \footnotesize Align & \footnotesize SGen & \footnotesize SLoc & \footnotesize Acc & \footnotesize Gen & \footnotesize Loc & \footnotesize Align & \footnotesize SGen & \footnotesize SLoc & \footnotesize Acc & \footnotesize Gen & \footnotesize Loc & \footnotesize Align & \footnotesize SGen & \footnotesize SLoc \\
% \midrule
% Sentence Encoder & 0.98 & 0.94 & 0.76 & 0.95 & 0.72 & 0.47 & 0.98 & 0.95 & 0.69 & 0.93 & 0.70 & 0.52 & 0.98 & 0.94 & 0.77 & 0.93 & 0.73 & 0.52 \\
% Last hidden state & 0.98 & 0.93 & 0.72 & 0.95 & 0.72 & 0.44 & 0.98 & 0.95 & 0.76 & 0.93 & 0.71 & 0.50 & 0.98 & 0.91 & 0.75 & 0.93 & 0.71 & 0.54 \\
% \bottomrule
% \end{tabular}
% }
% \caption{Ablation on the retrieval-key embedding method for editing LLMs on ProtoQA.}
% \label{tab:ablate_key}
% \end{table*}

\subsection{Additional Results} \label{app:additional_results}

\begin{table*}[t]
\centering

% (1) SGen-I on VQAv2 (19,186 samples)
\resizebox{\textwidth}{!}{%
\setlength{\tabcolsep}{3pt}%
\begin{NiceTabular}{l>{\centering\arraybackslash }b{4.6em}>{\centering\arraybackslash }b{10em}>{\centering\arraybackslash }b{4.6em}>{\centering\arraybackslash }b{10em}>{\centering\arraybackslash }b{4.6em}>{\centering\arraybackslash }b{10em}}
\toprule
 \Block{2-1}{\textbf{Editor}} & \multicolumn{2}{c}{\textbf{LLaVA-1.5-7B (4,436 errors)}} & \multicolumn{2}{c}{\textbf{InstructBLIP-7B (4,625 errors)}} & \multicolumn{2}{c}{\textbf{Qwen3-4B (4,157 errors)}} \\
\cmidrule(lr){2-3} \cmidrule(lr){4-5} \cmidrule(lr){6-7}
 & \footnotesize \shortstack{SGen-I\\(perturb)} & \footnotesize \shortstack{SGen-I\\(variant)} & \footnotesize \shortstack{SGen-I\\(perturb)} & \footnotesize \shortstack{SGen-I\\(variant)} & \footnotesize \shortstack{SGen-I\\(perturb)} & \footnotesize \shortstack{SGen-I\\(variant)} \\
\midrule
Unedited & 0.104 & 0.216 [0.214, 0.218] & 0.172 & 0.290 [0.287, 0.294] & 0.153 & 0.291 [0.287, 0.296] \\
\midrule
FT & 0.499 & 0.449 [0.445, 0.453] & 0.486 & 0.468 [0.463, 0.473] & 0.482 & 0.422 [0.417, 0.427] \\
MEND & 0.125 & 0.166 [0.166, 0.166] & 0.108 & 0.167 [0.165, 0.168] & 0.082 & 0.097 [0.097, 0.097] \\
GRACE & 0.434 & 0.289 [0.285, 0.292] & 0.394 & 0.352 [0.346, 0.357] & 0.232 & 0.295 [0.290, 0.299] \\
BalancEdit & 0.235 & 0.307 [0.304, 0.310] & 0.211 & 0.312 [0.308, 0.316] & 0.186 & 0.288 [0.285, 0.292] \\
ReasonEdit & 0.399 & 0.394 [0.393, 0.395] & 0.406 & 0.429 [0.428, 0.430] & 0.535 & 0.522 [0.518, 0.526] \\
\midrule
FT-Soft & 0.653 & 0.523 [0.520, 0.527] & 0.624 & 0.550 [0.545, 0.554] & 0.626 & 0.482 [0.478, 0.486] \\
MEND-Soft & 0.131 & 0.173 [0.172, 0.173] & 0.131 & 0.185 [0.185, 0.185] & 0.062 & 0.079 [0.078, 0.079] \\
GRACE-Soft & 0.664 & 0.333 [0.329, 0.337] & 0.571 & 0.397 [0.392, 0.402] & 0.277 & 0.296 [0.292, 0.300] \\
BalancEdit-Soft & 0.226 & 0.296 [0.294, 0.299] & 0.211 & 0.304 [0.301, 0.307] & 0.206 & 0.308 [0.304, 0.312] \\
ReasonEdit-Soft & 0.516 & 0.491 [0.490, 0.492] & 0.519 & 0.535 [0.533, 0.536] & 0.566 & 0.533 [0.527, 0.538] \\
\midrule
\textbf{SoftEdit} & \textbf{0.931} & \textbf{0.624 [0.619, 0.629]} & \textbf{0.881} & \textbf{0.756 [0.748, 0.764]} & \textbf{0.793} & \textbf{0.558 [0.551, 0.565]} \\
\bottomrule
\CodeAfter
\begin{tikzpicture}

\end{tikzpicture}
\end{NiceTabular}
}\par
\vspace{-2mm}
\caption{SGen-I on VQAv2, evaluated separately using photometric perturbations and generated image variants. The variant column is the mean [min, max] across $t\in\{0.0,0.3\}$.}
\label{tab:sgen_i_variant}
\end{table*}

\begin{figure}
    \centering
    \includegraphics[width=1\linewidth]{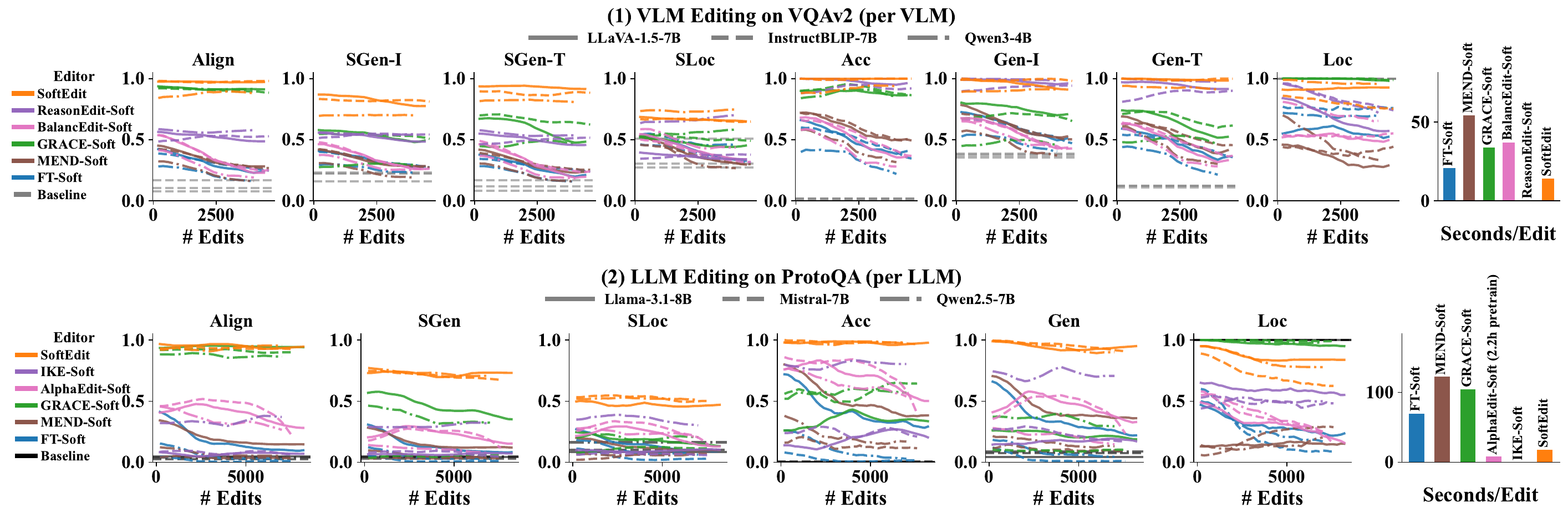}
    \caption{\small Sequential editing performance and computational efficiency across VLM and LLM editors adapted to one-shot calibration. SoftEdit achieves the best alignment, soft generality, and soft locality. Each line corresponds to an edited pretrained model.}
    \label{fig:seq_supp}
\end{figure}

\begin{figure}
    \centering
    \includegraphics[width=1\linewidth]{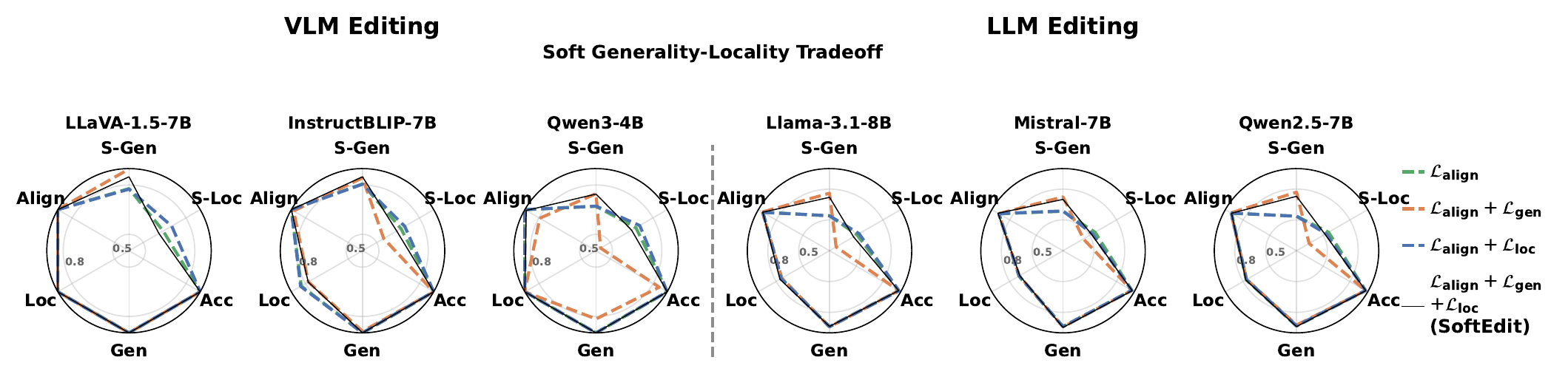}
    \caption{Ablation of all four loss combinations. $\mathcal{L}_{\mathrm{loc}}$ is designed to mitigate overgeneralization introduced by $\mathcal{L}_{\mathrm{gen}}$; thus, optimizing $\mathcal{L}_{\mathrm{align}}+\mathcal{L}_{\mathrm{loc}}$ is not a primary design choice shown in Fig.\ref{fig:ablations}.}
    \label{fig:ablations_tradeoff_supp}
\end{figure}

\xhdr{Separate SGen-I evaluations} In Table~\ref{tab:main_results}, we report the overall image soft generality (SGen-I) of VLM editors. Here, we separately evaluate SGen-I on photometric perturbations and generated image variants (mean [min, max] over $t\in\{0.0,0.3\}$ variants) in Table~\ref{tab:sgen_i_variant}. Photometric perturbations preserve the original image semantics, while generated variants assess generalization to unseen but semantically related images. SoftEdit consistently outperforms all baselines on both evaluation sets across three VLMs. Performance is generally lower on generated variants, which may reflect stronger visual variations and uncertainty in their target distributions. Overall, these results show that SoftEdit's image soft generality improvements extend beyond simple photometric perturbations.

\xhdr{Non-aggregated sequential editing} We report the non-aggregated sequential editing performance for each LLM and VLM individually in Fig.~\ref{fig:seq_supp}.

\xhdr{Additional ablation results on loss functions}
We further ablate all four loss combinations, including $\mathcal{L}_{\text{align}}+\mathcal{L}_{\text{loc}}$, in Fig.~\ref{fig:ablations_tradeoff_supp}. Across all six models, $\mathcal{L}_{\text{align}}+\mathcal{L}_{\text{loc}}$ yields performance similar to $\mathcal{L}_{\text{align}}$.

% \newpage
\section{Limitations}
SoftEdit has limitations. First, SoftEdit focuses on \emph{aleatoric uncertainty}: irreducible uncertainty arising from genuine ambiguity or subjective disagreement, where the observed human answer distribution is treated as reliable feedback. In practice, however, human feedback may also contain annotation noise, unreliable answers, or outliers that should not be interpreted as meaningful uncertainty. Distinguishing these cases is an important direction for future work. For example, properties of the observed answer distribution, such as its shape and entropy, could potentially be combined with outlier detection or annotator-consistency signals to identify unreliable answers.
Second, SoftEdit inherits a scalability limitation of retrieval-based editors (GRACE, IKE, BalancEdit, ReasonEdit).  Although SoftEdit stores only a compact weight perturbation matrix for each edit, the retrieval index and stored perturbations grow with the number of edits. For codebook containing a few thousand edits, as commonly considered in model editing settings, exact nearest-neighbor retrieval is efficient on CPU, with nearly constant latency in practice~\citep{hartvigsen2023aging}. As the codebook grows, more efficient GPU usage in retrieval, or advanced search algorithms on CPUs can mitigate retrieval overhead~\citep{johnson2019billion,malkov2018efficient,jegou2010product}.
% SoftEdit inherits a scalability limitation of retrieval-based editors (GRACE, IKE, BalancEdit, ReasonEdit). Although each edit stores only a compact weight perturbation matrix, the retrieval index and stored perturbations grow with the number of edits, while retrieval introduces additional inference-time overhead. These costs can be mitigated through more efficient GPU usage in retrieval, quantization of stored perturbations, and mechanisms for pruning obsolete entries. 
%
As the first study of one-shot calibration via model editing, we adopt first-token-based probability estimation following prior work~\citep{lee2023can, ma2025large}. 
Our primary objective is to investigate local calibration and its distributional generality--locality tradeoff, using the same evaluation procedure for SoftEdit and all adapted editors.
In our experiments, shared-first-token answers occur in only 4.65--5.06\% of ProtoQA samples and 0.62--0.63\% of VQAv2 samples, as determined by tokenization using the three LLMs and three VLMs, respectively. 
To better distinguish answers sharing the same first token, future work could explore more flexible token-level probabilistic factorizations.
Experimentally, for SGen-I, we acknowledge that the target distributions of generated ambiguous images are approximated through latent interpolation, which may introduce evaluation noise. To account for this uncertainty, we report the average performance across multiple interpolation strengths, together with the minimum and maximum scores.

\end{document}